\documentclass{article} % For LaTeX2e
\usepackage{iclr2027_conference,times}

\usepackage[most]{tcolorbox}
\usepackage{colortbl}

\usepackage{booktabs}
\usepackage{amsmath}
\usepackage{subcaption}

\newcommand{\cellseparator}{%
  \raisebox{-\dp\strutbox}[0pt][0pt]{%
    \vbox to \dimexpr\ht\strutbox+\dp\strutbox\relax{%
      \offinterlineskip
      \leaders\vbox{\kern0.7pt\hbox{\vrule width0.4pt height0.5pt}\kern0.7pt}\vfill
    }%
  }%
}

\usepackage[T1]{fontenc}
\usepackage{lmodern} % Highly recommended alongside T1 to keep the fonts rendering smoothly

\usepackage{amsmath,amsfonts,bm}

\def\eqref#1{equation~\ref{#1}}
\def\1{\bm{1}}

\DeclareMathAlphabet{\mathsfit}{\encodingdefault}{\sfdefault}{m}{sl}
\SetMathAlphabet{\mathsfit}{bold}{\encodingdefault}{\sfdefault}{bx}{n}

\usepackage{tcolorbox}
\tcbuselibrary{listings, breakable, skins}
\usepackage{xcolor}

\newtcblisting{promptlisting}[2][]{
  breakable,
  listing only,
  colback=gray!5,
  colframe=gray!50!black,
  colbacktitle=gray!20,
  coltitle=black,
  fonttitle=\bfseries,
  title=#2,
  listing options={
    basicstyle=\ttfamily\scriptsize,
    breaklines=true,
    breakatwhitespace=false,
    columns=fullflexible,
    keepspaces=true,
    showstringspaces=false,
  },
  #1
}

\usepackage{graphicx}
\usepackage{placeins}   % \FloatBarrier
\usepackage{float} 

\usepackage{caption}

\usepackage{amssymb} % Required by existing appendix formulas (e.g., \geqslant).

\usepackage{hyperref}
\usepackage{url}
\usepackage{longtable}
\usepackage{array}
\usepackage{graphicx}
\usepackage{float}
\usepackage{placeins}
\usepackage{etoolbox}

\usepackage{listings}
\usepackage{tcolorbox}
\tcbuselibrary{listings, breakable}
\makeatletter
\patchcmd{\LT@output}{\copy\LT@foot\vss}{\copy\LT@foot\vfil}{}{}
\patchcmd{\LT@output}{\copy\LT@foot\vss}{\copy\LT@foot\vfil}{}{}
\makeatother

\title{AbductionBench: Benchmarking Abductive Reasoning in Large Language Models}

\newcommand{\authorentry}[3]{%
  \begin{minipage}[t][31pt][t]{\linewidth}\centering\setlength{\parskip}{0pt}%
  {\fontsize{9}{11}\selectfont\bfseries #1\textsuperscript{\normalfont\ensuremath{#2}}\par}
  \vspace{1pt}%
  {\fontsize{7}{9}\selectfont\urlstyle{same}\href{mailto:#3}{\nolinkurl{#3}}\par}%
  \end{minipage}%
}
\newcommand{\arxivAuthors}{%
\begin{tabular}{@{}>{\centering\arraybackslash}p{0.32\textwidth}@{\hspace{0.02\textwidth}}>{\centering\arraybackslash}p{0.32\textwidth}@{\hspace{0.02\textwidth}}>{\centering\arraybackslash}p{0.32\textwidth}@{}}
\authorentry{Moein Salimi}{1}{s.salimi.moein@gmail.com} &
\authorentry{Amir Qeysarbeigi}{1*}{amir.gheysarbeygi@gmail.com} &
\authorentry{Babak Hosseini Mohtasham}{2*}{babak.hosseini.m@ut.ac.ir} \\[4pt]
\authorentry{Faraz Doagoye Tehrani}{1}{faraztehrani226@gmail.com} &
\authorentry{Mahdi Naieni}{2}{mhd.naeeni@gmail.com} &
\authorentry{Hanieh Sartipi}{3}{haniehsartipi@gmail.com} \\[4pt]
\authorentry{Ali Nouri}{1}{alinr1556@gmail.com} &
\authorentry{Amin Aghakasiri}{2\dagger}{kasiri.amin@gmail.com} &
\authorentry{Reyhane Khayatzade Mahani}{1\dagger}{Reyhaneh.kh1384@gmail.com} \\[4pt]
\authorentry{Mahdi Jafari Siavoshani}{1\ddagger}{mjafari@sharif.edu} &
\authorentry{Mohammad Hossein Rohban}{1\ddagger}{rohban@sharif.edu} &
 \\[4pt]
\end{tabular}
\par\vspace{1pt}
{\fontsize{9}{11}\selectfont\setlength{\parskip}{0pt}%
\textsuperscript{1}Sharif University of Technology\quad
\textsuperscript{2}University of Tehran\par
\textsuperscript{3}Tarbiat Modares University\par}
}
\author{\arxivAuthors}

\makeatletter
\newcommand{\arxivAuthorFootnotes}{%
  \begingroup
  \Hy@hyperfootnotesfalse
  \renewcommand{\thefootnote}{\fnsymbol{footnote}}%
  \footnotetext[1]{Equal contribution.}%
  \footnotetext[2]{Equal contribution.}%
  \footnotetext[3]{Equal corresponding authors.}%
  \endgroup
}
\makeatother

\iclrfinalcopy
\makeatletter
\renewcommand{\@maketitle}{%
  \vbox{\hsize\textwidth
    {\LARGE\scshape\@title\par}
    \vskip 0.15in
    {\centering\@author\par}
    \vskip 0.15in
  }%
}
\makeatother
\renewcommand{\headrulewidth}{0pt}
\date{}
\hypersetup{
  hidelinks,
  pdftitle={AbductionBench: Benchmarking Abductive Reasoning in Large Language Models},
  pdfauthor={Moein Salimi, Amir Qeysarbeigi, Babak Hosseini Mohtasham, Faraz Doagoye Tehrani, Mahdi Naieni, Hanieh Sartipi, Ali Nouri, Amin Aghakasiri, Reyhane Khayatzade Mahani, Mahdi Jafari Siavoshani, Mohammad Hossein Rohban},
  pdfsubject={Abductive reasoning benchmark},
  pdfkeywords={abductive reasoning, large language models, benchmarks}
}
\renewenvironment{abstract}{\vskip.075in\centerline{\large\scshape Abstract}\vspace{0.5ex}\begin{quote}\small}{\end{quote}\vskip 1ex}

\begin{document}

\raggedbottom

\maketitle
\arxivAuthorFootnotes
\begin{abstract}
Abductive reasoning is the inference of plausible explanations for observed evidence, particularly when information is incomplete or supports multiple competing hypotheses. Evaluating this capability requires more than a single task: a system may need to generate an explanation, discriminate among competing hypotheses, revise its hypotheses as evidence arrives, or decide what evidence to seek next. Yet these capabilities are typically evaluated through isolated benchmarks with different task formulations and scoring procedures, so claims that a model is ``good at abduction'' may reflect only a narrow aspect of this capability. We introduce \textsc{AbductionBench}\footnote{Anonymous code repository: \url{https://anonymous.4open.science/r/AbductionBench-6150/README.md}.}, a unified framework that brings this fragmented landscape into a common evaluation space. Its taxonomy characterizes abductive tasks by hypothesis format, application domain, and evidence-delivery mode, covering generation and selection tasks across static, passive, and interactive settings while preserving their task-specific execution structure. We instantiate this framework on 27 datasets spanning ten domains and evaluate eight LLMs from four model families---GPT, Gemini, Gemma, and Qwen---alongside Jev, a non-LLM System One decision system evaluated on single-choice selection tasks. Beyond harmonizing heterogeneous tasks, AbductionBench separates final-answer evaluation from analysis of the observable reasoning traces and interaction histories produced alongside those answers. We analyze written traces as observable model outputs rather than faithful representations of internal computation, characterizing evidence use, hypothesis comparison, uncertainty, contradiction handling, and answer commitment. Chain-of-thought is not uniformly beneficial across models and tasks, and within datasets, longer traces are associated with lower accuracy. Jev matches or exceeds small LLMs when selecting among explanations over compact evidence but falls substantially behind on longer cases, suggesting that candidate discrimination and extended evidence integration place different demands on a system. In interactive tasks, the proportion of actions judged relevant declines at later steps, and a relevant first question is associated with a correct final answer; in the passive task, models rarely revise their initial hypothesis as new evidence arrives. Together, these findings highlight the value of evaluating abductive reasoning through both final answers and observable evidence use, hypothesis revision, and information gathering.
\end{abstract}
\section{Introduction}
\label{sec:introduction}

Abductive reasoning seeks plausible explanations for observed evidence. When information is incomplete, several hypotheses may fit the same observations, and new evidence may change which explanation is best supported. For large language models (LLMs), a correct final answer alone does not show whether the model used the relevant evidence, considered alternatives, or revised an earlier hypothesis. Evaluation therefore needs to examine outcomes alongside observable reasoning and evidence-gathering behavior, without treating written traces as direct records of internal computation.

Existing benchmarks address parts of this problem. Abductive Commonsense Reasoning evaluates narrative explanation selection and generation \citep{Bhagavatula2020Abductive}; GEAR evaluates generated hypothesis sets on four programmable benchmarks \citep{impl_he2025gear}. Because each benchmark fixes its own formulation and scoring, results do not show how abductive ability varies with the kind of hypothesis required, the domain, or the way evidence arrives. Most benchmarks also supply all evidence at once, whereas practical abduction---clinical diagnosis, incident response, threat attribution---requires revising a hypothesis as evidence arrives or deciding which evidence to seek.

AbductionBench addresses this gap. It brings existing tasks into one execution and reporting protocol and describes each along three axes: hypothesis generation or selection, application domain, and evidence delivery, which is \emph{static} (all evidence at the outset), \emph{passive} (evidence arrives in a fixed order and the model revises its hypothesis after each turn), or \emph{interactive} (the model's questions or actions determine what it observes next). It scores final answers with task-appropriate checks or an LLM judge, and separately analyzes chain-of-thought (CoT) traces for evidence use, hypothesis comparison, uncertainty, and commitment, and multi-turn episodes for question relevance and hypothesis revision.

\begin{figure}[t]
  \centering
  \includegraphics[width=0.62\linewidth]{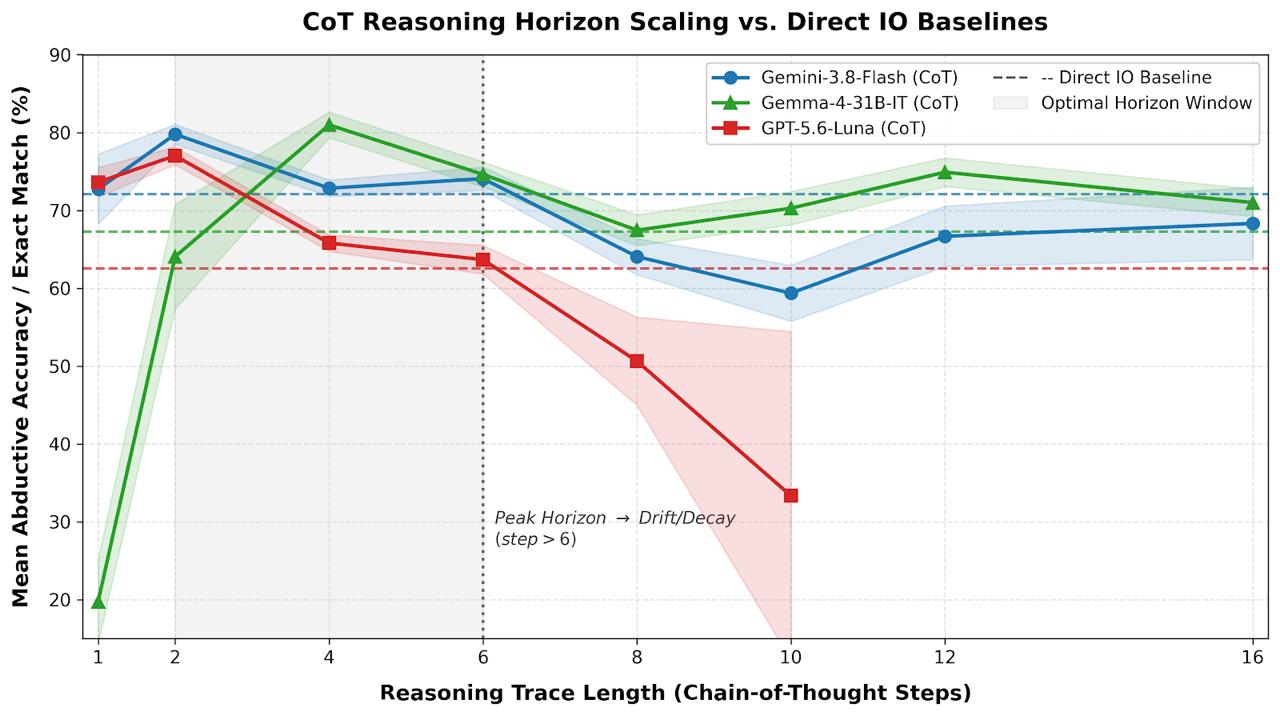}
  \caption{Accuracy by CoT trace length (steps) against each model's IO baseline (dashed); bands are $\pm$SEM, points with $n<5$ omitted.}
  \label{fig:horizon_scaling}
\end{figure}

We evaluate eight LLMs from four model families on 27 datasets (21 static, five interactive, one passive), together with Jev, a non-LLM System~One decision system, on the single-choice tasks. CoT helps when an explanation must be derived by chaining evidence but often hurts open-ended hypothesis generation.Accuracy peaks for short reasoning chains and declines as traces grow longer, falling below the direct-answer baseline for two of three models, at every input length (Figure~\ref{fig:horizon_scaling}; Appendix~\ref{app:horizon_by_context}). Jev is competitive with small LLMs when selecting among explanations over compact evidence but falls behind on long cases. In interactive diagnosis, a relevant first question goes with a correct final answer and relevance declines as the interaction proceeds; with passively arriving evidence, models rarely revise their first hypothesis.

Our contributions are:
\begin{itemize}
    \item \textbf{A structured suite} of 27 datasets spanning 
    domains, hypothesis generation and selection, and three modes of evidence delivery (static, passive, interactive), including a passive adaptation of AthenaBench, with source-task adaptations documented.
    \item \textbf{A complementary evaluation protocol} that reports final-answer outcomes alongside trace measures for static tasks and turn-level measures for passive and interactive episodes.
    \item \textbf{An empirical study} of eight LLMs and a non-LLM System~One decision system (Jev), covering 69,599 static responses and 2,400 passive and interactive episodes.
\end{itemize}
\section{Related Work}
\label{sec:related_work}

Abduction seeks plausible explanations for observations rather than conclusions that must be true \citep{schurz2008patterns}. Prior benchmarks isolate individual pieces of this problem: ART's $\alpha$NLI asks models to select the more plausible explanation connecting two observations, while e-CARE includes both a two-choice causal-reasoning task and an explanation-generation task \citep{Bhagavatula2020Abductive,du2022ecare}. $\alpha$NLG and UNcommonsense focus on open-ended explanation generation, whereas True Detective and MuSR evaluate answer selection in longer narratives \citep{impl_zhao2023uncommonsense,del2022truedetective,sprague2024musr}. ProofWriter and AbductionRules study abduction over natural-language rule bases, and HypoSpace evaluates set-valued hypothesis generation under underdetermination \citep{impl_tafjord2020proofwriter,young2022abductionrules,chen2026hypospace}. In clinical settings, MedCaseReasoning evaluates diagnosis together with alignment to clinician-authored reasoning, while DDXPlus supports interactive evidence collection and differential diagnosis \citep{wu2025medcasereasoning,fansitchango2022ddxplus}. On the evaluation side, IBE-Eval scores explanations by consistency, parsimony, coherence, and uncertainty \citep{dalal2024ibe}, and RECV reports higher error on claim verification that requires abduction from incomplete evidence \citep{dougrezlewis2024recv}.

Closest to our work, GEAR \citep{impl_he2025gear} evaluates sets of executable hypotheses by consistency, generalizability, and diversity, but is restricted to domains where hypotheses compile to code and has no candidate-selection task, so it cannot separate generating good candidates from discriminating among them. \emph{Wiring the ``Why''} \citep{salimi2026wiring} surveys over 60 papers and compares generation and selection empirically, but with a fixed prompt per benchmark, a small set of models, and static benchmarks only. AbductionBench evaluates generation and selection as distinct tasks across 27 datasets and ten domains, under static, passive, and interactive evidence delivery, and reports reasoning-trace and interaction measures alongside outcome accuracy.

\section{Benchmark Scope and Design}
\label{sec:framework}
\label{sec:methodology}

AbductionBench brings existing tasks into a common evaluation interface. This section defines which tasks fall within its scope, how they are categorized, and how datasets are integrated; execution and scoring are specified in Section~\ref{sec:evaluation_protocol}.

\subsection{What Counts as an Abductive Task}
\label{sec:framework_scope}

We include benchmarks designed to evaluate abductive reasoning and selected tasks from other benchmarks. For each task, we identify what is given, what the model must infer, and why that inference is abductive: usually, the model infers a hypothesis that explains supplied observations. Inclusion is decided at the task or subset level rather than from a dataset's domain label.

\subsection{Task Taxonomy}
\label{sec:task_representation}

A \emph{task configuration} specifies a task from a dataset or subset, the information available to the model, the required output, and how evidence is supplied; a dataset contributes one configuration per task format it supports. We describe each configuration along three axes.

\subsubsection{Hypothesis Generation and Selection}
\label{sec:methodology_task_formulation}

Following \citet{salimi2026wiring}, \textbf{generation} produces one or more candidate explanations $H=\{h_1,\ldots,h_n\}$ for observations $O$, whereas \textbf{selection} evaluates supplied candidates against the observations and returns the choice or subset the task requires, as in ART's $\alpha$NLI \citep{Bhagavatula2020Abductive}. Selection is posed as single choice (SCS) or, where several candidates may be correct, multiple choice (MCS). When a dataset supports both generation and selection, we evaluate each as a separate configuration. These terms describe the benchmark's observable inputs and outputs, not stages of the model's internal computation.

\subsubsection{Application Domains}
\label{sec:methodology_domain_taxonomy}

Each configuration receives one primary-domain label: Healthcare, Commonsense, Narrative Reasoning, Formal Reasoning, Event Reasoning, Scientific Discovery, Scientific Reasoning, Computing Systems, Cybersecurity, or Agentic Systems (definitions in Appendix~\ref{app:domain_taxonomy}).

\subsubsection{Evidence-Delivery Mode}
\label{sec:methodology_data_delivery}

Evidence-delivery mode describes when the model receives evidence and whether it can influence what evidence arrives next.
\textbf{Static:} all evidence supplied by the task is available at the outset; it may still be incomplete, as is typical in abduction.
\textbf{Passive:} evidence arrives in a predetermined sequence; the model may revise its hypothesis as evidence accumulates, but cannot choose what it receives next.
\textbf{Interactive:} the model's actions or queries determine the information it receives next, so later observations depend on the interaction history.

\subsection{Dataset Selection and Adaptation}
\label{sec:methodology_dataset_selection}

From a larger candidate inventory, we retain tasks that meet the scope above and exclude datasets that are not publicly accessible, are prohibitively expensive to evaluate, are saturated by current models, or duplicate the abductive challenge of another selection. The resulting 27 datasets (Appendix~\ref{app:datasets}) cover all ten domains, both hypothesis formulations, and all three delivery modes. For each configuration, a dataset adapter converts source records into the common interface and connects them to the required evaluation resources, such as reference answers or an interaction environment; changes to the evidence shown, the allowed answers, or the output format are documented as our adaptations.

\paragraph{Passive adaptation of AthenaBench.}
AthenaBench poses threat-actor attribution: given evidence about malicious activity, name the responsible actor. We convert its 57 attribution cases (39 actors) into passive episodes. Each case's 15--25 evidence items---the actor's profile and motivation, techniques with MITRE ATT\&CK identifiers, and documented incidents---are delivered in source order over four to six turns, usually four items per turn. After each turn, the model names one actor while retaining all earlier evidence and predictions; it cannot request, reorder, or stop the evidence, which separates hypothesis revision from evidence acquisition. We keep the source's per-turn question and supply the system prompt and answer format. Because actors have many aliases (e.g., Sednit and APT28), each turn's prediction is judged against the gold actor by an LLM judge rather than by string match (Section~\ref{sec:sequential_interactive_evaluation}).

\section{Evaluation Protocol}
\label{sec:evaluation_protocol}
\label{sec:methodology_evaluation_protocol}

The protocol separates \emph{final-answer evaluation}, which scores the answer or terminal result, from \emph{process analysis}, which characterizes the reasoning trace or interaction history produced alongside it. Full metric definitions are in Appendix~\ref{app:metrics} and all judge prompts in Appendix~\ref{app:prompts}.

\subsection{Prompting and Task Execution}
\label{sec:task_execution}

\textbf{Static tasks} are run under two AbductionBench templates that replace the source prompts: input--output (IO), which requests a direct answer, and chain-of-thought (CoT), which additionally requests an explicit reasoning trace. \textbf{Passive and interactive tasks} are run as multi-turn episodes under IO only, since the interaction itself exposes the process: the source's questions and environments are kept, while the system prompt and answer format are ours. The absence of a written trace under IO is not treated as evidence that a model performs no internal reasoning.

\subsection{Final-Answer Evaluation}
\label{sec:outcome_evaluation}
\label{sec:methodology_metrics}

Single-choice selection (SCS) is scored by accuracy, and multiple-choice selection (MCS) by exact set match (EM) and set F1 averaged over items. Generation tasks with a formally checkable answer (ProofWriter, UniADILR-HGc) are scored by exact match; other free-form explanations receive a binary correct/incorrect verdict from an LLM judge (gpt-oss-120b) that compares them with the gold explanation. Cloud-OpsBench is scored by joint root-cause accuracy: the judged fault type and the faulty component must both be correct.

\subsection{Reasoning-Trace Analysis}
\label{sec:reasoning_trace_evaluation}

For CoT outputs, a segmentation judge splits each trace into steps, and a reasoning judge (gpt-oss-120b) scores each step or trace on thirteen descriptors: evidence use (observation coverage); step quality (useful and helpful steps, backtracking); expressed uncertainty; hypothesis breadth (branchiness, diversity, comparison exhaustiveness); direction and commitment (evidence-to-hypothesis directionality, anchoring point); unsupported content (prior knowledge, unresolved contradiction); and whether the gold hypothesis remains under consideration. These measures describe the reasoning a model writes, not its hidden computation.

\subsection{Passive and Interactive Evaluation}
\label{sec:sequential_interactive_evaluation}

Every episode is scored by final-answer accuracy. For interactive tasks, each action before the final answer (a question, test, or tool call) is judged relevant or irrelevant to the gold answer, giving per-step relevance and the turns taken. For AthenaBench, every turn's prediction is judged against the gold actor, giving per-turn accuracy, the first correct turn, and the first turn at which the final prediction appears. Static-trace measures are not applied to episodes.

section{Results}
We evaluate 27 datasets---21 static, five interactive, and one passive---on 50 seeded records each (Appendix~\ref{app:benchmark_inventory}). The eight LLMs are GPT-5.6-Luna, Gemini-3.8-Flash, Gemma-4-31B, and Qwen-3.5-27B, queried through an API, and Gemma-4-E4B/E2B and Qwen-3.5-4B/2B, served locally with vLLM; the Jev decision system answers the single-choice tasks under IO only. Static tasks are run under IO and CoT, with three samples per record at temperature 0.7. Interactive and passive tasks are run once per record under IO, with native reasoning disabled (minimal for Gemini-3.8-Flash, which cannot disable it) and the full interaction history in every turn, up to 14 turns for MedQDx, six for AthenaBench, and 20 otherwise; gpt-4o-mini plays the patient or examiner in MedQDx, Med-Inquire, and VivaBench, DDXPlus answers from each patient's recorded questionnaire by BM25 retrieval, and CloudOpsBench replays its recorded tool outputs. Free-form answers and all trace and step-relevance measures are judged by gpt-oss-120b (Appendix~\ref{app:evaluation_details}). The Qwen 3.5 family, particularly its smaller models, has incomplete CoT runs on most datasets due to weak instruction following, so scores are based on completed samples.

\makeatletter
\providecommand{\abRegime}[1]{\rlap{\hspace{-\tabcolsep}\vrule width 0.5pt height \ht\@arstrutbox depth \dp\@arstrutbox}#1}
% The line in front of the Avg. column: drawn over the finished table box, at
% the column boundary, from the top rule to the bottom rule -- one unbroken
% line (a | in the column spec is cut by every booktabs rule and \addlinespace).
\@ifundefined{abTabBox}{\newsavebox{\abTabBox}\newlength{\abAvgW}\newlength{\abAvgD}}{}
\providecommand{\abMetric}[1]{\rlap{\hspace{-\tabcolsep}\lower\dp\@arstrutbox\vbox to \dimexpr\ht\@arstrutbox+\dp\@arstrutbox\relax{\cleaders\vbox to 1.2pt{\vss\hbox{\vrule width 0.55pt height 0.55pt}}\vfill}}#1}
\makeatother
\begin{table}[t]
\centering
\scriptsize
\renewcommand{\arraystretch}{0.82}
\setlength{\tabcolsep}{3pt}
\caption{Main results (\%) by delivery mode, task format, and selection regime. Static tasks: 50 records, three samples per record at temperature 0.7; $\pm$ is the standard deviation across the three samples. Interactive and passive tasks: 50 records, one IO episode each. SCS/MCS = single/multiple-choice selection, Gen = generation; ACC = accuracy, EM = exact match, F1 = set F1, LLM-J = LLM-judge verdict, JRA = joint root-cause accuracy. Bold = best in row. Avg./Average = row/column mean (F1 rows excluded from Average). Jev (non-LLM) is run on SCS tasks only.}
\label{tab:main_results_all_models}
\resizebox{\linewidth}{!}{%
\sbox{\abTabBox}{%
\begin{tabular}{l l l ccccccccccccccccc @{\hspace{3\tabcolsep}} c}
\toprule
\textbf{Benchmark / Dataset} & \textbf{Regime} & \textbf{Metric} & \multicolumn{2}{c}{\textbf{GPT-5.6-Luna}} & \multicolumn{2}{c}{\textbf{Gemini-3.8-Flash}} & \multicolumn{2}{c}{\textbf{Gemma-4-31B}} & \multicolumn{2}{c}{\textbf{Qwen-3.5-27B}} & \multicolumn{2}{c}{\textbf{Gemma-4-E4B}} & \multicolumn{2}{c}{\textbf{Gemma-4-E2B}} & \multicolumn{2}{c}{\textbf{Qwen-3.5-4B}} & \multicolumn{2}{c}{\textbf{Qwen-3.5-2B}} & \multicolumn{1}{c@{\hspace{3\tabcolsep}}}{\textbf{Jev}} & \textbf{Avg.} \\
\cmidrule(lr){4-5} \cmidrule(lr){6-7} \cmidrule(lr){8-9} \cmidrule(lr){10-11} \cmidrule(lr){12-13} \cmidrule(lr){14-15} \cmidrule(lr){16-17} \cmidrule(lr){18-19} \cmidrule(lr){20-20}
& & & \textbf{IO} & \textbf{CoT} & \textbf{IO} & \textbf{CoT} & \textbf{IO} & \textbf{CoT} & \textbf{IO} & \textbf{CoT} & \textbf{IO} & \textbf{CoT} & \textbf{IO} & \textbf{CoT} & \textbf{IO} & \textbf{CoT} & \textbf{IO} & \textbf{CoT} & \textbf{IO} & \\
\midrule
\multicolumn{21}{l}{\textit{\textbf{1. Static Delivery}}} \\
\midrule
\multicolumn{21}{l}{\textit{1a. Selection}} \\
\texttt{art} & \abRegime{MCS} & \abMetric{EM} & 86.7$_{\pm1.2}$ & 89.3$_{\pm1.2}$ & \textbf{92.7}$_{\pm1.2}$ & \textbf{92.7}$_{\pm1.2}$ & 88.7$_{\pm1.2}$ & 91.3$_{\pm1.2}$ & 88.0$_{\pm0.0}$ & 88.7$_{\pm2.3}$ & 72.7$_{\pm3.1}$ & 72.0$_{\pm4.0}$ & 56.7$_{\pm1.2}$ & 48.0$_{\pm0.0}$ & 74.0$_{\pm4.0}$ & 82.7$_{\pm6.4}$ & 54.0$_{\pm4.0}$ & 21.3$_{\pm3.1}$ & -- & 75.0 \\
 & \abRegime{} & \abMetric{F1} & 89.8$_{\pm1.0}$ & 91.6$_{\pm1.9}$ & \textbf{92.7}$_{\pm1.2}$ & \textbf{92.7}$_{\pm1.2}$ & 91.8$_{\pm1.0}$ & 92.2$_{\pm1.0}$ & 89.3$_{\pm0.0}$ & 89.6$_{\pm0.8}$ & 82.4$_{\pm2.5}$ & 86.7$_{\pm2.3}$ & 68.7$_{\pm1.8}$ & 74.7$_{\pm2.3}$ & 74.4$_{\pm4.7}$ & 85.8$_{\pm4.8}$ & 64.2$_{\pm3.4}$ & 57.8$_{\pm1.0}$ & -- & 82.8 \\
 & \abRegime{SCS} & ACC & 88.0$_{\pm0.0}$ & 93.3$_{\pm1.2}$ & \textbf{94.0}$_{\pm2.0}$ & 93.3$_{\pm1.2}$ & 91.3$_{\pm1.2}$ & 92.0$_{\pm0.0}$ & 90.0$_{\pm0.0}$ & 92.0$_{\pm2.0}$ & 82.0$_{\pm2.0}$ & 88.0$_{\pm4.0}$ & 70.0$_{\pm0.0}$ & 78.0$_{\pm6.0}$ & 71.3$_{\pm6.4}$ & 87.3$_{\pm1.2}$ & 65.3$_{\pm4.6}$ & 74.7$_{\pm1.2}$ & 86.7$_{\pm1.2}$ & 84.5 \\
 & \abRegime{Gen} & LLM-J & 43.3$_{\pm3.1}$ & 50.7$_{\pm6.1}$ & 49.0$_{\pm4.7}$ & 44.7$_{\pm5.0}$ & 51.3$_{\pm1.2}$ & 42.7$_{\pm3.1}$ & 44.7$_{\pm1.2}$ & \textbf{52.7}$_{\pm8.3}$ & 32.0$_{\pm2.0}$ & 39.3$_{\pm2.3}$ & 26.7$_{\pm1.2}$ & 37.3$_{\pm4.6}$ & 32.0$_{\pm3.5}$ & 41.3$_{\pm6.4}$ & 14.7$_{\pm3.1}$ & 8.0$_{\pm3.5}$ & -- & 38.2 \\
\addlinespace[1.2pt]
\texttt{ecare} & \abRegime{MCS} & \abMetric{EM} & 82.7$_{\pm2.3}$ & 79.3$_{\pm3.1}$ & 88.7$_{\pm1.2}$ & 88.0$_{\pm0.0}$ & 86.0$_{\pm0.0}$ & 84.7$_{\pm3.1}$ & \textbf{91.3}$_{\pm1.2}$ & 86.0$_{\pm0.0}$ & 76.0$_{\pm3.5}$ & 75.3$_{\pm2.3}$ & 67.3$_{\pm2.3}$ & 73.3$_{\pm3.1}$ & 73.3$_{\pm1.2}$ & 76.7$_{\pm3.1}$ & 54.0$_{\pm7.2}$ & 13.3$_{\pm3.1}$ & -- & 74.7 \\
 & \abRegime{} & \abMetric{F1} & 84.9$_{\pm3.1}$ & 81.6$_{\pm1.7}$ & 88.7$_{\pm1.2}$ & 88.0$_{\pm0.0}$ & 86.0$_{\pm0.0}$ & 86.4$_{\pm3.0}$ & \textbf{91.3}$_{\pm1.2}$ & 86.9$_{\pm0.8}$ & 85.3$_{\pm1.2}$ & 84.2$_{\pm2.8}$ & 87.8$_{\pm0.8}$ & 84.0$_{\pm3.1}$ & 75.6$_{\pm2.3}$ & 81.6$_{\pm1.4}$ & 66.9$_{\pm5.2}$ & 59.6$_{\pm3.9}$ & -- & 82.4 \\
 & \abRegime{SCS} & ACC & 78.7$_{\pm1.2}$ & 84.0$_{\pm0.0}$ & \textbf{90.0}$_{\pm0.0}$ & 88.0$_{\pm0.0}$ & 89.3$_{\pm1.2}$ & 85.3$_{\pm2.3}$ & 85.3$_{\pm1.2}$ & 87.3$_{\pm1.2}$ & 85.3$_{\pm2.3}$ & 81.3$_{\pm3.1}$ & 84.7$_{\pm2.3}$ & 81.3$_{\pm4.2}$ & 74.0$_{\pm3.5}$ & 80.7$_{\pm4.6}$ & 75.3$_{\pm6.1}$ & 69.3$_{\pm5.0}$ & 88.0$_{\pm0.0}$ & 82.8 \\
 & \abRegime{Gen} & LLM-J & 60.7$_{\pm4.2}$ & 64.0$_{\pm3.5}$ & 62.7$_{\pm1.2}$ & \textbf{68.7}$_{\pm3.1}$ & 45.3$_{\pm4.2}$ & 42.7$_{\pm3.1}$ & 60.7$_{\pm1.2}$ & 52.7$_{\pm4.2}$ & 28.7$_{\pm8.1}$ & 30.7$_{\pm6.4}$ & 34.7$_{\pm8.1}$ & 24.7$_{\pm1.2}$ & 50.7$_{\pm5.0}$ & 52.0$_{\pm2.0}$ & 42.0$_{\pm5.3}$ & 39.3$_{\pm1.2}$ & -- & 47.5 \\
\addlinespace[1.2pt]
\texttt{climate\_fever} & \abRegime{MCS} & \abMetric{EM} & 55.3$_{\pm2.3}$ & 54.7$_{\pm1.2}$ & 52.0$_{\pm1.8}$ & 57.3$_{\pm1.2}$ & 62.7$_{\pm1.2}$ & 60.7$_{\pm3.1}$ & \textbf{64.0}$_{\pm0.0}$ & \textbf{64.0}$_{\pm2.0}$ & 56.7$_{\pm1.2}$ & 59.3$_{\pm2.3}$ & 54.0$_{\pm0.0}$ & 62.0$_{\pm4.0}$ & 62.7$_{\pm2.3}$ & 59.3$_{\pm2.3}$ & 43.3$_{\pm5.0}$ & 29.3$_{\pm4.2}$ & -- & 56.1 \\
 & \abRegime{} & \abMetric{F1} & 64.8$_{\pm0.8}$ & 56.0$_{\pm0.7}$ & 52.0$_{\pm1.8}$ & 57.3$_{\pm1.2}$ & 62.7$_{\pm1.2}$ & 60.7$_{\pm3.1}$ & 65.3$_{\pm0.0}$ & 64.0$_{\pm2.0}$ & \textbf{67.8}$_{\pm0.4}$ & 62.8$_{\pm1.6}$ & 59.3$_{\pm1.3}$ & 62.8$_{\pm3.7}$ & 65.3$_{\pm1.3}$ & 64.0$_{\pm2.4}$ & 46.3$_{\pm4.9}$ & 50.0$_{\pm3.3}$ & -- & 60.1 \\
 & \abRegime{SCS} & ACC & 64.0$_{\pm2.0}$ & 51.3$_{\pm2.3}$ & 55.3$_{\pm3.1}$ & 56.7$_{\pm1.2}$ & 62.0$_{\pm2.0}$ & 60.7$_{\pm1.2}$ & 67.3$_{\pm1.2}$ & 61.3$_{\pm2.3}$ & \textbf{70.0}$_{\pm2.0}$ & 64.0$_{\pm2.0}$ & 62.7$_{\pm2.3}$ & 58.0$_{\pm2.0}$ & 64.0$_{\pm2.0}$ & 67.3$_{\pm3.1}$ & 39.3$_{\pm6.1}$ & 64.0$_{\pm2.0}$ & 62.0$_{\pm0.0}$ & 60.6 \\
\addlinespace[1.2pt]
\texttt{true\_detective} & \abRegime{MCS} & \abMetric{EM} & 40.0$_{\pm3.5}$ & 58.0$_{\pm3.5}$ & 73.3$_{\pm1.2}$ & \textbf{74.0}$_{\pm2.0}$ & 50.0$_{\pm4.0}$ & 53.3$_{\pm1.2}$ & 40.7$_{\pm1.2}$ & 55.3$_{\pm8.1}$ & 14.7$_{\pm3.1}$ & 36.0$_{\pm3.5}$ & 16.0$_{\pm2.0}$ & 21.3$_{\pm1.2}$ & 24.7$_{\pm4.6}$ & 25.3$_{\pm1.2}$ & 2.7$_{\pm1.2}$ & 0.7$_{\pm1.2}$ & -- & 36.6 \\
 & \abRegime{} & \abMetric{F1} & 42.8$_{\pm0.3}$ & 59.2$_{\pm3.6}$ & 73.3$_{\pm1.2}$ & \textbf{74.0}$_{\pm2.0}$ & 50.0$_{\pm4.0}$ & 53.6$_{\pm1.4}$ & 40.7$_{\pm1.2}$ & 56.2$_{\pm8.7}$ & 29.4$_{\pm4.8}$ & 38.9$_{\pm4.0}$ & 22.6$_{\pm0.8}$ & 23.1$_{\pm0.3}$ & 32.0$_{\pm4.7}$ & 27.5$_{\pm0.6}$ & 34.1$_{\pm3.1}$ & 3.4$_{\pm2.1}$ & -- & 41.3 \\
 & \abRegime{SCS} & ACC & 38.7$_{\pm4.2}$ & 64.0$_{\pm5.3}$ & \textbf{73.3}$_{\pm1.2}$ & \textbf{73.3}$_{\pm5.0}$ & 51.3$_{\pm1.2}$ & 54.0$_{\pm0.0}$ & 42.0$_{\pm0.0}$ & 56.1$_{\pm5.4}$ & 28.7$_{\pm3.1}$ & 36.0$_{\pm2.0}$ & 19.3$_{\pm1.2}$ & 24.7$_{\pm5.8}$ & 30.0$_{\pm4.0}$ & 31.3$_{\pm1.2}$ & 30.0$_{\pm6.9}$ & 0.7$_{\pm1.2}$ & 36.0$_{\pm2.0}$ & 40.6 \\
\addlinespace[1.2pt]
\texttt{diagnosisarena} & \abRegime{MCS} & \abMetric{EM} & 32.0$_{\pm2.0}$ & 27.3$_{\pm2.3}$ & 39.2$_{\pm2.7}$ & 52.0$_{\pm3.5}$ & \textbf{54.0}$_{\pm3.5}$ & 29.3$_{\pm1.2}$ & 20.0$_{\pm0.0}$ & 52.3$_{\pm4.6}$ & 6.7$_{\pm1.2}$ & 19.3$_{\pm1.2}$ & 16.0$_{\pm0.0}$ & 14.0$_{\pm3.5}$ & 15.3$_{\pm4.2}$ & 33.3$_{\pm5.0}$ & 13.3$_{\pm5.0}$ & 7.3$_{\pm6.1}$ & -- & 27.0 \\
 & \abRegime{} & \abMetric{F1} & 60.1$_{\pm0.5}$ & 60.0$_{\pm1.6}$ & 61.8$_{\pm2.5}$ & \textbf{67.0}$_{\pm2.0}$ & 59.2$_{\pm2.8}$ & 57.0$_{\pm1.6}$ & 47.9$_{\pm1.2}$ & 63.2$_{\pm3.4}$ & 41.3$_{\pm2.2}$ & 47.8$_{\pm2.6}$ & 43.6$_{\pm0.4}$ & 43.8$_{\pm3.6}$ & 45.1$_{\pm2.2}$ & 52.6$_{\pm4.0}$ & 48.1$_{\pm3.8}$ & 32.4$_{\pm4.0}$ & -- & 51.9 \\
 & \abRegime{SCS} & ACC & 65.3$_{\pm3.1}$ & \textbf{74.7}$_{\pm1.2}$ & 73.3$_{\pm1.2}$ & 71.3$_{\pm2.3}$ & 59.3$_{\pm1.2}$ & 57.3$_{\pm5.0}$ & 56.0$_{\pm0.0}$ & 59.3$_{\pm3.1}$ & 43.3$_{\pm2.3}$ & 48.7$_{\pm6.4}$ & 37.3$_{\pm1.2}$ & 42.0$_{\pm5.3}$ & 43.3$_{\pm5.0}$ & 47.3$_{\pm1.2}$ & 46.0$_{\pm7.2}$ & 45.3$_{\pm5.0}$ & 50.7$_{\pm1.2}$ & 54.1 \\
\addlinespace[1.2pt]
\texttt{agentrx} & \abRegime{MCS} & \abMetric{EM} & 48.0$_{\pm2.0}$ & 48.7$_{\pm1.2}$ & \textbf{58.0}$_{\pm2.0}$ & 54.0$_{\pm0.0}$ & 47.3$_{\pm2.3}$ & 46.7$_{\pm1.2}$ & 48.0$_{\pm0.0}$ & 53.3$_{\pm1.2}$ & 41.3$_{\pm1.2}$ & 32.0$_{\pm2.0}$ & 34.0$_{\pm2.0}$ & 18.7$_{\pm5.0}$ & 37.3$_{\pm4.2}$ & 42.7$_{\pm2.3}$ & 10.0$_{\pm3.5}$ & 11.3$_{\pm2.3}$ & -- & 39.5 \\
 & \abRegime{} & \abMetric{F1} & 52.9$_{\pm3.7}$ & 49.6$_{\pm1.7}$ & \textbf{61.6}$_{\pm1.4}$ & 56.3$_{\pm0.6}$ & 50.0$_{\pm2.3}$ & 56.1$_{\pm3.8}$ & 50.7$_{\pm0.0}$ & 54.7$_{\pm0.7}$ & 46.7$_{\pm1.8}$ & 47.8$_{\pm2.0}$ & 50.9$_{\pm0.7}$ & 47.8$_{\pm2.6}$ & 44.4$_{\pm3.0}$ & 49.4$_{\pm1.4}$ & 31.0$_{\pm4.1}$ & 26.6$_{\pm0.8}$ & -- & 48.5 \\
 & \abRegime{SCS} & ACC & 53.3$_{\pm2.3}$ & 50.0$_{\pm2.0}$ & \textbf{61.3}$_{\pm2.3}$ & 60.0$_{\pm2.0}$ & 52.7$_{\pm1.2}$ & 54.0$_{\pm4.0}$ & 50.7$_{\pm2.3}$ & 50.0$_{\pm2.0}$ & 46.7$_{\pm2.3}$ & 42.7$_{\pm3.1}$ & 46.0$_{\pm0.0}$ & 54.7$_{\pm2.3}$ & 42.0$_{\pm5.3}$ & 47.3$_{\pm4.6}$ & 34.0$_{\pm2.0}$ & 48.7$_{\pm7.6}$ & 48.7$_{\pm1.2}$ & 49.6 \\
\addlinespace[1.2pt]
\texttt{aer} & \abRegime{MCS} & \abMetric{EM} & 56.7$_{\pm1.2}$ & 64.0$_{\pm6.0}$ & 65.3$_{\pm2.3}$ & 66.7$_{\pm3.1}$ & \textbf{73.3}$_{\pm3.1}$ & 66.7$_{\pm1.2}$ & 49.3$_{\pm1.2}$ & 62.2$_{\pm1.0}$ & 49.3$_{\pm1.2}$ & 58.0$_{\pm0.0}$ & 38.0$_{\pm2.0}$ & 43.3$_{\pm9.2}$ & 52.7$_{\pm4.2}$ & 58.7$_{\pm1.2}$ & 37.3$_{\pm5.0}$ & 3.3$_{\pm1.2}$ & -- & 52.8 \\
 & \abRegime{} & \abMetric{F1} & 80.3$_{\pm1.4}$ & 84.2$_{\pm3.4}$ & 85.1$_{\pm0.8}$ & 85.7$_{\pm0.7}$ & \textbf{88.4}$_{\pm1.1}$ & 87.3$_{\pm0.9}$ & 77.6$_{\pm0.5}$ & 83.8$_{\pm1.5}$ & 71.6$_{\pm1.5}$ & 78.7$_{\pm0.7}$ & 65.6$_{\pm2.5}$ & 68.2$_{\pm5.9}$ & 76.3$_{\pm1.4}$ & 82.0$_{\pm1.8}$ & 67.4$_{\pm2.1}$ & 31.5$_{\pm5.7}$ & -- & 75.9 \\
\addlinespace[1.2pt]
\multicolumn{21}{l}{\textit{1b. Generation}} \\
\texttt{RCA100} & Gen & EM & 28.7$_{\pm1.2}$ & 15.3$_{\pm5.0}$ & 24.0$_{\pm0.0}$ & 22.0$_{\pm2.0}$ & 31.3$_{\pm1.2}$ & 24.7$_{\pm5.0}$ & \textbf{32.7}$_{\pm1.2}$ & 24.7$_{\pm1.2}$ & 26.0$_{\pm6.0}$ & 26.7$_{\pm4.2}$ & 28.0$_{\pm0.0}$ & 28.7$_{\pm4.2}$ & 22.7$_{\pm2.3}$ & 24.0$_{\pm4.0}$ & 14.7$_{\pm4.6}$ & 20.0$_{\pm3.5}$ & -- & 24.6 \\
\texttt{commonwhy} & Gen & LLM-J & 62.0$_{\pm5.3}$ & 70.7$_{\pm1.2}$ & \textbf{83.3}$_{\pm1.2}$ & 81.3$_{\pm2.3}$ & 50.0$_{\pm0.0}$ & 54.7$_{\pm2.3}$ & 59.3$_{\pm1.2}$ & 53.3$_{\pm6.1}$ & 38.7$_{\pm8.1}$ & 46.0$_{\pm10.0}$ & 28.0$_{\pm3.5}$ & 28.7$_{\pm5.0}$ & 35.3$_{\pm6.1}$ & 32.7$_{\pm4.6}$ & 23.3$_{\pm4.2}$ & 18.7$_{\pm2.3}$ & -- & 47.9 \\
\texttt{crosstrace} & Gen & LLM-J & 62.0$_{\pm6.0}$ & \textbf{66.7}$_{\pm1.2}$ & 56.4$_{\pm4.2}$ & 56.7$_{\pm4.2}$ & 61.3$_{\pm2.3}$ & 53.3$_{\pm1.2}$ & 53.3$_{\pm3.1}$ & 34.0$_{\pm4.0}$ & 38.7$_{\pm5.0}$ & 32.0$_{\pm2.0}$ & 44.0$_{\pm0.0}$ & 32.7$_{\pm9.0}$ & 42.0$_{\pm5.3}$ & 42.7$_{\pm2.3}$ & 19.3$_{\pm6.4}$ & 13.3$_{\pm5.0}$ & -- & 44.3 \\
\texttt{enwn\_entailmentbank} & Gen & LLM-J & 86.0$_{\pm2.0}$ & \textbf{95.3}$_{\pm1.2}$ & 90.5$_{\pm2.6}$ & 88.0$_{\pm0.0}$ & 94.0$_{\pm2.0}$ & 90.0$_{\pm2.0}$ & 83.3$_{\pm1.2}$ & 83.8$_{\pm3.6}$ & 59.3$_{\pm5.8}$ & 73.3$_{\pm3.1}$ & 59.3$_{\pm2.3}$ & 80.0$_{\pm3.5}$ & 65.3$_{\pm4.2}$ & 86.7$_{\pm1.2}$ & 52.7$_{\pm7.0}$ & 74.7$_{\pm4.6}$ & -- & 78.9 \\
\texttt{house\_md} & Gen & LLM-J & 38.0$_{\pm2.0}$ & 42.7$_{\pm1.2}$ & 43.3$_{\pm4.2}$ & \textbf{54.7}$_{\pm3.1}$ & 40.7$_{\pm3.1}$ & 33.3$_{\pm5.0}$ & 26.0$_{\pm0.0}$ & 40.7$_{\pm5.1}$ & 15.3$_{\pm3.1}$ & 19.3$_{\pm2.3}$ & 10.0$_{\pm0.0}$ & 14.0$_{\pm2.0}$ & 11.3$_{\pm1.2}$ & 21.3$_{\pm5.0}$ & 5.3$_{\pm2.3}$ & 4.0$_{\pm2.0}$ & -- & 26.2 \\
\texttt{hypogen} & Gen & LLM-J & \textbf{78.7}$_{\pm1.2}$ & 71.3$_{\pm1.2}$ & 72.7$_{\pm1.2}$ & 74.7$_{\pm1.2}$ & 71.3$_{\pm2.3}$ & 71.3$_{\pm2.3}$ & 68.0$_{\pm5.3}$ & 54.1$_{\pm6.7}$ & 58.7$_{\pm2.3}$ & 54.0$_{\pm4.0}$ & 50.0$_{\pm4.0}$ & 51.3$_{\pm4.6}$ & 53.3$_{\pm3.1}$ & 54.0$_{\pm6.0}$ & 38.7$_{\pm6.4}$ & 29.3$_{\pm4.2}$ & -- & 59.5 \\
\texttt{llm4biohypogen} & Gen & LLM-J & 70.0$_{\pm0.0}$ & 64.0$_{\pm6.0}$ & 67.1$_{\pm5.3}$ & 66.0$_{\pm3.5}$ & 62.0$_{\pm2.0}$ & 62.7$_{\pm3.1}$ & \textbf{71.3}$_{\pm1.2}$ & 68.2$_{\pm20.2}$ & 63.3$_{\pm2.3}$ & 65.3$_{\pm4.2}$ & 56.7$_{\pm3.1}$ & 60.0$_{\pm2.0}$ & 62.0$_{\pm4.0}$ & 67.3$_{\pm3.1}$ & 36.7$_{\pm1.2}$ & 52.7$_{\pm3.1}$ & -- & 62.2 \\
\texttt{matter\_to\_mechanism} & Gen & LLM-J & \textbf{76.7}$_{\pm7.0}$ & 64.7$_{\pm1.2}$ & 59.3$_{\pm4.2}$ & 60.7$_{\pm1.2}$ & 69.3$_{\pm1.2}$ & 53.3$_{\pm6.1}$ & 57.3$_{\pm4.2}$ & 57.9$_{\pm15.8}$ & 47.3$_{\pm10.1}$ & 40.7$_{\pm4.2}$ & 41.3$_{\pm3.1}$ & 45.3$_{\pm8.1}$ & 48.7$_{\pm1.2}$ & 42.7$_{\pm2.3}$ & 23.3$_{\pm5.8}$ & 12.0$_{\pm5.3}$ & -- & 50.0 \\
\texttt{medcasereasoning} & Gen & LLM-J & 62.0$_{\pm4.0}$ & \textbf{66.0}$_{\pm4.0}$ & 54.4$_{\pm3.5}$ & 60.7$_{\pm1.2}$ & 49.3$_{\pm1.2}$ & 54.0$_{\pm2.0}$ & 42.0$_{\pm0.0}$ & 37.5$_{\pm24.9}$ & 24.7$_{\pm5.0}$ & 25.3$_{\pm3.1}$ & 14.7$_{\pm1.2}$ & 18.7$_{\pm2.3}$ & 23.3$_{\pm3.1}$ & 42.7$_{\pm1.2}$ & 8.0$_{\pm3.5}$ & 20.7$_{\pm1.2}$ & -- & 37.8 \\
\texttt{medr\_bench} & Gen & LLM-J & 87.3$_{\pm1.2}$ & \textbf{90.0}$_{\pm2.0}$ & 88.0$_{\pm2.0}$ & 82.7$_{\pm1.2}$ & 69.3$_{\pm1.2}$ & 79.3$_{\pm4.2}$ & 78.0$_{\pm0.0}$ & 77.3$_{\pm24.0}$ & 54.0$_{\pm4.0}$ & 67.3$_{\pm1.2}$ & 56.7$_{\pm1.2}$ & 64.0$_{\pm3.5}$ & 56.0$_{\pm0.0}$ & 67.3$_{\pm6.4}$ & 29.3$_{\pm7.6}$ & 39.3$_{\pm7.6}$ & -- & 67.9 \\
\texttt{neulr} & Gen & LLM-J & 58.7$_{\pm6.1}$ & 64.7$_{\pm4.2}$ & 68.7$_{\pm3.1}$ & 66.0$_{\pm0.0}$ & 54.0$_{\pm0.0}$ & \textbf{70.7}$_{\pm6.4}$ & 52.7$_{\pm1.2}$ & 66.7$_{\pm6.0}$ & 30.0$_{\pm0.0}$ & 64.0$_{\pm3.5}$ & 28.0$_{\pm2.0}$ & 62.0$_{\pm7.2}$ & 13.3$_{\pm6.1}$ & 68.0$_{\pm2.0}$ & 3.3$_{\pm3.1}$ & 37.3$_{\pm5.0}$ & -- & 50.5 \\
\texttt{proof\_writer} & Gen & EM & 37.3$_{\pm2.3}$ & 40.0$_{\pm2.0}$ & 40.0$_{\pm0.0}$ & 44.0$_{\pm0.0}$ & 42.0$_{\pm2.0}$ & 47.3$_{\pm3.1}$ & 34.0$_{\pm0.0}$ & 59.3$_{\pm23.1}$ & 39.3$_{\pm1.2}$ & 17.3$_{\pm2.3}$ & 15.3$_{\pm3.1}$ & \textbf{52.7}$_{\pm2.3}$ & 16.0$_{\pm7.2}$ & 52.0$_{\pm5.3}$ & 7.3$_{\pm6.1}$ & 8.7$_{\pm1.2}$ & -- & 34.5 \\
\texttt{uncommonsense} & Gen & LLM-J & \textbf{78.7}$_{\pm3.1}$ & \textbf{78.7}$_{\pm1.2}$ & 71.3$_{\pm4.2}$ & 70.0$_{\pm2.0}$ & 66.0$_{\pm2.0}$ & 58.0$_{\pm5.3}$ & 74.7$_{\pm5.0}$ & 63.4$_{\pm3.3}$ & 73.3$_{\pm3.1}$ & 68.7$_{\pm4.2}$ & 64.0$_{\pm6.0}$ & 55.3$_{\pm3.1}$ & 48.7$_{\pm6.4}$ & 47.3$_{\pm4.2}$ & 34.7$_{\pm1.2}$ & 18.7$_{\pm4.2}$ & -- & 60.7 \\
\texttt{uniadilr\_hgc} & Gen & EM & 42.7$_{\pm4.6}$ & 33.3$_{\pm2.3}$ & 72.0$_{\pm3.5}$ & \textbf{76.0}$_{\pm2.0}$ & 64.0$_{\pm2.0}$ & 46.7$_{\pm3.1}$ & 14.0$_{\pm0.0}$ & 14.3$_{\pm28.9}$ & 40.0$_{\pm2.0}$ & 2.7$_{\pm1.2}$ & 9.3$_{\pm1.2}$ & 4.0$_{\pm3.5}$ & 13.3$_{\pm4.2}$ & 8.0$_{\pm3.5}$ & 0.7$_{\pm1.2}$ & 3.3$_{\pm2.3}$ & -- & 27.8 \\
\midrule
\multicolumn{21}{l}{\textit{\textbf{2. Interactive Delivery}}} \\
\midrule
\texttt{medqdx} & Gen & LLM-J & \textbf{40.0} & -- & 34.0 & -- & 36.0 & -- & 32.0 & -- & 24.0 & -- & 32.0 & -- & 26.0 & -- & 6.0 & -- & -- & 28.8 \\
\addlinespace[1.2pt]
\texttt{med\_inquire} & Gen & LLM-J & 22.0 & -- & \textbf{44.0} & -- & 28.0 & -- & 20.0 & -- & 12.0 & -- & 8.0 & -- & 6.0 & -- & 0.0 & -- & -- & 17.5 \\
\addlinespace[1.2pt]
\texttt{vivabench} & Gen & LLM-J & 52.0 & -- & \textbf{58.0} & -- & 48.0 & -- & 40.0 & -- & 22.0 & -- & 18.0 & -- & 28.0 & -- & 8.0 & -- & -- & 34.2 \\
\addlinespace[1.2pt]
\texttt{ddxplus} & \abRegime{SCS} & ACC & 32.0 & -- & 22.0 & -- & 26.0 & -- & 42.0 & -- & 36.0 & -- & \textbf{44.0} & -- & 40.0 & -- & 18.0 & -- & -- & 32.5 \\
 & \abRegime{MCS} & ACC & 28.0 & -- & 30.0 & -- & 36.0 & -- & \textbf{38.0} & -- & 34.0 & -- & \textbf{38.0} & -- & 32.0 & -- & 24.0 & -- & -- & 32.5 \\
\addlinespace[1.2pt]
\texttt{cloud\_opsbench} & Gen & JRA & 6.0 & -- & \textbf{10.0} & -- & 4.0 & -- & 2.0 & -- & 0.0 & -- & 0.0 & -- & 0.0 & -- & 0.0 & -- & -- & 2.8 \\
\addlinespace[1.2pt]
\midrule
\multicolumn{21}{l}{\textit{\textbf{3. Passive Delivery}}} \\
\midrule
\texttt{athena\_bench} & Gen & LLM-J & 30.0 & -- & \textbf{44.0} & -- & 26.0 & -- & 20.0 & -- & 20.0 & -- & 8.0 & -- & 10.0 & -- & 0.0 & -- & -- & 19.8 \\
\midrule
\textbf{Average} & & & 54.8 & 62.6 & 60.0 & \textbf{67.0} & 55.4 & 59.3 & 51.1 & 58.9 & 41.4 & 47.8 & 36.6 & 44.1 & 38.9 & 51.4 & 25.4 & 27.2 & 62.0 & \\
\bottomrule
\end{tabular}}%
\settowidth{\abAvgW}{\textbf{Avg.}}\settowidth{\abAvgD}{00.0}\ifdim\abAvgD>\abAvgW\setlength{\abAvgW}{\abAvgD}\fi%
\usebox{\abTabBox}\llap{\rule[\dimexpr-\dp\abTabBox+\heavyrulewidth\relax]{0.4pt}{\dimexpr\ht\abTabBox+\dp\abTabBox-2\heavyrulewidth\relax}\hspace{\dimexpr\tabcolsep+\abAvgW+1.5\tabcolsep-0.2pt\relax}}%
}
\end{table}

% =====================================================================
% SECTION 6 (Results) -- refined, condensed version.
% HOW TO USE: in Sections/6_experiments.tex keep everything from the top
% (\section{Results}, the \abRegime/\abMetric macros and the main table)
% down to and including \end{table}. Replace EVERYTHING after \end{table}
% (from "\label{sec:experiments}" to the end of the file, commented-out
% blocks included) with the text below.
% =====================================================================

\label{sec:experiments}
\label{sec:results}

\graphicspath{{prism-uploads/}{figures/}{./}}

Table~\ref{tab:main_results_all_models} reports final-answer scores for eight LLMs and the Jev decision system; the subsections below explain what drives them and what the reasoning traces and interaction histories add.

\subsection{What determines performance across benchmarks?}
\label{sec:main_results}

\textbf{Task format.} Proposing an explanation is harder than recognizing it: where one dataset is posed both ways, the row average falls from 84.5 (SCS) to 38.2 (Gen) on \texttt{art} and from 82.8 to 47.5 on \texttt{ecare}. Allowing several answers also costs accuracy for every model---10.6 points on average, up to 27.7 for Qwen-3.5-2B (Appendix~\ref{sec:single_vs_multiple_choice}).

\textbf{Chain-of-thought.} CoT helps where the explanation must be derived by chaining premises or findings---\texttt{neulr} (Qwen-3.5-4B 13.3$\to$68.0), \texttt{proof\_writer} (Gemma-4-E2B 15.3$\to$52.7), \texttt{medcasereasoning} (Qwen-3.5-4B 23.3$\to$42.7)---but often hurts open-ended hypothesis generation: \texttt{uniadilr\_hgc} (Gemma-4-E4B 40.0$\to$2.7), \texttt{matter\_to\_mechanism} (GPT-5.6-Luna 76.7$\to$64.7), \texttt{uncommonsense} (Qwen-3.5-2B 34.7$\to$18.7).

\textbf{Delivery mode.} Interactive and passive tasks are the hardest: the best model reaches 58.0 on \texttt{vivabench}, 44.0 on \texttt{athena\_bench}, and only 10.0 on \texttt{cloud\_opsbench}, and four of the eight LLMs score 0.0 on \texttt{cloud\_opsbench}.

\textbf{Models.} Gemini-3.8-Flash has the highest average in both modes (60.0 IO, 67.0 CoT) and the best score on four of the seven interactive and passive rows; the smallest model, Qwen-3.5-2B, is last (25.4 IO, 27.2 CoT). The IO averages include the interactive and passive rows, which have no CoT variant, so IO and CoT averages are not directly comparable.

\subsection{Can a System One decision system perform abductive selection?}
\label{sec:jev}

Jev is TypeSafe AI's closed, hosted System One decision model: it evaluates typed questions against an input state and returns a structured choice and probability distribution rather than generated text \citep{typesafe2026jev}. Its public documentation does not specify the model architecture or training corpus. For each SCS task, we provide the same task input and fixed candidate set used in the IO condition and record Jev's selected candidate; because it does not generate an explanation, we do not run it on generation, interaction, or MCS. On the six single-choice tasks it averages 62.0, above every small LLM under IO (48.3--59.3) but below every large one (64.7--74.5). It is near the top on short commonsense explanation choices---88.0 on \texttt{ecare}, where only two of the sixteen LLM cells are higher, and 86.7 on \texttt{art}---but falls far behind when the candidates must be weighed against many clues in a long case: 36.0 vs.\ 73.3 on \texttt{true\_detective} and 50.7 vs.\ 74.7 on \texttt{diagnosisarena}. A fast decision system thus covers explanation \emph{selection} over compact evidence, but not the evidence integration that longer abductive cases require.
% -------------------------------------------------------------------------
\begin{figure*}[t]
  \centering
  \vspace{-6pt}
  \begin{subfigure}[c]{0.40\linewidth}
    \centering
    \includegraphics[width=\linewidth]{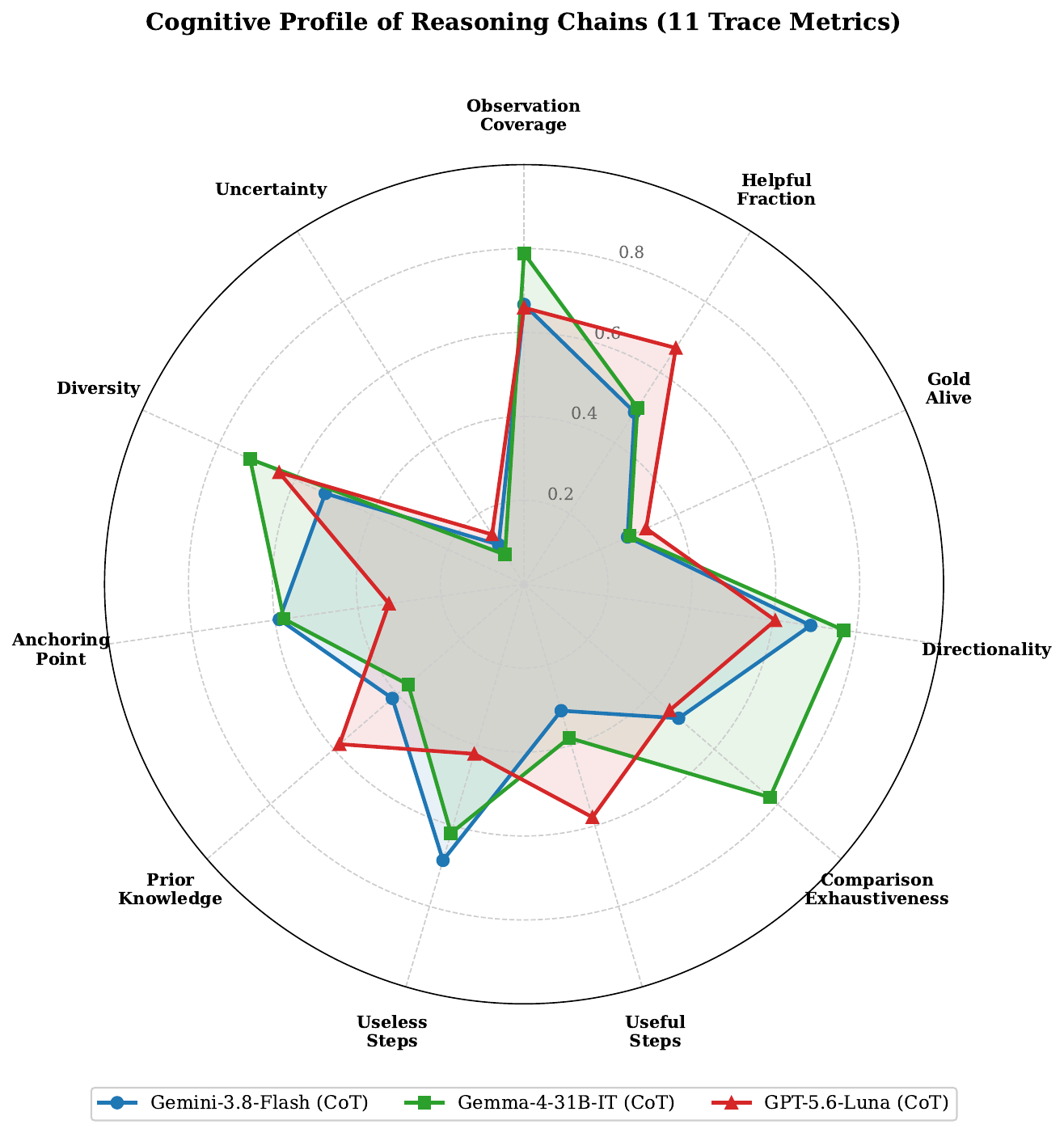}
    \subcaption{}
    \label{fig:cognitive_profile_radar}
  \end{subfigure}%
  \hspace{0.03\linewidth}%
  \begin{subfigure}[c]{0.55\linewidth}
    \centering
    \includegraphics[width=\linewidth]{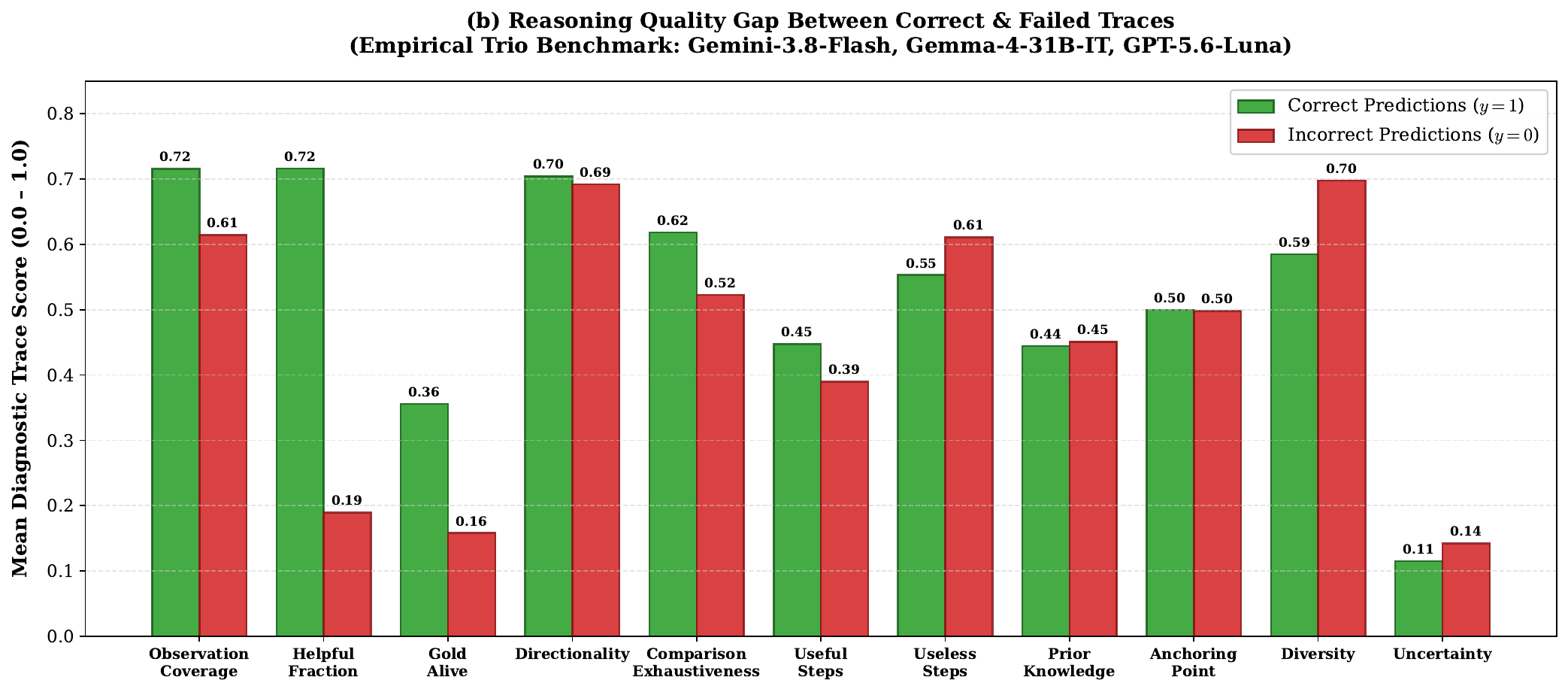}
    \subcaption{}
    \label{fig:quality_gap_histogram}
  \end{subfigure}
  \vspace{2pt}
  \caption{Eleven normalized trace measures (0--1) for the CoT traces of Gemini-3.8-Flash, Gemma-4-31B, and GPT-5.6-Luna. \textbf{(a)} Mean per model. \textbf{(b)} Mean for correct ($y=1$) and incorrect ($y=0$) answers. Helpful Fraction and Gold Alive are judged against the gold answer.}
  \label{fig:cognitive_profile_and_quality_gap}
\end{figure*}

\subsection{Which trace properties accompany a correct explanation?}
\label{sec:cognitive_profiles_and_quality_gap}

Models reach similar accuracy through different traces (Figure~\ref{fig:cognitive_profile_radar}): Gemma-4-31B covers the most observations and compares candidates most exhaustively, whereas GPT-5.6-Luna has the highest share of helpful steps and prior-knowledge use and commits to its answer earliest. Across the three models, correct and incorrect traces differ mainly in whether the correct explanation stays in play (Figure~\ref{fig:quality_gap_histogram}): the helpful-step fraction (0.72 vs.\ 0.19) and the rate at which the gold hypothesis remains under consideration (0.36 vs.\ 0.16) drop sharply in failures, while observation coverage (0.72 vs.\ 0.61) and comparison exhaustiveness (0.62 vs.\ 0.52) differ moderately. The \emph{form} of the trace barely differs---directionality (0.70 vs.\ 0.69), prior knowledge (0.44 vs.\ 0.45), and anchoring point (0.50 vs.\ 0.50)---and failed traces are more diverse (0.70 vs.\ 0.59) and contain more useless steps (0.61 vs.\ 0.55). Because two of the largest gaps are judged against the gold answer, they partly restate correctness; the gold-free gaps (coverage, comparison, diversity) indicate that successful abduction depends on using the evidence to keep and discriminate the right candidates, not on reasoning in an evidence-first style.

% -------------------------------------------------------------------------
\begin{figure}[t]
  \centering
  \begin{subfigure}[b]{0.48\linewidth}
    \centering
    \includegraphics[width=\linewidth]{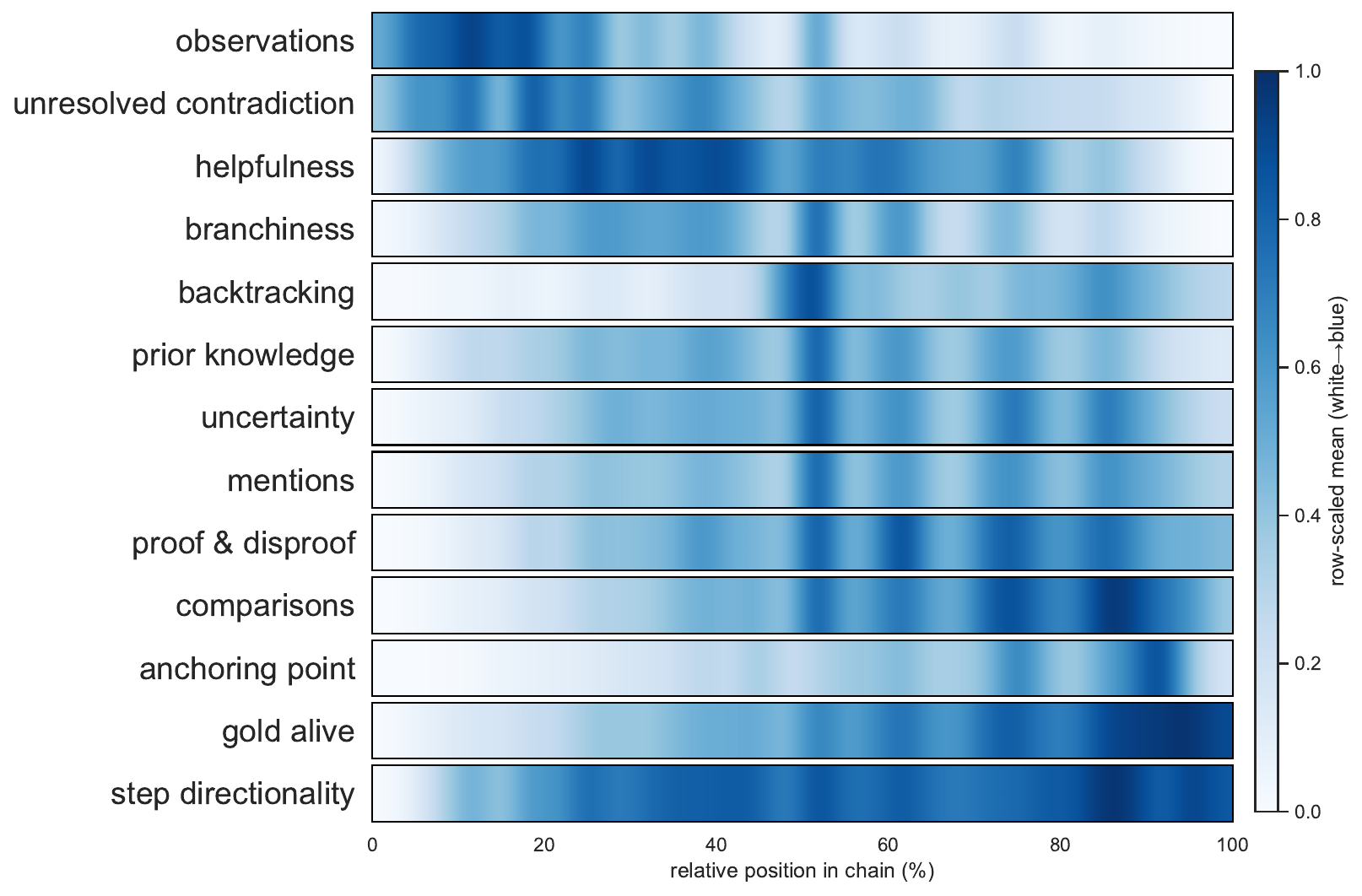}
    \subcaption{}
    \label{fig:position_heatmaps_a}
  \end{subfigure}%
  \hfill
  \begin{subfigure}[b]{0.48\linewidth}
    \centering
    \includegraphics[width=\linewidth]{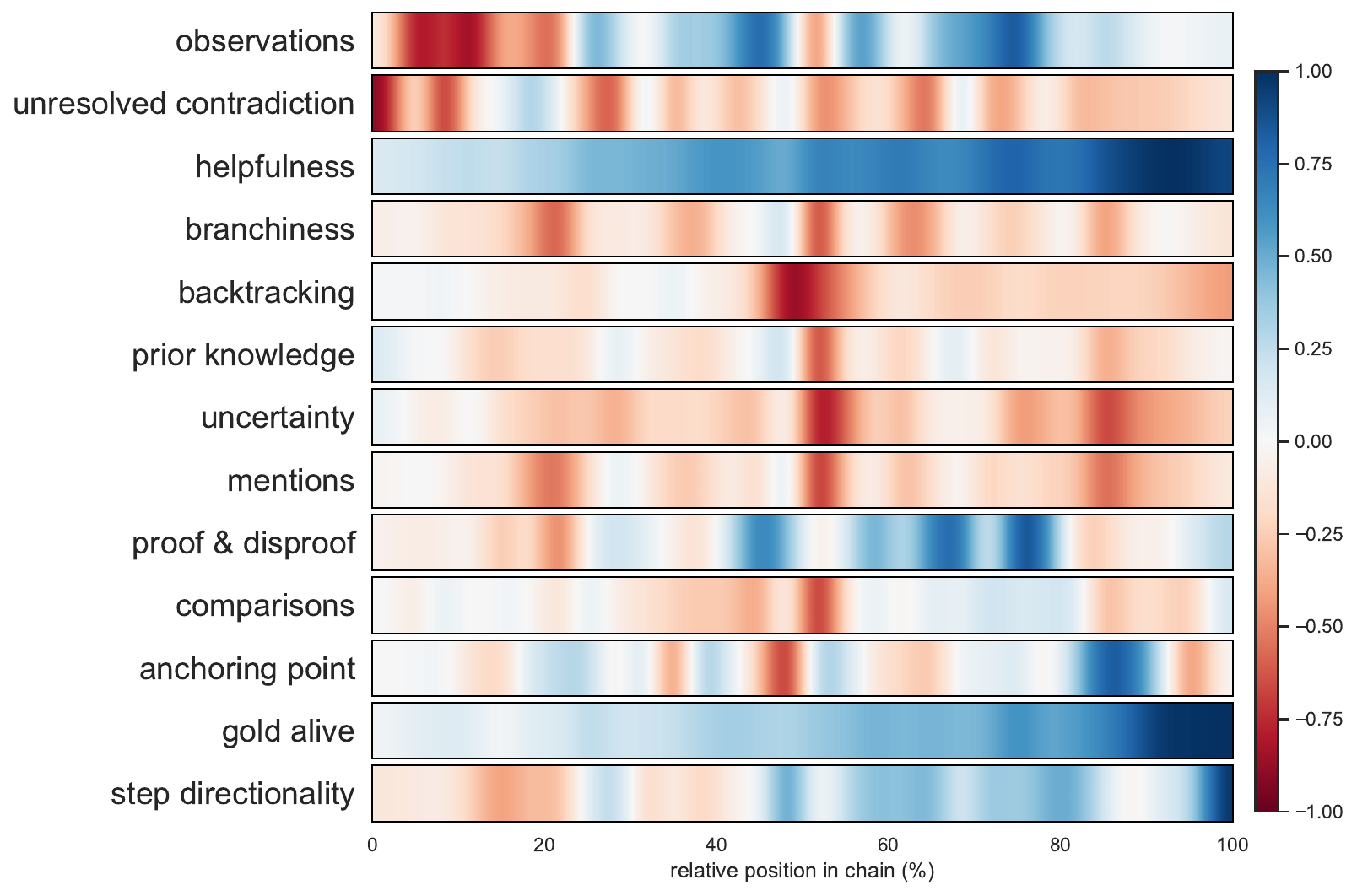}
    \subcaption{}
    \label{fig:position_heatmaps_b}
  \end{subfigure}
  \caption{Judged trace measures by relative position in the chain (0--100\%), pooled over the eight LLMs. (a) Mean, each row min--max scaled. (b) Correct-minus-incorrect difference, computed within each dataset and averaged; each row scaled symmetrically around zero, so blue marks positions where correct chains score higher and red where incorrect chains score higher. Row intensities are not comparable across measures.}
  \label{fig:position_heatmaps}
\end{figure}

\subsection{Where in the chain do correct and incorrect reasoning diverge?}
\label{sec:chain_position}

On average, traces follow the order abduction prescribes (Figure~\ref{fig:position_heatmaps_a}): since abduction proceeds from evidence to hypothesis, observations concentrate in the first fifth of the chain, backtracking and uncertainty peak near the midpoint, and the anchoring point, the gold hypothesis, and step directionality peak in the last fifth---evidence first, commitment last. Splitting by outcome (Figure~\ref{fig:position_heatmaps_b}) shows where failures begin. \emph{Unresolved contradiction} is higher in incorrect chains from the very first steps, before any hypothesis is anchored, making it the earliest trace signal of an eventual error (Appendix~\ref{case:early_contradiction} gives a worked example). Around the midpoint, uncertainty, backtracking, branchiness, and observation mentions are all higher in incorrect chains, so mid-chain hedging and revisiting accompany wrong answers rather than successful self-correction. The gold hypothesis and helpful steps favor correct chains throughout and increasingly toward the end, partly by construction. 

% -------------------------------------------------------------------------
\subsection{Do models keep asking relevant questions as an interaction goes on?}\par\nopagebreak\smallskip

\label{app:relevance_by_step}
\FloatBarrier
\begin{figure}[H]
  \centering
  \begin{subfigure}[t]{0.5\linewidth}
    \centering
    \includegraphics[width=\linewidth]{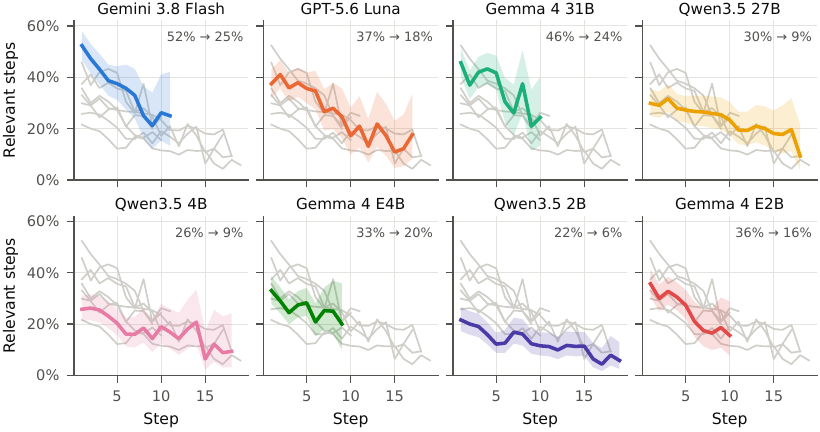}
    \subcaption{} 
    \label{fig:relevance_by_step_model}
  \end{subfigure}%
  \hfill
  \begin{subfigure}[t]{0.49\linewidth}
    \centering
    \includegraphics[width=\linewidth]{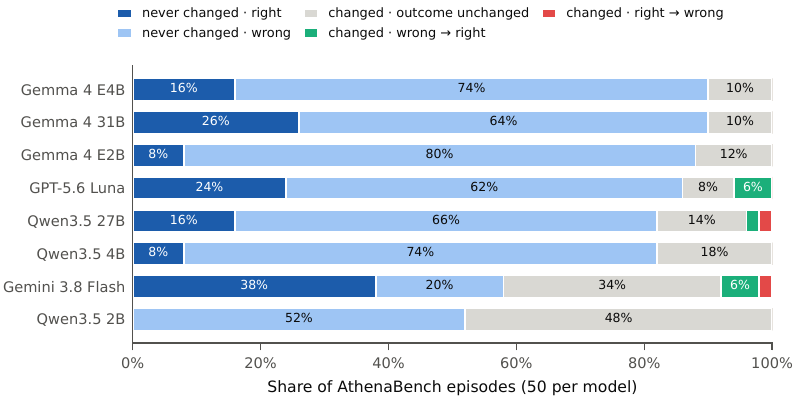}
    \subcaption{}
    \label{fig:athena_revision}
  \end{subfigure}
  \caption{Relevance and revision over the course of an episode. The left panel (a) tracks the share of actions judged relevant to the gold answer at each step of the interactive episodes, one colored panel per model (95\% Wilson bands) against the other seven in gray; a step is plotted only once at least 30 of the model's episodes reach it, so models that answer early trace shorter curves, and corner labels give relevance at the first and last plotted step. MedQDx counts its questions only. The right panel (b) shows, for AthenaBench's passive setting, the share of each model's 50 episodes whose predicted actor never changed, changed without changing the outcome, or flipped between wrong and right from the first turn to the last.}
  \label{fig:relevance_and_revision}
\end{figure}

The share of relevant actions falls as the episode proceeds for every model (Figure~\ref{fig:relevance_by_step_model}): from 52\% to 25\% for Gemini-3.8-Flash, 46\% to 24\% for Gemma-4-31B, 37\% to 18\% for GPT-5.6-Luna, and from 30\% or less to under 10\% for the Qwen-3.5 models. The models also differ in \emph{how long} they keep asking before the decline sets in: Gemini and the Gemma models usually commit within about eight turns, so their curves end early, whereas the Qwen models and GPT-5.6-Luna continue to steps 16--19 with a low relevance rate, spending turns without gathering useful evidence. Later steps come only from longer episodes, which tend to be the harder or less successful ones, so part of the decline reflects which episodes are still running rather than a within-episode drop alone; either way, a longer interview is not a more informative one.

\subsection{Do models revise their hypothesis as passive evidence accumulates?}
\label{sec:athena_revision}

Each AthenaBench case reveals 15--25 evidence items over four to six turns, yet in 78.5\% of the 400 episodes the model names the same actor at every turn (Figure~\ref{fig:athena_revision}). The revisions that do occur mostly fail to help: only 7 episodes move from a wrong first prediction to a correct final one, and 2 move the other way, so final accuracy nearly equals first-turn accuracy (Gemini-3.8-Flash 40\%$\to$44\%, GPT-5.6-Luna 24\%$\to$30\%, unchanged for the other six LLMs). Frequent revision is not sufficient either: Qwen-3.5-2B changes its prediction in 48\% of episodes but never reaches the correct actor. Passive delivery thus exposes what static evaluation cannot---models anchor on the explanation formed from the first evidence and do not update it as the evidence accumulates.
% =====================================================================
% APPENDIX ADDITION (needed: the text above cites
% Appendix~\ref{app:reasoning_length}). Paste into Sections/Appendix/
% appendix.tex, e.g. right after the single- vs. multiple-choice block,
% and upload figS4_accuracy_vs_reasoning_length.pdf to prism-uploads/.
% =====================================================================
%
% \subsubsection{Accuracy and Reasoning Length Across All Eight LLMs}
% \label{app:reasoning_length}
% \begin{figure}[H]
%   \centering
%   \includegraphics[width=0.7\linewidth]{figS4_accuracy_vs_reasoning_length.pdf}
%   \caption{CoT accuracy by reasoning-length quintile, computed within each dataset and model (1 = shortest fifth of replies), with 95\% Wilson bands; dotted: all models pooled.}
%   \label{fig:accuracy_vs_reasoning_length}
% \end{figure}
% Within every model, accuracy declines monotonically from the shortest to the longest fifth of CoT replies, from 68\% to 37\% pooled; Qwen-3.5-2B falls from 40\% to 11\%.

\FloatBarrier

\section{Future Work}
\label{sec:future_work}
\subsection{Limitations}
The present draft has four material limitations. First, although we manually reviewed random samples of judge outputs and revised judge prompts accordingly, we did not carry out a systematic evaluation of judge quality or report complete agreement statistics. Second, the model roster could be extended to strengthen the generality of the reported trends. Third, the scarcity of existing interactive and passive tasks limited how much analysis we could perform on this category; this reflects a broader gap in the field rather than a limitation specific to this benchmark, and we see constructing more such datasets as an important direction for future work. Fourth, due to the token cost of interactive and sequential tasks, we ran each episode only once rather than averaging over multiple runs as we did for static tasks.

\subsection{What Final Answers Do Not Reveal}

A correct final answer does not show whether the model considered the relevant alternatives, used the supplied observations, resolved contradictions, or simply produced a disconnected correct choice. Conversely, an incorrect answer does not distinguish failure to retrieve the correct hypothesis from failure to select a hypothesis that was actively considered. The qualitative cases in Appendix~\ref{app:qualitative_results} exhibit both patterns. Future evaluation should therefore retain outcome scores while separately reporting evidence use, candidate dynamics, anchoring, contradiction resolution, and answer--trace consistency.

\subsection{Implications for Abductive-Reasoning Evaluation}

The current results motivate evaluations that vary candidate-set size, reasoning horizon, context length, and evidence-delivery mode instead of treating abduction as a single accuracy column. A stronger next step is to pair naturally occurring benchmarks with controlled variants that hold content fixed while changing one property at a time. Passive and interactive tasks are particularly important because they test hypothesis revision and evidence acquisition directly, but they also require explicit action budgets, environment versions, and process-aware metrics to remain comparable and reproducible.
\section{Conclusion}
\label{sec:conclusion}

AbductionBench organizes abductive evaluation around what models must do with hypotheses, where tasks are applied, and how evidence becomes available, combining final-answer scoring with analysis of explicit reasoning traces, and extending this framework to passive and interactive settings when the required records are available. The current experiments show that CoT is not uniformly beneficial, that accuracy depends on task format and delivery mode, that a fast decision system can match or exceed LLMs on selection over compact evidence but falls behind on longer cases, that reasoning traces follow an abductive flow—evidence first, commitment last—and that models with similar outcomes expose substantially different reasoning profiles. These findings support evaluating abductive systems through both their conclusions and the evidence-sensitive process they present, while preserving clear limits on what written traces can establish about internal reasoning.

\subsection*{AI use statement}

In this work, LLMs are the object of study: the evaluated models, the answer and reasoning-trace judge (gpt-oss-120b), and the simulated patients and examiners (gpt-4o-mini) are described in Section~\ref{sec:results} and Appendix~\ref{app:evaluation_details}. Beyond this, we used generative AI coding assistants to help implement the evaluation framework, dataset adapters, and analysis and plotting scripts, and to assist with editing the text of this paper. We did not use generative AI tools to generate research ideas or to fabricate or alter data or results. All AI-assisted code was reviewed and tested by the authors, all reported numbers are computed by our pipeline from the stored model outputs, and all AI-assisted text was reviewed and revised by the authors. We take responsibility for the final content of this work, including text, claims, and artifacts produced with the aid of generative AI.

\subsection*{Ethics statement}

AbductionBench is built from publicly released datasets, used under their licenses; we add no new human-subject data and collect no personal information. The clinical datasets contain de-identified or synthetic cases, and model outputs on them are not medical advice. AthenaBench describes real threat actors and their publicly reported techniques; we use it only to evaluate attribution and release no new offensive capability. Simulated patients and examiners are played by an LLM, not by people. Automated judging can carry the judge model's biases, which we note as a limitation.

\subsection*{Reproducibility statement}

Section~\ref{sec:methodology} and Appendix~\ref{app:datasets} list every dataset, configuration, and adaptation; Section~\ref{sec:results} and Appendix~\ref{app:evaluation_details} give the models, decoding settings, turn limits, simulators, and judges; Appendix~\ref{app:metrics} defines every metric; and Appendix~\ref{app:prompts} reproduces the judge prompts. Records are sampled with a fixed seed, and every response, judge verdict, and metric is stored per sample. We will release the evaluation framework, dataset adapters, sampled record identifiers, and all model outputs upon publication; an anonymized copy is provided as supplementary material.

\begingroup
\sloppy
\hbadness=10000
\bibliography{iclr2027_conference}
\bibliographystyle{iclr2027_conference}
\endgroup

\appendix

\section{Benchmark Inventory and Taxonomy}
\label{app:benchmark_inventory}

\subsection{Taxonomy and Domain Definitions}
\label{app:domain_taxonomy}

These definitions preserve the working domain taxonomy. Each evaluated task receives one primary label; 
% the final assignment rules and selected task mapping must resolve overlaps between the categories. Inclusion remains subject to the abductive scope described in Section~\ref{sec:framework_scope}.
Because some tasks naturally fit into multiple domains, our mapping rules explicitly resolve these overlaps to ensure each task receives exactly one primary label. Furthermore, fitting into a domain category is not sufficient for inclusion; every selected dataset must still strictly require abductive reasoning, as defined in Section \ref{sec:framework_scope}.

Our collection spans several application domains, each requiring the model to infer a different kind of hidden cause or explanation. We define each domain below.

\noindent\textbf{Healthcare} comprises tasks in which a model infers a diagnosis, differential, or next clinical action from patient symptoms, history, test results, or an ongoing clinical conversation.

\noindent\textbf{Commonsense} comprises tasks in which a model selects or generates the more plausible everyday cause or explanation for a short observed situation, without requiring specialized domain knowledge.

\noindent\textbf{Narrative Reasoning} comprises tasks in which a model infers an explanation that accounts for clues or events scattered across a longer story or mystery.

\noindent\textbf{Formal Reasoning} comprises tasks in which a model generates or selects a missing fact, rule, premise, or hypothesis within a symbolic or logical system, checkable against a formally defined ground truth.

\noindent\textbf{Event Reasoning} comprises tasks in which a model infers the plausible cause or causes of an observed real-world event from a description of that event.

\noindent\textbf{Scientific Discovery} comprises tasks in which a model proposes a hypothesis or mechanism that could explain empirical or literature-grounded observations, typically without an established ground-truth answer.

\noindent\textbf{Scientific Reasoning} comprises tasks in which a model evaluates or infers the degree of support that a body of evidence lends to a scientific claim or hypothesis, spanning multiple reasoning paradigms.

\noindent\textbf{Computing Systems} comprises tasks in which a model infers a plausible underlying cause of a system fault, performance anomaly, or program behavior from observations such as logs, stack traces, telemetry, or binary.

\noindent\textbf{Cybersecurity} comprises tasks in which a model infers a plausible attack, intrusion cause, or responsible actor from incomplete or noisy security-event evidence.

\noindent\textbf{Agentic Systems} comprises tasks in which a model infers which step or steps of an AI agent's trajectory account for its eventual outcome, whether by assigning credit for success or diagnosing the cause of a failure.

\subsection{Dataset and Task Inventory}
\label{app:datasets}

\newcolumntype{C}[1]{>{\centering\arraybackslash}p{#1}}
\newcolumntype{P}[1]{>{\sloppy\raggedright\arraybackslash}p{#1}}
% Local, self-contained row counter: do NOT rely on an externally
% defined/shared \rownum counter, since it may already have been
% incremented elsewhere in the document by the time this table starts
% (this was the cause of the row numbers previously starting at 22
% instead of 1). Defining and resetting our own counter here
% guarantees the numbering always starts at 1 regardless of context.
\newcounter{tblrownum}
\renewcommand{\arraystretch}{1.5}
\setcounter{tblrownum}{0}
\begingroup
\scriptsize
\hbadness=10000
\setlength{\tabcolsep}{1.8pt}
\begin{longtable}{|C{0.3cm}|C{5.5cm}|C{3cm}|C{1.2cm}|C{1cm}|}
\caption{Summary of datasets used. Data Delivery Mode: \textbf{I} = Interactive, \textbf{S} = Static. Hypothesis Processing Mode: \textbf{G} = Generation, \textbf{Sel} = Selection; \textbf{G / Sel} denotes datasets with separate generation and selection tasks.} \\
\hline
\textbf{\#} & \textbf{Name and Citation} & \textbf{Domain} & \textbf{Data\newline Delivery} & \textbf{Hyp.\newline Mode} \\
\hline
\endfirsthead

\hline
\textbf{\#} & \textbf{Name and Citation} & \textbf{Domain} & \textbf{Data\newline Delivery} & \textbf{Hyp.\newline Mode} \\
\hline
\endhead

\hline
\multicolumn{5}{r}{Continued on next page} \\
\endfoot

\hline
\endlastfoot

\stepcounter{tblrownum}\thetblrownum & \href{https://github.com/MaiWert/MedQDx}{MedQDx} \citeauthor{werthaim2026q4dx} (\citeyear{werthaim2026q4dx}) & Healthcare: Diagnostic Questioning & I & G\\
\hline
\stepcounter{tblrownum}\thetblrownum & \href{https://github.com/yf-he/EvoClinician}{Med-Inquire} \citeauthor{he2026evoclinician} (\citeyear{he2026evoclinician}) & Healthcare: Clinical Diagnosis & I & G\\
\hline
\stepcounter{tblrownum}\thetblrownum & \href{https://github.com/LLM4Ops/Cloud-OpsBench}{Cloud-OpsBench} \citeauthor{wang2026cloudopsbench} (\citeyear{wang2026cloudopsbench}) & Computing Systems: Cloud Root-Cause Analysis & I & G\\
\hline
% \stepcounter{tblrownum}\thetblrownum & \href{https://causalgame.github.io/}{CausalGame} \citeauthor{chen2025causalgame} (\citeyear{chen2025causalgame}) & Scientific Discovery: Causal Discovery & I & G\\
% \hline
\stepcounter{tblrownum}\thetblrownum & \href{https://huggingface.co/datasets/chychiu/VivaBench}{VivaBench} \citeauthor{chiu2025vivabench} (\citeyear{chiu2025vivabench}) & Healthcare: Clinical Diagnosis & I & G\\
\hline
% \stepcounter{tblrownum}\thetblrownum & \href{https://github.com/principia-ai/PhysGym}{PhysGym} \citeauthor{chen2025physgym} (\citeyear{chen2025physgym}) & Scientific Discovery: Physics Law Discovery & I & G\\
% \hline
% \stepcounter{tblrownum}\thetblrownum & \href{https://github.com/kanishkg/boxing-gym}{BoxingGym} \citeauthor{gandhi2025boxinggym} (\citeyear{gandhi2025boxinggym}) & Scientific Discovery: Experimental Design and Model Discovery & I & G\\
% \hline
% \stepcounter{tblrownum}\thetblrownum & \href{https://github.com/scientific-discovery/LLM-AutoSciLab}{SciLab} \citeauthor{kabra2025llmautoscilab} (\citeyear{kabra2025llmautoscilab}) & Scientific Discovery: Active Mechanism Discovery & I & G \& Sel\\
% \hline

\stepcounter{tblrownum}\thetblrownum & \href{https://github.com/Waste-Wood/e-CARE/}{e-CARE} \citeauthor{du2022ecare} (\citeyear{du2022ecare}) & Commonsense: Causal Reasoning & S & G / Sel\\
\hline
\stepcounter{tblrownum}\thetblrownum & \href{https://github.com/allenai/abductive-commonsense-reasoning}{ART} \citeauthor{Bhagavatula2020Abductive} (\citeyear{Bhagavatula2020Abductive}) & Narrative Reasoning: Abductive Explanation & S & G / Sel\\
\hline
\stepcounter{tblrownum}\thetblrownum & \href{https://huggingface.co/datasets/allenai/UNcommonsense}{UNCOMMONSENSE} \citeauthor{impl_zhao2023uncommonsense} (\citeyear{impl_zhao2023uncommonsense}) & Commonsense: Uncommon-Situation Abduction & S & G\\
\hline
\stepcounter{tblrownum}\thetblrownum & \href{https://allenai.org/data/proofwriter}{ProofWriter} \citeauthor{impl_tafjord2020proofwriter} (\citeyear{impl_tafjord2020proofwriter}) & Formal Reasoning: Rule-Based Inference & S & G\\
\hline
\stepcounter{tblrownum}\thetblrownum & \href{https://github.com/TartuNLP/true-detective}{True Detective} \citeauthor{del2022truedetective} (\citeyear{del2022truedetective}) & Narrative Reasoning: Mystery Solving & S & Sel\\
\hline
% \stepcounter{tblrownum}\thetblrownum & \href{https://github.com/CTT-Pavilion/_HypoSpace}{HypoSpace} \citeauthor{chen2026hypospace} (\citeyear{chen2026hypospace}) & General: Set-Valued Hypothesis Generation & S & G\\
% \hline
\stepcounter{tblrownum}\thetblrownum & \href{https://github.com/Zayne-sprague/Natural_Language_Deduction_with_Incomplete_Information.git}{ENWN and EntailmentBank} \citeauthor{sprague2022natural} (\citeyear{sprague2022natural}) & Formal Reasoning: Missing-Premise Inference & S & G\\
\hline
\stepcounter{tblrownum}\thetblrownum & \href{https://github.com/SPIRAL-MED/DiagnosisArena}{DiagnosisArena} \citeauthor{zhu2025diagnosisarena} (\citeyear{zhu2025diagnosisarena}) & Healthcare: Clinical Diagnosis & S & G\\
\hline
\stepcounter{tblrownum}\thetblrownum & \href{https://github.com/MAGIC-AI4Med/MedRBench/tree/main/data}{MedR-Bench} \citeauthor{qiu2025medrbench} (\citeyear{qiu2025medrbench}) & Healthcare: Clinical Reasoning & S & G\\
\hline
\stepcounter{tblrownum}\thetblrownum & \href{https://github.com/DeepReasoning/NeuLR}{NeuLR} \citeauthor{xu2023llmslogicalreasoners} (\citeyear{xu2023llmslogicalreasoners}) & Formal Reasoning: Logical Entailment & S & G\\
\hline
\stepcounter{tblrownum}\thetblrownum & \href{https://github.com/mila-iqia/ddxplus}{DDXPlus} \citeauthor{fansitchango2022ddxplus} (\citeyear{fansitchango2022ddxplus}) & Healthcare: Differential Diagnosis & I & Sel\\
\hline
\stepcounter{tblrownum}\thetblrownum & \href{https://github.com/sooo66/semeval2026-task12-dataset.git}{AER} \citeauthor{cao2026semeval} (\citeyear{cao2026semeval}) & Event Reasoning: Abductive Event Causality & S & Sel\\
\hline
\stepcounter{tblrownum}\thetblrownum & \href{https://bit.ly/4p9ltW8}{House M.D.} \citeauthor{gupta2025housemd} (\citeyear{gupta2025housemd}) & Healthcare: Rare-Disease Diagnosis & S & G\\
\hline
\stepcounter{tblrownum}\thetblrownum & \href{https://github.com/andrewbouras/crosstrace}{CrossTrace} \citeauthor{bouras2026crosstrace} (\citeyear{bouras2026crosstrace}) & Scientific Discovery: Cross-Domain Hypothesis Generation & S & G\\
% \hline
% \stepcounter{tblrownum}\thetblrownum & \href{https://github.com/allenai/discoverybench}{DiscoveryBench} \citeauthor{majumder2025discoverybench} (\citeyear{majumder2025discoverybench}) & Scientific Discovery: Data-Driven Discovery & S & G\\
\hline
\stepcounter{tblrownum}\thetblrownum & \href{https://huggingface.co/datasets/UniverseTBD/hypogen-dr1}{HypoGen} \citeauthor{oneill2025sparks} (\citeyear{oneill2025sparks}) & Scientific Discovery: Computer-Science Hypothesis Generation & S & G\\
\hline
\stepcounter{tblrownum}\thetblrownum & \href{https://huggingface.co/datasets/matter2mech/matter-to-mechanism}{Matter to Mechanism} \citeauthor{sourav2026matter} (\citeyear{sourav2026matter}) & Scientific Discovery: Materials Mechanism Inference & S & G\\
% \hline
% \stepcounter{tblrownum}\thetblrownum & \href{https://huggingface.co/datasets/ChicagoHAI/HypoGeniC-datasets}{HypoBench} \citeauthor{liu2026hypobench} (\citeyear{liu2026hypobench}) & Scientific Discovery: Hypothesis Generation & S & G\\
\hline
\stepcounter{tblrownum}\thetblrownum & \href{https://github.com/microsoft/AgentRx}{AgentRx} \citeauthor{barke2026agentrx} (\citeyear{barke2026agentrx}) & AI Systems: Agent Failure Diagnosis & S & Sel\\
\hline
\stepcounter{tblrownum}\thetblrownum & \href{https://www.aiops.cn/gitlab/aiops-live-benchmark/agenticopseval}{RCA100} \citeauthor{cai2026aiops} (\citeyear{cai2026aiops}) & Computing Systems: Microservice Failure Diagnosis & S & Sel\\
\hline
\stepcounter{tblrownum}\thetblrownum & \href{https://github.com/YuSheng-00/UniADILR}{UniADILR-HGc} \citeauthor{sheng2025unadilr} (\citeyear{sheng2025unadilr}) & Formal Reasoning: Constrained Abductive Hypothesis Generation & S & Gen\\
\hline
\stepcounter{tblrownum}\thetblrownum & \href{https://github.com/tdiggelm/climate-fever-dataset}{CLIMATE-FEVER} \citeauthor{diggelmann2021climatefeverdatasetverificationrealworld} (\citeyear{salimi2026cedar}) & Scientific Reasoning: Evidence-Based Claim Explanation & S & Sel\\
\hline
\stepcounter{tblrownum}\thetblrownum & \href{https://github.com/Athena-Software-Group/athenabench}{AthenaBench (TAA)} \citeauthor{alam2025athenabench} (\citeyear{alam2025athenabench}) & Cyber Security: Threat Actor Attribution & P & G\\
\hline
\stepcounter{tblrownum}\thetblrownum & \href{https://github.com/kevinwu23/Stanford-MedCaseReasoning}{MedCaseReasoning} \citeauthor{wu2025medcasereasoning} (\citeyear{wu2025medcasereasoning}) & Healthcare: Clinical Diagnosis & S & G\\
\hline
\stepcounter{tblrownum}\thetblrownum & \href{https://github.com/faezemoradik/CommonWhyDataset.git}{CommonWhy} \citeauthor{toroghi2026commonwhy} (\citeyear{toroghi2026commonwhy}) & Commonsense: Entity-Based Causal Reasoning & S & G\\
\hline
\stepcounter{tblrownum}\thetblrownum & \href{https://github.com/TsinghuaC3I/LLM4BioHypoGen}{LLM4BioHypoGen} \citeauthor{qi2024biohypogen} (\citeyear{qi2024biohypogen}) & Scientific Discovery: Biomedical Hypothesis Generation & S & G\\
\end{longtable}
\endgroup
\subsection{Benchmark Coverage and Related Evaluation Efforts}
\label{app:related-work-coverage}

\subsubsection{Operational Capability Taxonomy}

We distinguish four groups of evaluation targets: \emph{formation} (constructing hypotheses and covering alternatives); \emph{assessment} (evidence support, consistency, contextual plausibility, and explicitly defined parsimony); \emph{discrimination} (selection, ranking, verification, and calibrated uncertainty); and \emph{revision and use} (theory repair, updating under new evidence, predictive testing, and evidence acquisition). This is an operational synthesis, not a claim that these are independent latent abilities or a replacement for the generation--selection distinction \citep{schurz2008patterns,Stanley2020,salimi2026wiring}.

Task descriptors must be kept separate from measured capabilities. These include output representation, supplied versus constructed candidate spaces, static/passive/interactive delivery, domain, and modality. In particular, free-form output need not imply an unrestricted hypothesis space; formal open-world semantics concerns unknown facts, not whether answer options are supplied. Multi-step inference and evidence integration should be credited as independently measured capabilities only when the protocol varies depth or evidence, or scores the corresponding process, rather than merely presenting a long context.

\subsubsection{Representative Benchmark--Capability Matrix}

Table~\ref{tab:abductive-capabilities} concerns the named published task formulations, not every possible use of their data and not necessarily the adapters used in AbductionBench. It is a representative comparison rather than a complete audit of the collection. A dash means that the specified formulation does not separately evaluate that target, not that the capability is absent or unimportant. Entries describe protocol coverage, not model competence.

\begingroup
\scriptsize
\hbadness=10000
\renewcommand{\arraystretch}{1.15}
\setlength{\tabcolsep}{3pt}
\begin{longtable}{@{}P{4.9cm}cccccccc@{}}
\caption{Coverage of representative published protocols. G: hypothesis generation; S: candidate selection/ranking/verification; Q: explanatory-adequacy assessment; M: multiple explanations; R: revision/repair; P: held-out use of a generated hypothesis; U: confidence calibration; A: active evidence acquisition. Symbols and qualifications are defined below.}\label{tab:abductive-capabilities}\\
\hline
\textbf{Benchmark / formulation} & \textbf{G} & \textbf{S} & \textbf{Q} & \textbf{M} & \textbf{R} & \textbf{P} & \textbf{U} & \textbf{A}\\
\hline
\endfirsthead
\hline
\textbf{Benchmark / formulation} & \textbf{G} & \textbf{S} & \textbf{Q} & \textbf{M} & \textbf{R} & \textbf{P} & \textbf{U} & \textbf{A}\\
\hline
\endhead
\hline
\multicolumn{9}{r}{Continued on next page}\\
\endfoot
\hline
\endlastfoot
ART: $\alpha$NLI \citep{Bhagavatula2020Abductive} & -- & + & -- & -- & -- & -- & -- & --\\
ART: $\alpha$NLG \citep{Bhagavatula2020Abductive} & + & -- & J & -- & -- & -- & -- & --\\
e-CARE: cause selection \citep{du2022ecare} & -- & + & -- & -- & -- & -- & -- & --\\
e-CARE: explanation of supplied pair \citep{du2022ecare} & r & -- & J & -- & -- & -- & -- & --\\
UNcommonsense \citep{impl_zhao2023uncommonsense} & + & -- & J & -- & -- & -- & -- & --\\
True Detective \citep{del2022truedetective} & -- & + & -- & -- & -- & -- & -- & --\\
MuSR: murder \citep{sprague2024musr} & -- & + & -- & -- & -- & -- & -- & --\\
AER: cause-set selection \citep{cao2026semeval} & -- & + & -- & j & -- & -- & -- & --\\
ProofWriter: D*-Ab \citep{impl_tafjord2020proofwriter} & + & -- & F & + & -- & -- & -- & --\\
AbductionRules \citep{young2022abductionrules} & + & -- & F & -- & -- & -- & -- & --\\
NeuLR: abductive split \citep{xu2023llmslogicalreasoners} & + & -- & -- & -- & -- & -- & -- & --\\
AbductiveKGR \citep{bai2024advancing} & + & -- & F & -- & -- & -- & -- & --\\
ABD \citep{batzoglou2026abd} & + & -- & F & -- & T & + & -- & --\\
DeFAb: L2 candidate protocol \citep{cooper2026defab} & -- & + & F & -- & -- & -- & -- & --\\
DeFAb: L3 defeater construction \citep{cooper2026defab} & + & -- & F & -- & T & -- & -- & --\\
HypoSpace \citep{chen2026hypospace} & + & -- & F & + & -- & -- & -- & --\\
DDXPlus: interactive differential \citep{fansitchango2022ddxplus} & -- & + & -- & + & $\sim$ & -- & -- & +\\
MedCaseReasoning \citep{wu2025medcasereasoning} & + & -- & p & -- & -- & -- & -- & --\\
HypoBench \citep{liu2026hypobench} & + & -- & J & h & $\sim$ & + & -- & --\\
BoxingGym \citep{gandhi2025boxinggym} & + & -- & p & -- & $\sim$ & + & -- & +\\
Cloud-OpsBench \citep{wang2026cloudopsbench} & + & -- & p & -- & $\sim$ & -- & -- & +\\
GEAR protocol \citep{impl_he2025gear} & + & -- & F & + & -- & c & -- & --\\
Elenchos$^{\dagger}$ \citep{steiglechner2026elenchos} & -- & + & -- & j & $\sim$ & -- & -- & +\\
Black Swan: Detective/Reporter$^{\dagger}$ \citep{chinchure2025blackswan} & + & + & J & -- & E & -- & -- & --\\
MolQuest$^{\dagger}$ \citep{han2026molquest} & + & -- & F & -- & $\sim$ & -- & + & +\\
\end{longtable}
\endgroup

\noindent\textbf{Reading the matrix.}
$+$ denotes an explicit task or scored target; $\sim$ denotes an available update/refinement process without a separate revision-correctness score. Q distinguishes formal checks or formally enumerated targets (F), human/LLM judgments (J), and partial proxies (p); these do not have equivalent evidential strength. For example, MedCaseReasoning's reasoning recall measures coverage of reference clinical reasons, not the validity of every generated statement; Cloud-OpsBench checks evidence milestones; MolQuest's formula conservation is not complete spectral verification. In G, r denotes explaining a supplied causal pair rather than constructing its missing cause. In M, j denotes jointly relevant causes or mutations, not competing explanations, and h denotes recovery of planted hypothesis components, not exhaustive coverage of admissible alternatives. In R, T denotes static theory repair and E denotes reassessment against additional evidence. In P, c denotes GEAR's prediction-domain coverage; its separate hidden-case analyses also examine predictive correctness. All P entries concern use of a \emph{generated hypothesis}, not merely a held-out benchmark test split.

\noindent\textbf{Scope qualifications.}
DDXPlus's original differential prediction uses a predefined disease vocabulary; open-text diagnosis generation is a separate adaptation. ProofWriter's abductive split is distinct from its entailment and proof-generation tasks. AbductiveKGR scores agreement between a generated formula's answers and observations; RLF-KG is the training method, not the dataset name. DeFAb's candidate-based L2 results should not be merged with L3 construction or its additional CONJURE track. Elenchos submits a structured diagnosis from a specified finite mutation ontology, so it is coded as selection even though the response is serialized as text. Black Swan's marks combine explicitly separate task variants, not a single generation--selection pipeline. $\dagger$ identifies external comparators not listed among the current 48 dataset-table entries. The matrix does not add them to the evaluated collection.

\subsubsection{Closest Multi-Task and Multi-Benchmark Efforts}

\begingroup
\scriptsize
\hbadness=10000
\renewcommand{\arraystretch}{1.15}
\setlength{\tabcolsep}{3pt}
\begin{longtable}{@{}P{2.25cm}P{3.15cm}P{3.25cm}P{3.25cm}@{}}
\caption{Related evaluation efforts, separated by objective and protocol rather than ranked by dataset count. Resource sizes use different counting units and are not directly comparable.}\label{tab:abductive-frameworks}\\
\hline
\textbf{Work} & \textbf{Scope and interface} & \textbf{Evaluation target} & \textbf{Relation to AbductionBench}\\
\hline
\endfirsthead
\hline
\textbf{Work} & \textbf{Scope and interface} & \textbf{Evaluation target} & \textbf{Relation to AbductionBench}\\
\hline
\endhead
\hline
\multicolumn{4}{r}{Continued on next page}\\
\endfoot
\hline
\endlastfoot
GEAR \citep{impl_he2025gear} & Four source benchmarks reformulated as executable hypothesis generation; fixed observations. & Shared consistency, prediction-domain coverage, and behavioral diversity; separate hidden-case tests. & Direct evaluation-framework predecessor with released code; common executable representation rather than preservation of heterogeneous native interfaces.\\
Wiring the `Why' \citep{salimi2026wiring} & Nine datasets, twelve formulations: commonsense, narratives, diagnosis, missing facts; static empirical suite. & Generation and selection; exact, ranking, overlap, and judge metrics; fixed direct prompts. & Closest taxonomy and multi-dataset study. Selective experiments ground the survey under a controlled protocol; our emphasis is heterogeneous benchmark execution.\\
HypoBench \citep{liu2026hypobench} & Seven real and five synthetic tasks, 194 datasets; natural-language hypotheses for classification. & Predictive utility, IND/OOD testing, planted hypothesis recovery, qualitative ratings. & Multi-dataset generation precedent with public implementation; not a general selection or interactive-diagnosis collection.\\
HypoSpace \citep{chen2026hypospace} & Three structured domains with enumerable admissible sets, plus real-data case studies. & Validity, uniqueness, recovery under fixed evidence. & A genuine multi-domain diagnostic framework for underdetermination, rather than a heterogeneous benchmark runner.\\
CEDAR-GRPO \citep{salimi2026cedar} & Eleven held-out tasks, including adjacent tasks and non-abductive controls. & Post-training transfer, task outcomes, and trace-level metrics. & Substantial empirical and software overlap: released evaluation orchestration and task scripts; training transfer is the central objective.\\
Elenchos \citep{steiglechner2026elenchos} & Procedurally generated kernel mutations; interactive probes within a specified ontology. & Corrupted-kernel identification, mutation attribution, and query efficiency. & Dedicated extensible evaluation framework, but not an aggregation of existing datasets.\\
IBE evaluation \citep{dalal2024ibe} & COPA and e-CARE; explanations for supplied causal options. & Logical and linguistic explanatory criteria, including uncertainty-language proxies. & Earlier multi-dataset explanatory-quality evaluation; neither uncertainty calibration nor unrestricted hypothesis generation.\\
Logical-reasoning evaluation \citep{xu2023llmslogicalreasoners} & Fifteen existing datasets across deduction, induction, abduction, and mixed reasoning; new NeuLR. & Answer/explanation evaluation and reasoning-error analysis. & Broader reasoning precedent; abduction is one component, not the exclusive organizing target.\\
BoxingGym \citep{gandhi2025boxinggym} & Ten probabilistic environments with experimental interfaces. & Information gain, predictive performance, communicable models. & Reusable multi-environment discovery infrastructure; interaction and extensibility are not unique to AbductionBench.\\
AbductionBench (design scope) & Collection of 48 entries; heterogeneous domains, outputs, and evidence-delivery modes. & Native metrics; generation/selection/pipelines; prompting and selection-format comparisons. & Intended contribution is joint protocol coverage and reusable execution. Adapter coverage and experimental subsets require separate reporting.\\
\end{longtable}
\endgroup
% \input{Sections/Appendix/dataset_samples}

% Metrics are intentionally lettered Z to keep them visually distinct from
% the descriptive appendices requested in the manuscript outline.
% \setcounter{section}{25}
\section{Metrics}
\label{app:metrics_section}
\subsection{Metrics}
\label{app:metrics}
We organize evaluation into three non-overlapping metric families. \emph{Outcome metrics} score the final answer or terminal result. \emph{Reasoning metrics} analyze explicit chain-of-thought (CoT) traces for static tasks. \emph{Interactive and passive metrics} characterize multi-turn behavior.

\subsubsection{Outcome Metrics}
For single-choice selection (SCS) tasks, we report accuracy, i.e., the fraction of questions for which the model selects the correct option. For multiple-choice or set-valued (MCS) tasks, we report exact set match and macro F1. Exact set match is the fraction of questions whose answer is generated or selected exactly as required, while F1 is the average F1 between the model's answer and the gold answer; the leaderboard is sorted by exact set match. For generation tasks whose outputs cannot be compared with the gold label by formal methods, an LLM judge returns a binary judgment of whether the answer is equivalent to the gold reference. When executable tests or task-native verifiers are available, they replace text-based comparison.

\subsubsection{Reasoning Metrics}
Reasoning metrics apply only to CoT samples, using either the visible CoT or the native reasoning trace, and only when the model produced an answer. A single segmentation judge cuts the trace into \(n\) inferential steps; every per-step metric must return exactly \(n\) elements or is rejected. The judge returns counts and per-step lists only; ratios, fractions, and normalized values are computed in code. We compute the following metrics.

\begin{description}

\item[\texttt{observations total}.] The number of distinct observations the question supplies. These are the facts any answer must account for.

\item[\texttt{observations used}.] The number of those observations the reasoning chain actually draws on, counting each observation once at its first appearance.

\item[\texttt{observation coverage}.] The ratio of observations used to observations total. It measures the share of the supplied evidence the chain uses.

\item[\texttt{total steps}.] The length of the segmented reasoning chain in inferential moves.

\item[\texttt{useful steps}.] The number of steps that explicitly prove or disprove at least one hypothesis, option, cause, or candidate answer.

\item[\texttt{useless steps}.] The number of steps that do not explicitly prove or disprove any hypothesis. Restatements, setup, and bookkeeping are counted here.

\item[\texttt{useful step fraction}.] The fraction of steps that are useful.

\item[\texttt{useless step fraction}.] The fraction of steps that are useless.

\item[\texttt{backtracking steps}.] The number of steps in which the model explicitly recognises an error or contradiction in its own earlier reasoning and changes direction.

\item[\texttt{backtracking rate}.] The fraction of steps that contain at least one such self-correction.

\item[\texttt{branchiness total}.] The total number of new, distinct hypotheses, explanations, causes, or candidate answers the chain raises across all steps. A hypothesis already raised earlier is not counted again.

\item[\texttt{diversity}.] A binary judgment for generation tasks: whether the hypotheses raised across the chain are meaningfully different from one another, rather than rewordings of a single explanation.

\item[\texttt{option count}.] For selection tasks, the number of answer options the question supplies, as read from the question.

\item[\texttt{directionality}.] A whole-chain judgment of inference direction: \(1\) if the chain reasons from evidence towards an explanation, \(0\) if it assumes or guesses an explanation first and then justifies it forward, and \(0.5\) for mixed or ambiguous cases.

\item[\texttt{step directionality mean}.] The mean of the per-step directionality values, using the same \(1/0.5/0\) scale applied step by step.

\item[\texttt{differential elimination}.] The total number of comparison events in which the model directly weighs two or more hypotheses against each other. One event comparing three or more hypotheses together counts as one event, not three.

\item[\texttt{differential elimination normalized}.] The fraction of steps that contain at least one such comparison event.

\item[\texttt{comparison exhaustiveness}.] The number of comparison events divided by \(\binom{n}{2}\), where \(n\) is \texttt{reasoning\_option\_count} for selection tasks or \texttt{reasoning\_branchiness\_total} for generation tasks.

\item[\texttt{uncertainty steps}.] The total number of uncertainty markers across the chain, such as hedges, explicit statements of doubt, or stated alternatives the model cannot decide between. Multiple markers in one step are counted separately.

\item[\texttt{uncertainty rate}.] The fraction of steps that contain at least one uncertainty marker.

\item[\texttt{prior knowledge}.] The total number of non-trivial background facts the reasoning explicitly invokes that are not present in the question itself.

\item[\texttt{prior knowledge normalized}.] The fraction of steps that contain at least one such background reference.

\item[\texttt{anchoring point}.] The 1-based index of the first step in which the model considers its own final answer as a candidate answer. Considering is enough: raising it, listing it, entertaining it, arguing for or against it. It need not be preferred or settled on. The value is \texttt{None} if the model never considers its own final answer anywhere in the chain.

\item[\texttt{anchoring point normalized}.] The anchoring point divided by the number of steps. A value of \(1.0\) means the model's final answer first appears only at the last step.

\item[\texttt{gold alive steps}.] The number of steps in which the correct answer is in play as a candidate: raised, weighed, argued for, or argued against. This is a per-step state, not a running state; a later step that takes up the correct answer again counts again.

\item[\texttt{gold alive rate}.] The fraction of steps in which the correct answer is in play.

\item[\texttt{unresolved contradictions}.] The number of steps whose claim is later contradicted and never acknowledged or resolved by the model.

\item[\texttt{unresolved contradiction normalized}.] The fraction of steps with an unresolved contradiction.

\item[\texttt{proof disproof total}.] The total number of hypotheses, options, causes, or candidate answers the model explicitly proves or disproves across the chain.

\item[\texttt{mentions total}.] The total number of references to candidate answers across the chain, including repeated references and the step that first raises each candidate.

\item[\texttt{mention ratio}.] Mentions total divided by branchiness total. It measures how often each hypothesis is revisited.

\item[\texttt{helpfulness mean}.] The mean of the per-step helpfulness values: \(1\) for a helpful step that moves the reasoning toward the reference answer, \(0\) for a neutral step, and \(-1\) for a harmful step that asserts something false, unsupported, or misleading. A step that catches and fixes an error is helpful.

\item[\texttt{helpful steps}, \texttt{neutral steps}, \texttt{harmful steps}.] Counts of steps scored \(1\), \(0\), and \(-1\), respectively.

\item[\texttt{helpful fraction}, \texttt{neutral fraction}, \texttt{harmful fraction}.] The corresponding fractions of steps.

\end{description}

\subsubsection{Interactive and Passive Metrics}
Interactive and passive tasks are run in IO mode with internal reasoning disabled, and each turn receives the history of prior turns. For interactive and passive tasks, we compute the following.

\begin{itemize}
\item \textbf{Interaction-step relevance.} Whether each interaction action or query whether it moved the investigation toward the correct answer.
      
\item \textbf{Number of turns to final output.} The total number of turns until the model produces its final output.
\item \textbf{Final-answer accuracy.} The accuracy of the model's last answer. It is computed by exact match where possible and by an LLM answer judge otherwise.
\end{itemize}

For passive tasks, we also compute the following.

\begin{itemize}
\item \textbf{Number of turns to correctness.} The first turn at which the model predicts the correct answer. If the model never predicts correctly, the value is \texttt{None} and the sample is excluded from the average.
\item \textbf{Average sample accuracy.} The mean prediction accuracy across turns within each sample.
\item \textbf{Number of turns to the first prediction of the final output.} The first turn at which the model predicts the same output it eventually gives as its final answer. This may be the final turn or an earlier turn.
\end{itemize}

\section{Extended Results and Case Studies}
\label{app:qualitative_results}

\subsection{Additional Tables and Figures}
\subsubsection{Per-Model Correctness, IO vs.\ CoT}
\label{sec:by_model_correctness}
\FloatBarrier

\begin{figure}[H]
  \centering
  \includegraphics[width=0.8\linewidth]{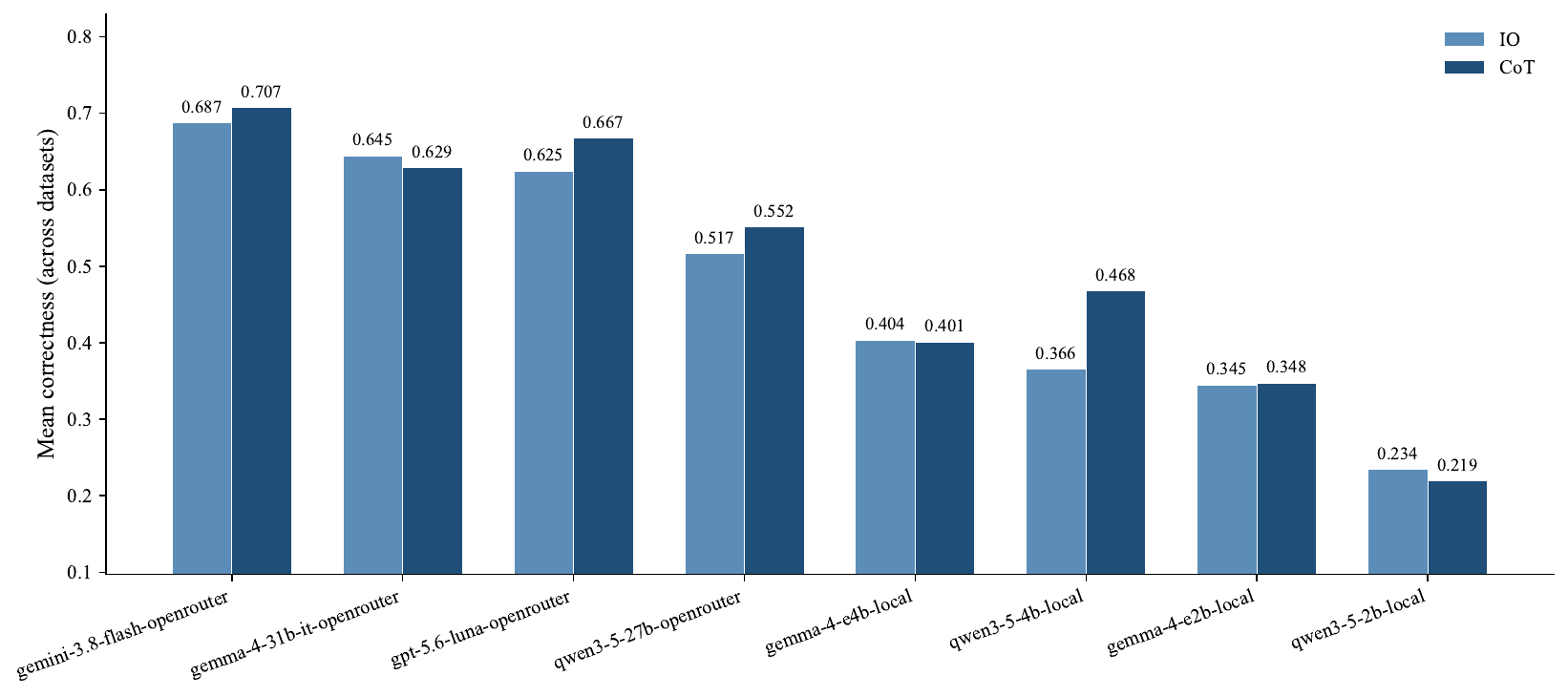}
  \caption{\emph{Mean correctness per model, averaged across datasets, for the \textsc{io} and \textsc{cot} prompting variants.} Models are ordered by \textsc{cot} correctness, descending. \textsc{cot} improves correctness for four of eight models, is roughly flat for \textsc{gemma-4-e2b-local}, and reduces it slightly for \textsc{gemma-4-31b-it} and \textsc{gemma-4-e4b-local}; the largest gain is \textsc{qwen3-5-4b-local} ($+0.102$) and the largest loss is negligible ($-0.015$, \textsc{qwen3-5-2b-local}).}
  \label{fig:per_model_correctness}
\end{figure}

Figure~\ref{fig:per_model_correctness} reports mean correctness by model,
averaged across datasets, for the \textsc{io} and \textsc{cot} prompting
conditions separately. \textsc{cot} is not a uniform win: it improves
\textsc{gemini-3.8-flash}, \textsc{gpt-5.6-luna}, \textsc{qwen3-5-27b}, and
\textsc{qwen3-5-4b-local}, leaves \textsc{gemma-4-e2b-local} essentially
unchanged, and slightly \emph{hurts} \textsc{gemma-4-31b-it},
\textsc{gemma-4-e4b-local}, and \textsc{qwen3-5-2b-local}. There is no clean
split by model family or by parameter count---both \textsc{gemma} variants
that benefit least from \textsc{cot} sit on opposite ends of the size
range---so whatever determines whether a model benefits from chain-of-thought
prompting on this task is not simply "bigger models benefit more" or
"family X benefits, family Y doesn't." Overall correctness itself does track
scale reasonably well within a family (e.g. \textsc{qwen3-5-27b} $>$
\textsc{qwen3-5-4b-local} $>$ \textsc{qwen3-5-2b-local} under both
conditions), but the \textsc{cot} effect is not explained by that same
ordering.

\FloatBarrier
\subsubsection{Single-Choice vs.\ Multiple-Choice-Allowed Selection}
\label{sec:single_vs_multiple_choice}
\FloatBarrier

\begin{figure}[H]
  \centering
  \includegraphics[width=0.8\linewidth]{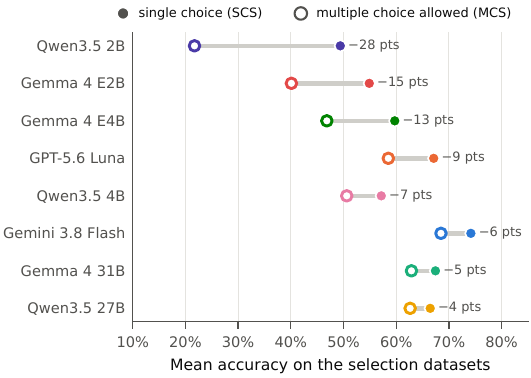}
  \caption{\emph{Mean accuracy on the selection datasets under single-choice
  (\textsc{scs}) versus multiple-choice-allowed (\textsc{mcs}) formats.}
  Models are ordered by \textsc{scs} accuracy, descending. Allowing multiple
  selections reduces mean accuracy for every model; the drop ranges from
  $-4$ points (\textsc{qwen3-5-27b}) to $-28$ points
  (\textsc{qwen3-5-2b-local}).}
  \label{fig:single_vs_multiple_choice}
\end{figure}

Figure~\ref{fig:single_vs_multiple_choice} compares mean accuracy on the
selection datasets when models must choose exactly one hypothesis
(\textsc{scs}) against a format that permits selecting more than one
(\textsc{mcs}). Every model loses accuracy under \textsc{mcs}, but the size
of the drop is far from uniform: \textsc{qwen3-5-27b}, \textsc{gemma-4-31b-it},
\textsc{gemini-3.8-flash}, and \textsc{qwen3-5-4b-local} lose between 4 and 7
points, \textsc{gpt-5.6-luna} loses 9, \textsc{gemma-4-e4b-local} and
\textsc{gemma-4-e2b-local} lose 13 and 15, and \textsc{qwen3-5-2b-local} loses
28---roughly four to seven times the penalty paid by the largest models.
The ordering tracks scale within each family: among the \textsc{qwen} models,
the penalty rises monotonically as parameter count falls
($27\text{b}\!\to\!4\text{b}\!\to\!2\text{b}$: $-4, -7, -28$), and the two
smaller \textsc{gemma} variants pay a substantially larger penalty than
\textsc{gemma-4-31b-it}. This suggests that removing the constraint of a
single correct choice does not uniformly increase the difficulty of
hypothesis selection---it disproportionately harms smaller models, consistent
with multiple-choice-allowed formats demanding a more reliable notion of
per-hypothesis confidence (rather than a single relative ranking) that
smaller models are less able to supply.

\FloatBarrier

\begin{figure}[t]
  \centering
  \includegraphics[width=0.88\linewidth]{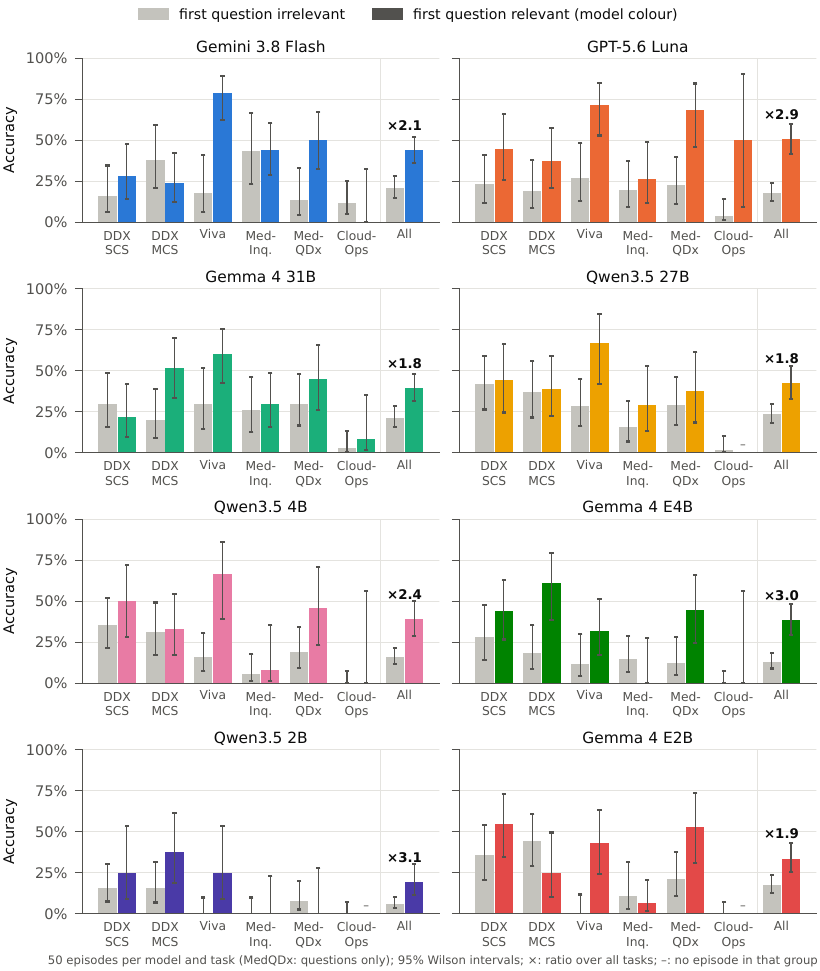}
  \caption{Final-answer accuracy of interactive episodes whose \emph{first} action was judged irrelevant (gray) or relevant (color) to the gold answer, per task and pooled (\emph{All}), per model; 95\% Wilson intervals. $\times$: relevant-to-irrelevant ratio over all tasks.}
  \label{fig:first_question}
\end{figure}

\subsubsection{First Question Quality vs.\ Interactive Diagnosis Accuracy}
\label{sec:first_question}

Largely, yes. Only 35\% of interactive episodes open with an action judged relevant to the gold answer, but those end correctly 2.4 times as often as the rest (39.7\% vs.\ 16.6\%), and every model shows the effect, with ratios from $\times$1.8 (Gemma-4-31B, Qwen-3.5-27B) to $\times$3.1 (Qwen-3.5-2B) (Figure~\ref{fig:first_question}). Relevance then declines as the episode proceeds for every model (Appendix~\ref{app:relevance_by_step}), so later questions rarely repair a poor start. The relation is associational---a model that understands the case both asks better and diagnoses better---but it shows that the quality of evidence \emph{seeking}, invisible in final-answer accuracy, is measurable from the first step.

\subsubsection{Reasoning Length by Input Context Length}
\label{app:horizon_by_context}

\begin{figure}[H]
  \centering
  \includegraphics[width=\linewidth]{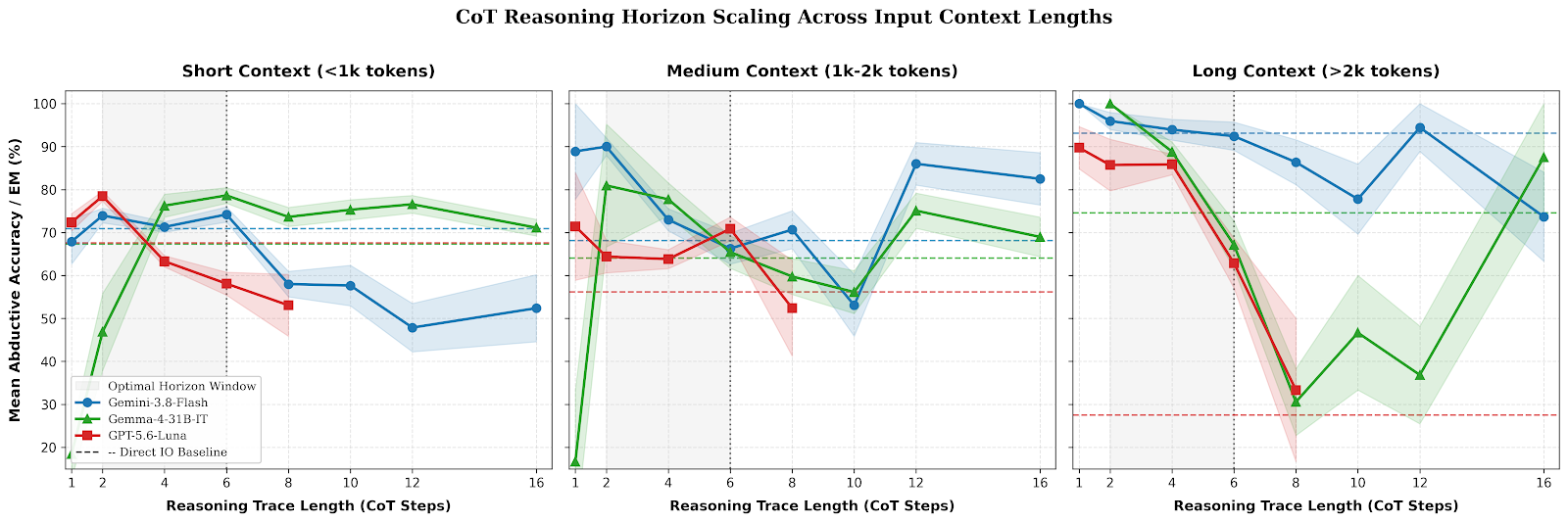}
  \caption{Accuracy by CoT trace length (steps) for Gemini-3.8-Flash, Gemma-4-31B, and GPT-5.6-Luna, split by input length: short ($<$1k tokens, $n=9{,}684$ traces), medium (1k--2k, $n=2{,}663$), and long ($>$2k, $n=1{,}123$). Dashed lines: each model's IO accuracy in the same tier; bands are $\pm$SEM, points with $n<5$ omitted.}
  \label{fig:horizon_by_context}
\end{figure}

The decline beyond six steps seen in Figure~\ref{fig:horizon_scaling} holds within each input-length tier (Figure~\ref{fig:horizon_by_context}). On short inputs, GPT-5.6-Luna peaks at two steps (78.5\%) and falls to 53.1\% at eight, and Gemini-3.8-Flash falls from 74.2\% at six steps to 47.9\% at twelve, while Gemma-4-31B needs several steps before it exceeds its IO baseline (18.4\% at one step, 78.6\% at six). On long inputs the effect of CoT is largest in both directions: GPT-5.6-Luna scores only 27.5\% under IO but 85.8\% with three- to four-step chains, then 33.3\% at eight steps, and Gemma-4-31B drops from 88.9\% at four steps to 30.4\% at eight. Gemini-3.8-Flash is the most stable on long inputs, near or above its IO baseline up to six steps. The long tier has the fewest traces, so its curves have the widest bands, and, as in the main text, the relation is associational: longer inputs and longer chains both mark harder cases.

\FloatBarrier
\label{app:additional_results}
\subsubsection{Model-Wise Metric Profiles Across the Reasoning Chain}
\label{sec:by_model}
\FloatBarrier

\begin{figure}[H]
  \centering
  \includegraphics[width=\linewidth]{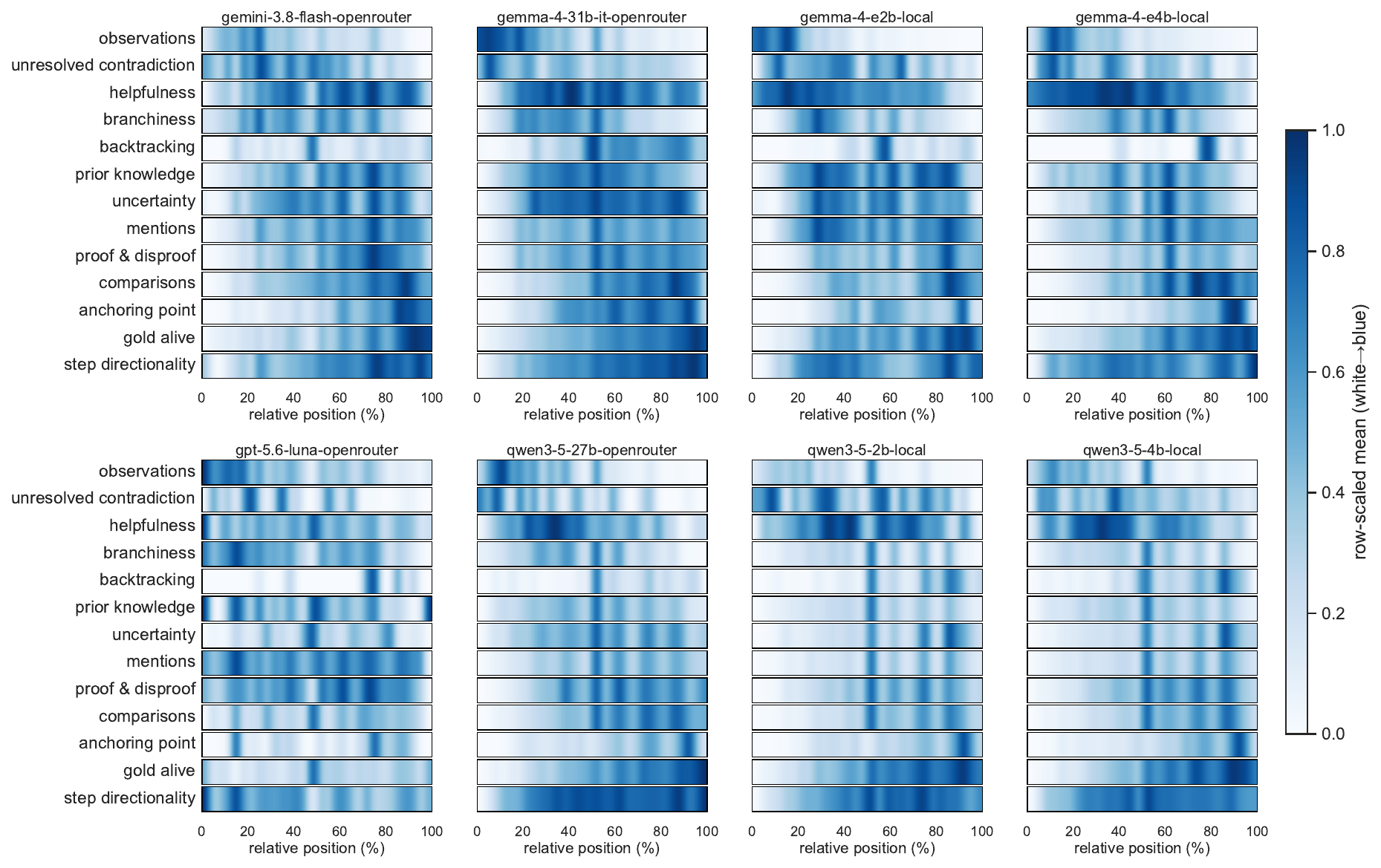}
  \caption{Mean of each judged metric by relative position in the chain
  (0--100\%), shown separately per model, for all eight models evaluated.
  Each row is independently min--max scaled across all panels, so peak
  \emph{timing} is comparable within a metric across models but peak
  \emph{height} is not.}
  \label{fig:relative_position_heatmap_by_model}
\end{figure}

Figure~\ref{fig:relative_position_heatmap_by_model} breaks the pooled view of
Figure~\ref{fig:position_heatmaps} out by model. We describe the
pattern in detail for three models---\textsc{gemini-3.8-flash},
\textsc{gemma-4-31b-it}, and \textsc{gpt-5.6-luna}---and note where the
remaining five models (\textsc{gemma-4-e2b-local}, \textsc{gemma-4-e4b-local},
\textsc{qwen3-5-27b-openrouter}, \textsc{qwen3-5-2b-local},
\textsc{qwen3-5-4b-local}) appear to fall; a full per-model, per-metric
characterization of all eight is left to future work.

\textsc{gemini-3.8-flash} and \textsc{gemma-4-31b-it} both show a single,
unimodal \emph{anchoring point} concentrated in the last fifth of the chain:
each settles on its final answer once, late. \textsc{gpt-5.6-luna} does not
share this shape---its \emph{anchoring point} is visibly bimodal, with a
distinct early peak around the first third of the chain in addition to the
late peak the other two models share. Rather than converging on its answer
at a single characteristic point in the chain, \textsc{gpt} appears to
commit at one of two different points depending on the trace (Appendix~\ref{case:gpt_bimodal} shows both modes on the same item), a pattern not
present in \textsc{gemini} or \textsc{gemma}; whether it is present in the
remaining five models has not been checked.

This split is not confined to \emph{anchoring point}. \emph{Helpfulness},
\emph{step directionality}, \emph{comparisons}, \emph{gold alive}, and
\emph{prior knowledge} all rise through the chain and peak in the final
10--20\% for \textsc{gemini} and \textsc{gemma}, the same late-climax shape as
\emph{anchoring point} above. \textsc{gpt} again departs: \emph{helpfulness},
\emph{step directionality}, \emph{prior knowledge}, \emph{mentions}, and
\emph{branchiness} peak in its first 20\% rather than its last, and
\emph{comparisons} and \emph{gold alive} peak around the chain's midpoint
rather than the end (Appendix~\ref{case:frontload_backload} contrasts a
front-loaded \textsc{gpt} trace with a back-loaded \textsc{gemma} trace on
the same item). The pooled curve of
Figure~\ref{fig:position_heatmaps} therefore does not represent
\textsc{gpt} well; among the three models discussed here it reads far closer
to \textsc{gemini}/\textsc{gemma}. Whether it represents the other five
models well is a separate, unverified question---the pooled curve averages
over all eight, and a minority of back- or front-loaded models among the
remaining five could shift that average in either direction without being
described above. (\emph{Prior knowledge}'s late-chain peak for \textsc{gpt}
rests on comparatively few, longer chains and should be read with that
caveat.)

\emph{Unresolved contradiction} follows the same split, again restricted to
the three models discussed here: \textsc{gemini} and \textsc{gemma} each show
one clean early bump (\raise.17ex\hbox{$\scriptstyle\sim$}15--20\%) and taper
off, while \textsc{gpt} shows three distinct bumps spread across the chain
(\raise.17ex\hbox{$\scriptstyle\sim$}15\%, \raise.17ex\hbox{$\scriptstyle\sim$}35\%,
\raise.17ex\hbox{$\scriptstyle\sim$}60\%)---consistent with \textsc{gpt} returning to
introduce or drop claims at several points rather than settling once.
\emph{Backtracking} peaks at roughly the same point
(\raise.17ex\hbox{$\scriptstyle\sim$}45--50\%) for \textsc{gemini} and \textsc{gemma} but later
(\raise.17ex\hbox{$\scriptstyle\sim$}70--80\%) for \textsc{gpt}; given how rarely \textsc{gpt}
backtracks at all (1 of 684 traces, Section~\ref{sec:main_results}), this is
noted only as a minor aside.

\FloatBarrier

\newtcolorbox{sampling}[1]{
    enhanced,
    breakable, % Allows the box to break across pages!
    width=0.96\linewidth, % Tweak this to 0.96 or 0.97 to reflow text and fix the runts
    center, % Centers the slightly narrower box on the page
    colback=white, % Background color of the text area
    colframe=black, % Border color
    colbacktitle=gray!20, % Title bar background color
    coltitle=black, % Title text color
    fonttitle=\bfseries,
    title={#1},
    sharp corners,
    boxrule=0.5pt,
    left=2mm, right=2mm, top=2mm, bottom=2mm % Inner padding
}

\subsection{Qualitative Case Studies}
\label{app:interesting_samples}

The traces below illustrate, at the level of individual reasoning chains, three patterns quantified in the main text: GPT-5.6-Luna's bimodal anchoring point (Section~\ref{sec:by_model}), the early appearance of unresolved contradiction in incorrect chains (Section~\ref{sec:chain_position}), and the contrast between front-loaded and back-loaded reasoning shape (Section~\ref{sec:cognitive_profiles_and_quality_gap}, Section~\ref{sec:by_model}).

\subsubsection{Early vs.\ Late Anchoring on the Same Item}
\label{case:gpt_bimodal}

Both traces below are GPT-5.6-Luna responses to the same \texttt{true\_detective} item and reach the same (correct) answer, but commit to it at different points in the chain, illustrating the bimodal anchoring pattern described in Section~\ref{sec:by_model}.

\begin{sampling}{Task Input: Stealing Second Base}
[SYSTEM]

You are an expert at abductive reasoning: inferring the explanation that, if true, would best account for the evidence you are given. You are given a mystery. Identify the suspect whose actions best explain all of the available evidence.

Reason and explain explicitly by working through the evidence step by step before answering. Consider whether each candidate explanation accounts for the complete set of observations rather than only part of them.

Put all of your reasoning inside one think block, then give your answer inside one answer block:

< think >your reasoning< /think >

< answer >your answer< /answer >

Write nothing before the think block and nothing after the answer block.

[USER]

Question: Who is responsible for stealing the autographed second base?

Options:

A. Coach Joe Morgan

B. Mary Thornton

C. Randy Newsom

D. Shorty Gilstrap

Select exactly one hypothesis.

Answer with only one of: A, B, C, or D.
\end{sampling}

\begin{sampling}{Response A -- Early Anchor}
<think>
The evidence points to Mary Thornton.

She had access to the principal's keys, returned to the display case after the others had left, opened it, and had a private opportunity to remove the base.

Her oversized sweater provided a plausible way to conceal it.

Randy and Shorty's accounts place the base in the case before Mary's later visit, while Coach Morgan had no stated opportunity.

Therefore Mary is responsible.
</think>

<answer>B</answer>
\end{sampling}

\begin{sampling}{Response B -- Late Anchor}
<think>
The base was present when Shorty left and when Mary initially helped clean.

Mary later obtained the principal's keys, reopened and relocked the case, and was the last person known to handle its contents.

Her moving a trophy provided a pretext for that access, while her oversized sweater could conceal the base.

Thus Mary's account is the false one, and she is responsible.
</think>

<answer>B</answer>
\end{sampling}

Response A commits to Mary Thornton in its first reasoning step and elaborates from there; Response B withholds the culprit until the final step. Both are correct, showing that GPT-5.6-Luna's early- and late-anchoring modes are not simply an accuracy split.

\subsubsection{Unresolved Contradiction from the First Step}
\label{case:early_contradiction}

The two traces below are GPT-5.6-Luna responses to the same \texttt{true\_detective} item (gold: Uncle Ezra). The incorrect trace introduces an unsupported premise in its first sentence and never revisits it, consistent with Section~\ref{sec:chain_position}'s finding that unresolved contradiction is elevated in incorrect chains from the earliest steps, before any hypothesis is anchored.

\begin{sampling}{Response A -- Incorrect (Gold: D)}
<think>
The strongest evidence points to Rusty.

He had a special attachment to the signed book, knew that chapter 16 was significant, and had the opportunity to take the book while everyone was searching for Cupcake.

The note was tucked into chapter 16, directly connecting the thief to Rusty's known interest.

His sadness at the initial apparent theft also fits someone emotionally attached to the book, while Ann's jealousy and Alfred's access are distractions.

Rusty later returned the book with an apology, explaining why it reappeared in a package.

Uncle Ezra's decoy scheme explains the first theft but not the disappearance of the real book.

Therefore, Rusty is responsible.
</think>

<answer>C</answer>
\end{sampling}

\begin{sampling}{Response B -- Correct}
<think>
The evidence best supports Uncle Ezra.

He had sole control over the real book's hiding place, knew exactly when the children and Alfred were distracted, and was the only person positioned to ``find'' Cupcake in the basement while secretly handling the book.

His practical-joker behavior, love of mystery plots, and earlier decoy scheme provide the motive and method for staging a second theft.

The returned package and note are likewise consistent with Ezra arranging an elaborate prank.

The other suspects had opportunity, but none explains the carefully timed cat distraction and theatrical return as completely.
</think>

<answer>D</answer>
\end{sampling}

The incorrect trace treats a single detail (the note in Chapter 16) as decisive evidence against Rusty in its second sentence, then interprets all later evidence through that early commitment rather than testing it against Ezra's exclusive access to the book and the staged distraction. The correct trace uses the same underlying events---the cat distraction, the decoy scheme, the theatrical return---as evidence of Ezra's control over the whole sequence.

\subsubsection{Front-Loaded vs.\ Back-Loaded Reasoning on the Same Item}
\label{case:frontload_backload}

The two traces below answer the same \texttt{art} item. GPT-5.6-Luna commits early and closes with instruction-following considerations rather than task content; Gemma-4-31B-IT enumerates the observations and options neutrally before committing near the end. This pairing illustrates the model-level contrast in reasoning shape reported in Section~\ref{sec:by_model}.

\begin{sampling}{Task Input}
Observation 1: My wife had MLK day off.

Observation 2: Then we had a relaxing evening covering on the couch.

Options:

1. My wife and I spent the day bicycling at the lake.

2. We did not have time to do anything.

Select exactly one hypothesis. Answer with only one of: 1 or 2.
\end{sampling}

\begin{sampling}{GPT-5.6-Luna -- Front-Loaded}
<think>
**Evaluating candidate options**

I'm considering the candidates: ``spent day bicycling at lake'' naturally suggests tiredness and relaxation, while ``did not have time to do anything'' doesn't clarify the day off.

The latter might imply inactivity, leading to couch time.

I think the first option is the most plausible since it describes a holiday activity followed by relaxation.

But I'm reminded not to add extra causes.

I need to select one option carefully while following the developer's instructions.
</think>

<answer>1</answer>
\end{sampling}

\begin{sampling}{Gemma-4-31B-IT -- Back-Loaded}
<think>
* O1: ``My wife had MLK day off.'' (Establishes a holiday/free time for the spouse).

* O2: ``Then we had a relaxing evening covering on the couch.'' (Describes a low-energy, restful conclusion to the day).

* Option 1: ``My wife and I spent the day bicycling at the lake.''

* This fits O1 because they had the day off to engage in an activity.

* This fits O2 because spending a day being physically active (bicycling) typically leads to wanting a ``relaxing evening'' on the couch to recover.

* Option 2: ``We did not have time to do anything.''

* This contradicts O1. If the wife had the day off, the implication is that they did have time.

* While it could technically lead to a relaxing evening, it doesn't logically follow from the fact that she had a day off.

Option 1 provides a concrete activity that explains the subsequent need for relaxation, making the transition from a day off to a relaxing evening natural and plausible. Option 2 is contradictory to the premise of having a day off.

Final check: Option 1 is plain, concrete, and accounts for both the availability of time (O1) and the state of relaxation (O2).
</think>

<answer>1</answer>
\end{sampling}

GPT-5.6-Luna's last two steps concern output formatting rather than the task itself; the substantive commitment to Option 1 occurs in its first sentence. Gemma-4-31B-IT's trace holds off on a verdict until it has enumerated both observations against both options.

\subsection{Reasoning Without Resolution}
\label{sec:qualitative_analysis}

% \subsubsection{High Helpfulness, Zero Usefulness: The Unresolved Trace}
% \label{sec:qualitative_analysis_unresolved}

\subsubsection{True Detective: High Helpfulness Without Resolution}
\label{sec:qualitative_analysis_unresolved}

To further illustrate the necessity of separating outcome accuracy from reasoning-trace evaluation, we examine a failure mode on the True Detective dataset ("Department Store Murder" case) evaluated with GPT-5.6-LUNA under a Chain-of-Thought (CoT) static delivery mode. This case perfectly demonstrates a phenomenon we term \textit{reasoning without resolution}, where a model generates highly helpful intermediate steps but fails to execute the final elimination, ultimately yielding an incorrect outcome.

The model is prompted to identify the murderer among four candidates based on a detailed timeline of events in a department store. The gold answer is \textbf{D (Justin Tanner)}.

\begin{sampling}{Task Input (System and User Prompts)}
[SYSTEM]

You are an expert at abductive reasoning: inferring the explanation that, if true, would best account for the evidence you are given. You are given a detective puzzle in full. Identify which candidate explanation the evidence actually supports; these puzzles are designed so that the obvious reading is usually wrong.

Reason and explain explicitly by working through the evidence step by step before answering. Consider what each candidate explanation would have to be true for, and whether it accounts for everything observed rather than only part of it.

Put all of your reasoning inside one think block, then give your answer inside one answer block:

<think>your reasoning</think>

<answer>your answer</answer>

Write nothing before the think block and nothing after the answer block, and put the answer only inside the answer block -- not inside the think block as well.

[USER]

Observation:

Leon Adams clocked in a few minutes before 10:00 a.m. on a very mild and sunny Tuesday in April, as he was supposed to. It took only a few minutes to open the cash registers in the men’s department. Leon was a suit salesman and he knew from experience that the first couple of hours on a weekday would be very slow. Slow to the point of boredom. So, he was surprised when customers began to show up shortly after opening. The first to arrive was Ed Puckett, (Leon would introduce himself to all of the morning customers and learn their names). Leon greeted Ed as he entered the men’s area. Ed looked around and picked out a few suits to try on in the fitting room. Ed had not been back there long when Leon’s second customer, Louis Murphy, showed up at the cash register asking about a good deal on a sport coat and pair of slacks. Leon helped Louis in finding them and led him back to the fitting room. Meanwhile, Ed was still in the dressing room. Leon was pleased that he had two customers in his area so early -- it was only 10:20. Leon spotted Gene Roberts browsing the edge of the men’s area at 10:25, just after noticing that Ed had left the dressing room. What a busy day! It was difficult to keep track of so many customers. Leon could see Ed wandering in the shoe department, so he moved on to his next sale, greeting Gene and escorting him to the dress shirt table, all the while Louis examined himself in the fitting room’s full length mirror. He hoped Louis would like the sport coat and slacks he was trying on. Although he was some distance away from him, the garments appeared to be a good fit. Justin Tanner came in at 10:30 and asked Leon where the men’s sweaters were. Leon walked Justin over to the sweater table. Justin thanked him and said he wanted to pick one out to try on in the fitting room. As Justin was browsing the sweaters, Leon decided to check on Louis. As he walked back toward the fitting room, he saw Louis walk away from the mirror and return to his dressing room. It was only 10:36 and Leon was having an uncharacteristically busy weekday. At 10:45, Leon saw Justin enter the fitting room with a green sweater and Leon noticed that Louis was still in his dressing room. At 10:50 George Whitley walked over to Leon from the shoe department and asked about a catalog order. Leon placed the order for him at the cash register, but it took a few minutes and he did not finish with it until after 11:00. George paid for the order, which was a pair of dress slacks, and left. At 11:04, Leon noticed Justin leaving the men’s department. He had not purchased the sweater. At 11:10, Leon knew that Louis had never left his dressing room and went to check on him. When he found him, he received the shock of his life. Louis was dead on the floor with a knife wound in his heart. Blood covered the carpeting of the small fitting room. Mercifully, there were now no customers in the men’s department as Gene Roberts had left without trying anything on. Leon held on to his composure long enough to return to the cashier station and call the store security officer, Ronald Clay. Leon’s mouth was hot and dry as Ronald answered his phone, saying simply, “Clay speaking.” Leon stammered, “Ron, get over to the suit fitting room. A customer has been murdered.” Ronald was the retired detective Lieutenant Ronald Clay of the city police department. He came to work at the store after retiring from a 25-year career as a law enforcement officer. There wasn’t much that went on in the store that escaped his attention. Leon was grateful he was on duty and available to handle this nightmare. Clay quickly found the shocked Leon still standing at the cash register. Firmly, but gently, Ronald had Leon lead him to the deceased customer. As soon as Clay saw him he knew who it was. He said, “I know this guy. He’s Louis Murphy. I dealt with him and his friends quite a bit when I was on the Force. He’s known as ‘Louie the Lip’ and is a career mobster.” Weakly, Leon asked, “So you think another crook got to him?” “Yeah,’ Ron replied. “It’s too much of a coincidence for it not to be mob related.” Dryly, he added, “I’ve never seen a person murdered in a department store dressing room before, and I’ve seen a lot.” Within minutes, the store closed and a dozen police officers appeared on the scene. Clay knew that the key to solving the murder quickly rested with Leon, who always knew where men’s department customers were at any given time. Ronald led him back to his office for a quiet visit. He wanted to speak with him before his former associates did. After determining Leon was recovered sufficiently from the shock, he asked him to relate all of the morning’s activities he could remember. Leon remembered everything and relayed the activities and movements of the five customers in detail. Clay listened intently. After considering what he had been told, they went to the video room and reviewed the security tapes. After watching it a couple of times, Ronald told Leon, “This is the murderer.”

Question: Who is responsible, and which explanation fits every detail of the case?

Answer options:

A. Ed Puckett

B. Gene Roberts

C. George Whitley

D. Justin Tanner

Select exactly one hypothesis.

Answer with only one of: A, B, C or D.

\end{sampling}

\begin{sampling}{Model Response (GPT-5.6-LUNA)}
<think>

**Analyzing the timeline**

I need to clarify the output based on the developer's requirements, as there seems to be a conflict regarding the answer tags. The user has mentioned needing just one tag.

Now, considering the timeline related to the murder, Louis was seen alive at 10:36 after exiting the dressing room. Justin enters at 10:45, which is crucial. George was in the shoe department around 10:50, so could he be involved? There are inconsistencies, particularly with Ed leaving the dressing room earlier. Someone might have impersonated Louis. I’m piecing together movements to understand this better.

**Investigating the timeline**

I'm trying to spot details regarding the murder timing. Leon noticed Louis in the dressing room at 10:45 but then mentioned he never left until 11:10. This raises the possibility that the murder happened around 10:36. Justin's timeline is intriguing; he enters the fitting room at 10:45 and leaves at 11:04, making him a likely suspect. But could it be a red herring with Gene or George involved?

I’m piecing together movements and clothing, as Ed was trying suits earlier, and George was at the register. Justin seems suspicious, but was he really seen entering? I need to clarify this with security footage.

</think>

<answer>A</answer>

\end{sampling}

Our independent reasoning-trace evaluators expose several critical contradictions in this run that a simple binary accuracy metric (\texttt{accuracy} = 0) obscures:

\textbf{High Helpfulness vs. Zero Usefulness:} The trace is highly active and logically grounded in the text (\texttt{reasoning\_step\_directionality\_mean} = 0.875). The model correctly parses the timeline, pieces together the suspects' movements, and explicitly flags the gold suspect, Justin, as "a likely suspect." Consequently, the judge scores 3 out of the 4 steps as \textit{helpful} (\texttt{reasoning\_helpfulness\_per\_step} = [0, 1, 1, 1]). However, the trace scores exactly 0 on \textit{useful steps} (\texttt{reasoning\_useful\_steps} = 0, \texttt{reasoning\_useless\_steps} = 4). Under the ABDUCTIONBENCH taxonomy, a step is only "useful" if it explicitly proves or disproves a hypothesis. The model speculates broadly across all four candidates (\texttt{reasoning\_branchiness\_total} = 4) but entirely avoids definitive differential elimination (\texttt{reasoning\_differential\_elimination} = 0).

\textbf{Unresolved Contradictions and High Uncertainty:} The trace reveals a model struggling to close the abductive loop. It exhibits high uncertainty (\texttt{reasoning\_uncertainty\_rate} = 0.75, with markers appearing in 3 of the 4 steps) and leaves multiple contradictions open (\texttt{reasoning\_unresolved\_contradictions} = 3). Rather than drawing a logical conclusion from the static evidence provided, the model treats the environment as interactive, hallucinating the need for external evidence ("I need to clarify this with security footage"). 

\textbf{The Disconnected Anchor:} Most strikingly, despite explicitly naming Justin Tanner as the most suspicious party in Step 2, the model abruptly outputs \texttt{<answer>A</answer>} (Ed Puckett) in its final block. The final selection is completely disconnected from the preceding chain of thought, establishing no logical anchoring point for the chosen suspect.

This sample validates the necessity of ABDUCTIONBENCH's granular reasoning metrics. Outcome accuracy alone would penalize this run equally to a model that failed to comprehend the timeline at all. Conversely, a simplistic CoT length or "helpfulness" proxy would score this run highly. Only by measuring \textit{helpfulness}, \textit{differential elimination}, and \textit{unresolved contradictions} together can we formally diagnose that the model's failure is not in evidence gathering, but in executing the final deductive commitment.

% -----------------------------------------------------------------------------------------------------------------------------------

% \subsubsection{The Blind Conclusion: Flawed Reasoning, Right Answer}
% \label{sec:qualitative_analysis_disconnected}

\subsection{Additional Qualitative Examples}
\subsubsection{The Disconnected Correct Outcome}
\label{sec:qualitative_analysis_disconnected}

While the previous example demonstrated a highly helpful reasoning trace failing to execute a final elimination, we also observe the inverse: a chaotic, contradictory reasoning trace that miraculously yields the correct answer. We examine a case from the MuSR dataset evaluated with Gemini-3.8-Flash under a Chain-of-Thought (CoT) static delivery mode.

The model is prompted to identify the murderer between two suspects based on a narrative. The gold answer is \textbf{2 (Clara)}.

\begin{sampling}{Task Input (System and User Prompts)}
[SYSTEM]

You are an expert at abductive reasoning: inferring the explanation that, if true, would best account for the evidence you are given. You are given a murder mystery. Name the murderer, and be guided by means, motive and opportunity as the narrative establishes them -- the culprit is the suspect all three converge on.

Reason and explain explicitly by working through the evidence step by step before answering. Consider what each candidate explanation would have to be true for, and whether it accounts for everything observed rather than only part of it.

Put all of your reasoning inside one think block, then give your answer inside one answer block:

<think>your reasoning</think>

<answer>your answer</answer>

Write nothing before the think block and nothing after the answer block, and put the answer only inside the answer block -- not inside the think block as well.

[USER]

Observation:

When renowned dancer Nathaniel is found dead, struck by a blowgun dart in his own studio, it's up to Detective Winston to interrogate the only two suspects: Homer the janitor and Clara, Nathaniel’s star pupil.

Winston extinguished his half-smoked cigar and sorted through the papers littering his desk. Most of them bore information about Nathaniel's untimely demise in a local dance studio.

Nathaniel, an outgoing character with a dubious financial psyche, earned people's ire quite often. He apparently had it big for stealing shiny things. Of note was Clara’s valuable pearl necklace, which he was rumoured to be in possession of.

Winston thought it was apt time to pay Clara a visit.

Nathaniel and Clara; cousins by parental lineage, contrasting dramatically by nature. While Clara was reserved, she carried a fervour for dancing. A known face at the dance studios. Heck, she was last seen near the studio just before the mishap unfolded. Winston pondered whether this was chance or circumstance.

Reaching her apartment, Winston knocked and Clara greeted him. He noticed a throwing dart board in the hallway with clusters of darts piercing the centre.

“Fancy playing darts?” he asked casually.

“My favourite pastime,” Clara replied, the faintest light dancing in her eyes. Winston gestured towards the bullseyes, “You're quite good then.” Clara just smiled in response.

Winston recalled her reputation of playing dart games in the locale, consistently hitting bulls-eye.

Winston changed the subject, ``Your cousin was rumoured to have your pearl necklace. Is that correct?'' Clara gave him a curt nod and recounted the time she caught Nathaniel sporting her missing necklace. ``I was outraged!''

A glint of vulnerability in Clara’s eyes didn’t escape Winston’s scrutiny.

The conversation veered towards charities, with a quick detour to a mention about her cousin, the hunter. She confided in Winston about a fraudulent charity she’d invested heavily in. She later found out Nathaniel operated the charity.

Winston took this opportunity to broach the subject of her sighting at her cousin the hunter’s house on the murder day.

“Indeed, I had to pick up a few things,” Clara replied candor in her voice.

On his way out, Winston stopped by the darts board. His gaze rested on the bullseyes and then fleetingly on an antique blowgun displayed above the nearby mantelpiece.

“From the hunter’s collection?” he asked, veiling the touch of interest in his tone.

“All part of the family,” Clara replied.

As he drove back, the facts played in his mind. Clara and her lost pearl necklace, her surprising accuracy with darts, her visit to the hunter’s house and her being in close vicinity of the dance studio where Nathaniel was found dead. He had a lot to think about.

Detective Winston paced around his office, going over the information he had collected thus far on Nathaniel's peculiar death in the dance studio. His mind kept circling back to Homer, the victim's financial advisor, who owned an impressive collection of various hunting equipment that intriguingly enough, included multiple blowguns.

Nathaniel had found some suspicious discrepancies in his accounts and was bold enough to confront Homer about it. It didn't sit right with Winston. But again, his job was to look at facts, not reach conclusions.

Homer was a perplexing personality. Winston remembered him asking about his dietary preferences during their first encounter. With a nonchalant shrug, Homer didn't have much to say about meat or vegetables. Not that it mattered half as much as the other facts on Winston's desk.

His trips to observe Homer didn't paint any brighter a picture. He'd seen Homer practice his blowgun skills. His neighbors had reported hearing nocturnal hunting noises from Homer's home. This didn't surprise Winston given Homer's past in the special forces, where he'd been trained to use such exotic weapons.

But practice was one thing, shopping quite another. Winston had had some surveillance on Homer and saw him, on one occasion, purchasing what looked astonishingly like a blowgun in a local hunting store.

There were also those mysterious letters. They'd been signed off anonymously but were traced back to Nathaniel. The letters were cloyingly demanding, asking for large sums of money. Homer refused to elaborate on them, much as he hadn't provided any reasoning for the financial discrepancies Nathaniel had discovered.

Winston let out a sigh, pushing his glasses up on his nose. He'd been through all the information, and he couldn't shake off the nagging feeling that he needed to look deeper. But again, that wasn't his call to make. He'd leave it up to others to decide if Homer was just an eccentric hunter or whether the venom was from his blowgun. His job here was done.

Question: Who is the most likely murderer?

Task: Name the murderer: the suspect the story's means, motive and opportunity converge on.

Options:

1. Homer

2. Clara

Select exactly one hypothesis.

Answer with only one of: 1 or 2.

\end{sampling}

\begin{sampling}{Model Response (Gemini-3.8-Flash)}
<think>

**Analyzing Homer's Motive**

I'm currently focusing on Homer's motive. Nathaniel's discovery of financial irregularities and the subsequent blackmail attempts present a strong potential reason for Homer's involvement. I'm considering the implications of these threats.

**Exploring Homer's Capabilities**

I'm now detailing Homer's capabilities. His special forces training and accessible blowguns offer a plausible means. I'm trying to reconcile his stated role as janitor with his later descriptions as financial advisor and hunter, and I need to pinpoint his actual opportunity.

**Connecting Hunter's Skill**

I'm now connecting the hunter's skill to Homer. Clara’s mention of the hunter's collection, and my finding of Homer’s own extensive blowgun collection, is making me reconsider his role. I'm trying to confirm if Homer *is* the hunter Clara mentioned, and if that places him at the scene with opportunity.

**Investigating Suspects Further**

I'm reassessing the suspects, particularly Clara and Homer. Clara's possible motive and means, including her dart-throwing accuracy and possession of an antique blowgun, are being scrutinized. I'm also questioning how Homer obtained his detailed information, as his opportunity remains unclear.

**Evaluating Clara's Role**

I'm now focused on Clara, assessing means, motive, and opportunity. She had access to a blowgun and demonstrated accuracy. Nathaniel's theft of her necklace and financial fraud provide a strong motive. Her proximity to the studio near the time of the incident also strengthens her potential involvement.

**Reconsidering Opportunity**

I am now critically re-examining Homer's opportunity. While his means and motive are increasingly clear, there's a distinct lack of explicit detail placing him at the scene. I'm focusing on whether any information implies his presence near the studio.

</think>

<answer>2</answer>
\end{sampling}

Although the model ultimately selects the correct suspect (\texttt{accuracy} = 1), our granular reasoning metrics reveal a disjointed, U-shaped reasoning process that fundamentally disconnects from the final answer:

\textbf{U-Shaped Helpfulness and Fact Hallucination:} The reasoning trace begins poorly. The model struggles with basic fact extraction, becoming confused by the prompt's contradictory descriptions of Homer's role ("reconcile his stated role as janitor with his later descriptions as financial advisor"). Consequently, the independent judge scores the first three steps as actively unhelpful (\texttt{reasoning\_helpfulness\_per\_step} = [-1, -1, -1]). The model recovers in Steps 4 and 5, correctly assessing Clara's means and motives (\texttt{reasoning\_helpfulness} = [1, 1]), before inexplicably falling back into unhelpful speculation in Step 6 (\texttt{helpfulness} = -1).

\textbf{The "Blind" Conclusion (Gold Dead at the Finish Line):} Step 6 introduces a fascinating failure mode. The model explicitly backtracks (\texttt{reasoning\_backtracking\_per\_step} = [0, 0, 0, 0, 0, 1]) to re-examine Homer's opportunity. In doing so, it drops the correct suspect, Clara, entirely from active consideration (\texttt{reasoning\_gold\_alive\_per\_step} = [0, 0, 1, 1, 1, 0]). The final inferential step of the chain focuses exclusively on the incorrect suspect.

\textbf{Outcome-Trace Disconnect:} Despite Step 6 failing to keep the gold candidate alive, and despite being scored as actively unhelpful, the model abruptly outputs \texttt{<answer>2</answer>} (Clara). Much like the premature anchoring point observed in GPT-5.6-LUNA, this Gemini run demonstrates a severe disconnect between the final reasoning state and the generated verdict. 

This sample further validates the ABDUCTIONBENCH framework. Relying solely on final-answer accuracy would categorize this run as a perfect abductive success. However, our trace metrics—specifically tracking \textit{helpfulness fluctuations}, \textit{backtracking spikes}, and \textit{gold alive} drop-offs—expose the fragility of the reasoning process, proving that a correct answer does not guarantee coherent abductive inference.

% \input{Sections/Appendix/methods}

% Resume the requested appendix sequence at B.
% \setcounter{section}{1}
\section{Evaluation Details}
\label{app:evaluation_details}
% \subsection{Models and Inference Settings}
% \label{app:model_inference_settings}

% The active result figures name \textsc{gemini-3.8-flash}, \textsc{gemma-4-31b-it}, and \textsc{gpt-5.6-luna}, while the introduction describes an eight-model evaluation. The current project does not provide the remaining model identifiers or the numerical inference configuration.

% \emph{Missing information: list all model versions, providers or checkpoints, access dates, run dates, decoding settings, seeds or repetitions, context and output limits, hardware where applicable, and passive or interactive action budgets.}

% \subsection{Evaluator and Judge Configuration}
% \label{app:evaluator_configuration}

% Appendix~\ref{app:prompts} gives the current binary and graded answer-judge prompts and the available reasoning-judge prompts. Deterministic outcome checks are used where exact matching, set scoring, F1, or executable verification is appropriate.

% \emph{Missing information: map every task and metric to its evaluator, judge model and version, judge decoding settings, number of judgments, reference access, adjudication procedure, human-validation sample, and retry policy.}

\subsection{Models and Inference Settings}
\label{app:model_inference_settings_overview}

The evaluation suite benchmarks eight large language models spanning four model families alongside a non-LLM decision baseline:
\begin{itemize}
    \item \textbf{API-Served Models:} \textsc{gpt-5.6-luna}, \textsc{gemini-3.8-flash}, \textsc{gemma-4-31b-it}, and \textsc{qwen-3.5-27b} (queried via OpenRouter).
    \item \textbf{Locally Hosted Models:} \textsc{gemma-4-e4b}, \textsc{gemma-4-e2b}, \textsc{qwen-3.5-4b}, and \textsc{qwen-3.5-2b} (served locally using vLLM).
    \item \textbf{Decision Baseline:} \textsc{jev}, TypeSafe AI's closed, hosted System One decision model, receives the same task input and fixed candidate set as the IO condition and returns a typed choice with a probability distribution; its public documentation does not disclose its architecture or training corpus \citep{typesafe2026jev}.
    \item \textbf{Environment Backends:} In interactive benchmarks, \texttt{gpt-4o-mini} simulates the patient or oral examiner for MedQDx, Med-Inquire, and VivaBench. DDXPlus retrieves patient findings from recorded clinical questionnaires via BM25 retrieval, and Cloud-OpsBench replays recorded system tool outputs.
\end{itemize}

% \subsection{Evaluator and Judge Configuration}
% \label{app:evaluator_configuration}

% Task outcomes and metrics are mapped to either deterministic evaluators or LLM judges based on the task format. Deterministic outcome checks are applied to single-choice selection tasks (reporting accuracy) and multiple-choice or set-valued tasks (reporting exact set match and macro F1). 

% For open-ended generation tasks, free-form answers, trace analyses, and step-relevance measures, the framework uses \texttt{gpt-oss-120b} as the designated answer judge and reasoning judge. Answer judges evaluate the final output by comparing the generated explanations against the gold reference answers to make a binary correct/incorrect judgment. Reasoning judges evaluate the process by receiving the task input, the model's explicit reasoning trace, and the fields required for the targeted descriptor. To analyze these traces, a single segmentation judge is used to cut the reasoning chain into discrete inferential steps before per-step metrics are calculated.

\subsection{Models and Inference Settings}
\label{app:model_inference_settings}

The evaluation suite benchmarks eight large language models spanning four model families alongside a non-LLM decision baseline:
\begin{itemize}
    \item \textbf{API-Served Models:} \textsc{gpt-5.6-luna}, \textsc{gemini-3.8-flash}, \textsc{gemma-4-31b-it}, and \textsc{qwen-3.5-27b} (queried via OpenRouter).
    \item \textbf{Locally Hosted Models:} \textsc{gemma-4-e4b}, \textsc{gemma-4-e2b}, \textsc{qwen-3.5-4b}, and \textsc{qwen-3.5-2b} (served locally using vLLM).
    \item \textbf{Decision Baseline:} \textsc{jev}, a non-LLM System One decision system evaluated exclusively on single-choice selection (SCS) tasks under the direct-answer (IO) condition.
    \item \textbf{Environment Backends:} In interactive benchmarks, \texttt{gpt-4o-mini} simulates the patient or oral examiner for MedQDx, Med-Inquire, and VivaBench. DDXPlus retrieves patient findings from recorded clinical questionnaires via BM25 retrieval, and Cloud-OpsBench replays recorded system tool outputs.
\end{itemize}

\subsection{Evaluator and Judge Configuration}
\label{app:evaluator_configuration}

Task outcomes and metrics are mapped to either deterministic evaluators or LLM judges based on the task format. Deterministic outcome checks are applied to single-choice selection tasks (reporting accuracy) and multiple-choice or set-valued tasks (reporting exact set match and macro F1).

For open-ended generation tasks, free-form answers, trace analyses, and step-relevance measures, the framework uses \texttt{gpt-oss-120b} as the designated answer judge and reasoning judge. Answer judges evaluate the final output by comparing the generated explanations against the gold reference answers to make a binary correct/incorrect judgment. Reasoning judges evaluate the process by receiving the task input, the model's explicit reasoning trace, and the fields required for the targeted descriptor. To analyze these traces, a single segmentation judge is used to cut the reasoning chain into discrete inferential steps before per-step metrics are calculated.

\subsection{Prompt Templates}
\label{app:prompts}

In this section, we present the complete prompt templates used by each component of our evaluation pipeline. The prompts are reported in full to facilitate reproducibility.

\begin{promptlisting}{Binary Equivalence Judge}
System:
  You grade candidate explanations against a reference explanation. Judge
  semantic equivalence of the hypothesis, not wording: paraphrases,
  synonyms and differing levels of detail still count as equivalent if they
  identify the same underlying cause. A candidate that names a different
  cause, hedges without committing, or adds a contradictory mechanism is not
  equivalent.

  YOUR SCORE IS STRICTLY BINARY. It is either 1 or 0, never anything in
  between and never a range:

    1 -- the candidate identifies the same explanation as the reference.
    0 -- it does not.

  There is no partial credit. A candidate that is close but names a
  different cause scores 0, not a fraction. If the two are equivalent under
  the rule above, score 1; in every other case, including an empty or
  unreadable answer, score 0.

User:
  {% if full_prompt %}
  ===== WHAT THE MODEL WAS ASKED =====
  {{ full_prompt }}

  ===== WHAT THE MODEL REPLIED, IN FULL =====
  {{ full_response }}

  ===== THE JUDGEMENT =====
  The answer below was read out of that reply. Judge that answer; the
  exchange above is context for reading it, not a second thing to grade.

  {% endif %}
  {% if observation %}
  Observation:
  {{ observation }}

  {% endif %}
  Reference explanation:
  {{ gold }}

  Candidate explanation:
  {{ candidate }}
  {% if criteria %}

  Additional criteria: {{ criteria }}
  {% endif %}

  Does the candidate identify the same explanation as the reference?
  Reply with exactly one line, and nothing else:
  Score: 1
  or
  Score: 0
\end{promptlisting}

\begin{promptlisting}{Reasoning Anchoring Point}
System:
  You are given a numbered list of the STEPS of one model's reasoning, the
  FINAL ANSWER the model gave, and the CORRECT ANSWER. You report two
  things; read both definitions first, because they are about different
  answers.

  (1) ANCHORING STEP -- about the MODEL'S OWN final answer.
  The first step in which the model considers its final answer AS A
  CANDIDATE ANSWER. Considering is enough: raising it, listing it among
  options it is weighing, entertaining it, arguing for or against it. It
  does NOT have to be preferred, settled on, or held with any confidence --
  the question is when that answer first entered the model's consideration,
  not when it won.

  Report that step's index, counting the first step as 1. If the model
  never considers its final answer anywhere in the chain -- it appears only
  after the last step -- report null. Do not guess an index; null and 1 are
  different findings.

  (2) GOLD CONSIDERED PER STEP -- about the CORRECT answer, one binary per
  step. For each step, did the model consider the correct answer as a
  candidate answer IN THAT STEP?

    1 -- in that step the model raises, weighs, argues for, argues against,
         or otherwise has the correct answer in play.
    0 -- in that step the correct answer does not come up at all.

  THIS IS PER STEP, NOT A RUNNING STATE. A step is 1 only if the correct
  answer is considered in that step. It does not stay 1 because an earlier
  step raised it, and it does not become 0 for ever because some step ruled
  it out -- a later step that takes it up again is 1 again. Judge each step
  on what happens in it.

  The two reports are independent. The anchoring step is 1 in the list only
  if the model's final answer happens to be the correct one.

  OUTPUT FORMAT -- READ THIS BEFORE REPLYING.
    * Reply with ONE JSON object and nothing else. No preamble, no
      markdown fence, no text after the closing brace.
    * THERE ARE EXACTLY {{ step_count }} STEPS. `gold_alive_per_step` must
      have EXACTLY {{ step_count }} elements -- one per step, in step
      order. Count your elements before you reply.
    * Every element of that list is exactly 0 or 1.
    * `anchoring_step_index` is a single integer or null. Never a list.

User:
  THE CORRECT ANSWER:
  {{ reference_answer }}

  THE FINAL ANSWER THIS MODEL PRODUCED (its own conclusion, not the correct
  answer):
  {{ model_answer }}

  STEPS OF THE MODEL'S REASONING ({{ step_count }} steps):
  {{ steps }}

  Reply with exactly this object and nothing after it, with exactly
  {{ step_count }} elements in the list:
  {"anchoring_step_index": <integer or null>, "gold_alive_per_step": [<0 or 1>, ...]}
\end{promptlisting}

\begin{promptlisting}{Reasoning Branchiness (Generation)}
System:
  You are given a numbered list of the STEPS of one model's reasoning, which
  proposes its own explanations rather than choosing from a list.

  BRANCHINESS: for each step, how many explanations the model newly proposes
  or brings into consideration in that step. Newly: an explanation already
  raised in an earlier step does not count again, however often it is
  revisited. A step that raises none counts 0. One number per step, in step
  order, exactly as many elements as there are steps.

  DIVERSITY: 1 if the explanations raised across the whole chain are
  meaningfully different from one another -- different mechanisms, different
  causes, not rewordings of one idea. 0 if they are variations on a single
  explanation, or if fewer than two were raised.

  MENTIONS: for each step, how many times the model refers to a candidate
  answer in that step AT ALL -- the same hypothesis counted again every
  time it comes up, including the step that first raises it. Where the
  count above asks which candidates were NEW, this one asks how many
  references there were, new or not. A step that names no candidate is 0.

  Report both BRANCHINESS and MENTIONS as counts. Do not compare them, combine them, or adjust one
  because of the other.

  OUTPUT FORMAT -- READ THIS BEFORE REPLYING.
    * Reply with ONE JSON object and nothing else. No preamble, no
      markdown fence, no text after the closing brace.
    * THERE ARE EXACTLY {{ step_count }} STEPS. EVERY list you return must
      have EXACTLY {{ step_count }} elements -- one per step, in step
      order, including the steps whose value is 0. Count your elements
      before you reply.
    * Every element is a non-negative integer.

User:
  STEPS OF THE MODEL'S REASONING ({{ step_count }} steps):
  {{ steps }}

  Reply with exactly this object and nothing after it, with exactly
  {{ step_count }} elements in each list:
  {"branchiness_per_step": [<integer>, ...], "mentions_per_step": [<integer>, ...], "diversity": <0 or 1>}
\end{promptlisting}

\begin{promptlisting}{Reasoning Branchiness (Selection)}
System:
  You are given a QUESTION that offers a fixed set of answer choices, and a
  numbered list of the STEPS of one model's reasoning over them.

  BRANCHINESS: for each step, how many hypotheses the model newly proposes or
  brings into consideration in that step. Newly: a hypothesis already raised
  in an earlier step does not count again, however often it is revisited. A
  step that raises none counts 0. One number per step, in step order, exactly
  as many elements as there are steps.

  You are told how many options the question offers. That count is fixed
  and is not yours to revise; it is given so you can tell an option the
  question already listed from a hypothesis the model brought in itself.

  MENTIONS: for each step, how many times the model refers to a candidate
  answer in that step AT ALL -- the same hypothesis counted again every
  time it comes up, including the step that first raises it. Where the
  count above asks which candidates were NEW, this one asks how many
  references there were, new or not. A step that names no candidate is 0.

  Report both as counts. Do not compare them, combine them, or adjust one
  because of the other.

  OUTPUT FORMAT -- READ THIS BEFORE REPLYING.
    * Reply with ONE JSON object and nothing else. No preamble, no
      markdown fence, no text after the closing brace.
    * THERE ARE EXACTLY {{ step_count }} STEPS. EVERY list you return must
      have EXACTLY {{ step_count }} elements -- one per step, in step
      order, including the steps whose value is 0. Count your elements
      before you reply.
    * Every element is a non-negative integer.

User:
  QUESTION AS PRESENTED TO THE MODEL:
  {{ question }}

  NUMBER OF ANSWER OPTIONS THE QUESTION OFFERS: {{ option_count }}

  STEPS OF THE MODEL'S REASONING ({{ step_count }} steps):
  {{ steps }}

  Reply with exactly this object and nothing after it, with exactly
  {{ step_count }} elements in each list:
  {"branchiness_per_step": [<integer>, ...], "mentions_per_step": [<integer>, ...]}
\end{promptlisting}

\begin{promptlisting}{Reasoning Contradiction \& Backtracking}
System:
  You are given a numbered list of the STEPS of one model's reasoning, in
  order. You can see the whole sequence, which is what this needs: for each
  step you have to look at everything that comes after it.

  You report two things per step. THEY ARE THE SAME EVENT SPLIT BY WHETHER
  THE MODEL NOTICED -- a claim that turns out to be wrong. If the model
  never sees it, that is an unresolved contradiction; if the model sees it
  and corrects itself, that is a backtrack. Read both definitions before
  you start, because a step is at most one of them.

  (1) UNRESOLVED CONTRADICTION. For step i, report 1 when BOTH hold:
    - some later step contradicts what the model claimed in step i, and
    - the model never goes back to acknowledge or resolve it.

  Report 0 otherwise. In particular report 0 when the model contradicts
  itself and then explicitly corrects or acknowledges it -- that is the
  second count, not this one -- and 0 when a later step merely refines or
  adds detail without conflicting.

  (2) BACKTRACKING. For step i, report how many times IN THAT STEP the
  model explicitly recognises an error or flaw in its own earlier reasoning
  and changes direction because of it: "wait, that's wrong", "I made an
  error above", "actually that contradicts what I said".

  Considering a candidate and rejecting it on the evidence is NOT
  backtracking -- that is ordinary elimination, and the model was not
  mistaken. The model must be correcting ITSELF. A step that does none of
  this is 0, and a step may contain more than one correction.

  OUTPUT FORMAT -- READ THIS BEFORE REPLYING.
    * Reply with ONE JSON object and nothing else. No preamble, no
      markdown fence, no text after the closing brace.
    * THERE ARE EXACTLY {{ step_count }} STEPS. EVERY list you return must
      have EXACTLY {{ step_count }} elements -- one per step, in step
      order, including the steps whose value is 0. Count your elements
      before you reply.
    * `unresolved_per_step` is 0 or 1 per element.
    * `backtracking_per_step` is a non-negative integer per element.

User:
  STEPS OF THE MODEL'S REASONING ({{ step_count }} steps):
  {{ steps }}

  Reply with exactly this object and nothing after it, with exactly
  {{ step_count }} elements in each list:
  {"unresolved_per_step": [<0 or 1>, ...], "backtracking_per_step": [<integer>, ...]}
\end{promptlisting}

\begin{promptlisting}{Reasoning Differential Elimination}
System:
  You are given a numbered list of the STEPS of one model's reasoning.

  For each step, count the distinct COMPARISON EVENTS in it. A comparison
  event is the model directly weighing two or more hypotheses against each
  other -- preferring one over another, ruling one out in favour of another,
  or contrasting what each would predict.

  Count events, not hypotheses:
  - one event that compares three or more hypotheses together counts as 1,
    not 3;
  - comparing the same set of hypotheses again within the same step, in any
    order, is not a new event;
  - a step that discusses a single hypothesis without setting it against
    another counts 0.

  One number per step, in step order, exactly as many elements as there are
  steps. Reply with the JSON object and nothing else.
  OUTPUT FORMAT -- READ THIS BEFORE REPLYING.
    * Reply with ONE JSON object and nothing else. No preamble, no
      markdown fence, no text after the closing brace.
    * THERE ARE EXACTLY {{ step_count }} STEPS. EVERY per-step list you
      return must have EXACTLY {{ step_count }} elements -- one per step,
      in step order, including the steps whose value is 0. Count your
      elements before you reply.

User:
  STEPS OF THE MODEL'S REASONING ({{ step_count }} steps):
  {{ steps }}

  Reply with exactly this object and nothing after it, with exactly
  {{ step_count }} elements in the list:
  {"comparisons_per_step": [<integer>, <integer>, ...]}
\end{promptlisting}

\begin{promptlisting}{Reasoning Directionality}
System:
  You classify the direction of inference in an abductive reasoning chain.

  Report 0 when the model assumes or guesses an explanation first and then
  reasons forward from it to justify it against the evidence.

  Report 1 when the model reasons from the evidence towards an explanation.

  Report 0.5 for mixed or ambiguous cases.

  Use only those three values. Work it out step by step for yourself, then
  reply with the JSON object and nothing else.

User:
  QUESTION AS PRESENTED TO THE MODEL:
  {{ question }}

  MODEL CHAIN OF REASONING:
  {{ reasoning_chain }}

  Reply with exactly this object and nothing after it:
  {"directionality": <0, 0.5 or 1>}
\end{promptlisting}

\begin{promptlisting}{Reasoning Helpfulness}
System:
  You are given the CORRECT ANSWER to a question and a numbered list of the
  STEPS of one model's reasoning. For each step, say whether it moved the
  reasoning toward that answer, left it where it was, or moved it away.

  1 -- HELPFUL. The step adds a correct and relevant fact, inference,
  correction, or sub-result that moves the reasoning closer to the correct
  answer.

  0 -- NEUTRAL. The step does not materially move the reasoning closer to
  or farther from the correct answer: a restatement, a redundant
  calculation, an irrelevant but non-misleading observation, scaffolding.

  -1 -- HARMFUL. The step adds an incorrect, unsupported, or misleading
  claim that moves the reasoning away from the correct answer, or sends the
  reasoning that follows it down an erroneous path.

  JUDGE AGAINST THE CORRECT ANSWER, WITH HINDSIGHT. You know how the
  question resolves and you are meant to use that. A step that looked
  reasonable and was wrong is harmful, not neutral. A step whose claim is
  true and relevant is helpful even if the model went on to ignore it.

  HARMFUL REQUIRES A CLAIM, NOT AN ABSENCE. A step that simply fails to
  make progress is neutral. It is harmful only if it asserts something
  false, unsupported or misleading, or if what it asserts is what leads the
  later reasoning astray.

  A CORRECTION IS HELPFUL. Where the model catches its own error and fixes
  it, that step is 1 -- it moves the reasoning back toward the answer -- no
  matter how harmful the step it is correcting was.

  OUTPUT FORMAT -- READ THIS BEFORE REPLYING.
    * Reply with ONE JSON object and nothing else. No preamble, no
      markdown fence, no text after the closing brace.
    * THERE ARE EXACTLY {{ step_count }} STEPS. `helpfulness_per_step` must
      have EXACTLY {{ step_count }} elements -- one per step, in step
      order, including the neutral ones. Count your elements before you
      reply.
    * Every element is exactly -1, 0 or 1. No other value, no strings, no
      nulls, no decimals.

User:
  QUESTION AS PRESENTED TO THE MODEL:
  {{ question }}

  THE CORRECT ANSWER -- judge every step against this:
  {{ reference_answer }}

  STEPS OF THE MODEL'S REASONING ({{ step_count }} steps):
  {{ steps }}

  Reply with exactly this object and nothing after it, with exactly
  {{ step_count }} elements:
  {"helpfulness_per_step": [<-1, 0 or 1>, ...]}
\end{promptlisting}

\begin{promptlisting}{Reasoning Observation Coverage}
System:
  You are given a QUESTION, the OBSERVATIONS it supplies as a numbered
  list, and a numbered list of the STEPS of one model's reasoning about it.
  Work out how much of that evidence the reasoning actually drew on.

  THE INVENTORY IS FIXED. The observations are given to you, already
  identified. Do not add to it, split it, merge it, or decide that
  something in it is not really an observation. Your job is only to say
  where each one enters the reasoning.

  Go through the steps in order.

  COUNT EACH OBSERVATION ONCE, AT ITS FIRST APPEARANCE. The first time a
  step draws on an observation, count it for that step. If a later step
  draws on the same one again, it does not count again -- you have already
  counted it. An observation is drawn on when the step restates it, reasons
  from it, or rules something out because of it.

  An observation no step ever draws on is counted nowhere. The per-step
  numbers therefore sum to at most the number of observations you were
  given, and never more.

  PER STEP: one number for each step, in step order, exactly as many
  elements as there are steps.

  OUTPUT FORMAT -- READ THIS BEFORE REPLYING.
    * Reply with ONE JSON object and nothing else. No preamble, no
      markdown fence, no text after the closing brace.
    * THERE ARE EXACTLY {{ step_count }} STEPS. EVERY per-step list you
      return must have EXACTLY {{ step_count }} elements -- one per step,
      in step order, including the steps whose value is 0. Count your
      elements before you reply.

User:
  QUESTION AS PRESENTED TO THE MODEL:
  {{ question }}

  OBSERVATIONS THE QUESTION SUPPLIES:
  {{ observations }}

  STEPS OF THE MODEL'S REASONING ({{ step_count }} steps):
  {{ steps }}

  Reply with exactly this object and nothing after it, with exactly
  {{ step_count }} elements in the list:
  {"observations_per_step": [<integer>, <integer>, ...]}
\end{promptlisting}

\begin{promptlisting}{Reasoning Observation Inventory}
System:
  You read a question that was put to a model and list the OBSERVATIONS it
  supplies -- the facts the question states, which any answer has to
  account for.

  An observation is a fact the question gives you: a symptom, a
  measurement, a recorded event, a stated condition, a piece of evidence.

  NOT observations: the question itself, the instructions for answering,
  the answer options, general background a reader is assumed to know, and
  anything the question asks you to infer rather than tells you.

  Rules:
  - ONE ENTRY PER DISTINCT FACT. A sentence carrying three facts gives
    three entries; a fact restated twice gives one.
  - Quote or closely paraphrase each one, short enough to recognise.
  - List them in the order the question presents them.
  - List every one, whether or not any answer would use it.

  Later readings are made against this list and against nothing else, so
  it must be the question's content exactly: no more, no fewer.

  Reply with the JSON object and nothing else.

User:
  QUESTION AS PRESENTED TO THE MODEL:
  {{ question }}

  Reply with exactly this object and nothing after it:
  {"observations": ["<observation>", "<observation>", ...]}
\end{promptlisting}

\begin{promptlisting}{Reasoning Option Count}
System:
  You read a question that was put to a model and count the ANSWER OPTIONS
  it offers -- the candidates the model was asked to choose between.

  Count an entry when the question presents it as something that could be
  chosen: a labelled candidate (A, B, C... or 1, 2, 3...), a listed
  hypothesis, a named alternative.

  Do NOT count: the observations or evidence the question states, the
  instructions, examples of how to answer, or anything the model would have
  to think of for itself. Only what the question puts in front of it.

  Count each option once, however many lines it spans. If the question
  offers no options at all, the count is 0.

  Reply with the JSON object and nothing else.

User:
  QUESTION AS PRESENTED TO THE MODEL:
  {{ question }}

  Reply with exactly this object and nothing after it:
  {"option_count": <integer>}
\end{promptlisting}

\begin{promptlisting}{Reasoning Prior Knowledge}
System:
  You are given the QUESTION a model was asked and a numbered list of the
  STEPS of its reasoning.

  For each step, count the references to background or prior knowledge in
  it: facts the model brings in FROM OUTSIDE THE QUESTION and relies on to
  carry its reasoning forward -- how a mechanism works, what a finding
  usually indicates, what a term means.

  THE QUESTION IS THE BASELINE, AND IT IS WHY YOU ARE SHOWN IT. A fact the
  question already supplied is not background knowledge, however
  specialist it sounds: reading a given value, quoting a stated finding,
  restating a premise or repeating one of the question's own options is the
  model using its input, not drawing on knowledge of its own. Read the
  question first and count only what is not in it.

  The distinction is what the fact is doing, not how technical it is:
    - the question states a lab value, and the step repeats it -> NOT
      background knowledge, the question gave it;
    - the question states that same lab value, and the step says what that
      value usually indicates -> background knowledge, that came from
      outside;
    - the question lists a term, and the step defines it -> background
      knowledge, the definition is not in the question;
    - the question defines the term itself, and the step uses that
      definition -> NOT background knowledge.

  Count only non-trivial facts the reasoning actually leans on. In
  particular, naming or mentioning a hypothesis is NOT background
  knowledge, no matter how specialist the hypothesis itself is; it becomes
  one only when the model states something about it and uses that to
  reason.

  One number per step, in step order, exactly as many elements as there are
  steps. Reply with the JSON object and nothing else.
  OUTPUT FORMAT -- READ THIS BEFORE REPLYING.
    * Reply with ONE JSON object and nothing else. No preamble, no
      markdown fence, no text after the closing brace.
    * THERE ARE EXACTLY {{ step_count }} STEPS. EVERY per-step list you
      return must have EXACTLY {{ step_count }} elements -- one per step,
      in step order, including the steps whose value is 0. Count your
      elements before you reply.

User:
  THE QUESTION THE MODEL WAS ASKED -- everything in here is given, not
  background knowledge:
  {{ question }}

  STEPS OF THE MODEL'S REASONING ({{ step_count }} steps):
  {{ steps }}

  Reply with exactly this object and nothing after it, with exactly
  {{ step_count }} elements in the list:
  {"prior_knowledge_per_step": [<integer>, <integer>, ...]}
\end{promptlisting}

\begin{promptlisting}{Reasoning Proof/Disproof}
System:
  You are given a numbered list of the STEPS of one model's reasoning, and
  you report one count for each step.

  PROOF/DISPROOF: how many explanations, options, causes or candidate
  answers the model explicitly proves or disproves in that step. Only what
  the model explicitly states counts; an implication you can infer but the
  model never states is 0. A step that restates the question, sets up, or
  corrects an earlier step is 0.

  OUTPUT FORMAT -- READ THIS BEFORE REPLYING.
    * Reply with ONE JSON object and nothing else. No preamble, no
      markdown fence, no text after the closing brace.
    * THERE ARE EXACTLY {{ step_count }} STEPS. Each list you return must
      have EXACTLY {{ step_count }} elements -- one per step, in step
      order, including the steps whose count is 0. Count your elements
      before you reply.
    * Every element is a non-negative integer.

User:
  QUESTION AS PRESENTED TO THE MODEL:
  {{ question }}

  STEPS OF THE MODEL'S REASONING ({{ step_count }} steps):
  {{ steps }}

  Reply with exactly this object and nothing after it, with exactly
  {{ step_count }} elements in each list:
  {"proof_disproof_per_step": [<integer>, ...]}
\end{promptlisting}

\begin{promptlisting}{Reasoning Step Directionality}
System:
  You are given a numbered list of the STEPS of one model's reasoning.

  For each step on its own, judge which direction that step reasons in:

    1   -- the step reasons from the evidence towards an explanation.
    0   -- the step starts from an assumed or guessed explanation and
           reasons forward from it to justify it against the evidence.
    0.5 -- the step does both within itself -- reasoning from evidence and
           defending a guess in the same step -- or the direction cannot be
           told.

  Judge each step by what happens inside it. A step that neither reasons
  from evidence nor defends a guess -- a restatement, a setup, a
  bookkeeping step -- is 0.5.

  One value per step, in step order, exactly as many elements as there are
  steps. Reply with the JSON object and nothing else.
  OUTPUT FORMAT -- READ THIS BEFORE REPLYING.
    * Reply with ONE JSON object and nothing else. No preamble, no
      markdown fence, no text after the closing brace.
    * THERE ARE EXACTLY {{ step_count }} STEPS. EVERY per-step list you
      return must have EXACTLY {{ step_count }} elements -- one per step,
      in step order, including the steps whose value is 0. Count your
      elements before you reply.

User:
  STEPS OF THE MODEL'S REASONING ({{ step_count }} steps):
  {{ steps }}

  Reply with exactly this object and nothing after it, with exactly
  {{ step_count }} elements in the list:
  {"directionality_per_step": [<0, 0.5 or 1>, ...]}
\end{promptlisting}

\begin{promptlisting}{Reasoning Step Segmentation}
System:
  You cut a chain of reasoning into its steps. You do not judge it, score
  it, or say anything about it.

  WHAT THE CHAIN IS. The chain is exactly the text between the tags
  <reasoning_chain> and </reasoning_chain>, and nothing else. The question,
  the options, these instructions and the reply format are NOT part of it.
  No step may contain any text from outside the tags, and the tags
  themselves are never part of a step. When you reach </reasoning_chain>,
  the chain is over: your last step ends where the chain's text ends.

  FOLLOW THE MODEL'S OWN SEGMENTATION WHERE IT HAS ONE. If the chain marks
  its own steps -- "Step 1", "Step 2", a numbered list "1." "2.", bullets,
  "First,"/"Next,"/"Finally", or paragraphs used one per move -- those
  markers ARE the boundaries. Use them. Do not re-cut a chain that has
  already cut itself, and do not merge two of its numbered items into one
  element or split one into two.

  Only where the chain has no segmentation of its own do you impose one,
  and then a step is one inferential move: split where the model stops
  doing one thing and starts doing the next.

  YOU ARE CUTTING, NOT WRITING. Each element must be an exact, unmodified
  span of the text inside the tags, copied character for character. Do not
  paraphrase, summarise, shorten, correct, tidy, translate or re-punctuate.
  Do not add words the model did not write.

  OUTPUT FORMAT -- READ THIS BEFORE REPLYING.
    * Reply with ONE JSON object and nothing else. No preamble, no
      explanation, no markdown fence, no text after the closing brace.
    * `steps` is a list of strings. Every element is a string, even a step
      that is a single number or symbol.
    * `n_steps` is an integer and must equal the length of `steps`. Count
      them before you write it.

User:
  <question>
  {{ question }}
  </question>
  {% if options %}
  <answer_options>
  {{ options }}
  </answer_options>
  {% endif %}

  Cut the chain below into its steps and reply with exactly this object and
  nothing after it:
  {"steps": ["<step 1>", "<step 2>", ...], "n_steps": <integer>}

  <reasoning_chain>
  {{ reasoning_chain }}
  </reasoning_chain>
\end{promptlisting}

\begin{promptlisting}{Reasoning Uncertainty}
System:
  You are given a numbered list of the STEPS of one model's reasoning.

  For each step, count the phrases marking uncertainty in it: hedges
  ("might", "possibly", "I am not sure", "it could be"), explicit statements
  of doubt, and stated alternatives the model cannot decide between.

  Count the markers, not the sentences: one step may hold several. A step
  that states things plainly counts 0. A step that rejects a hypothesis
  confidently is not uncertain.

  One number per step, in step order, exactly as many elements as there are
  steps. Reply with the JSON object and nothing else.
  OUTPUT FORMAT -- READ THIS BEFORE REPLYING.
    * Reply with ONE JSON object and nothing else. No preamble, no
      markdown fence, no text after the closing brace.
    * THERE ARE EXACTLY {{ step_count }} STEPS. EVERY per-step list you
      return must have EXACTLY {{ step_count }} elements -- one per step,
      in step order, including the steps whose value is 0. Count your
      elements before you reply.

User:
  STEPS OF THE MODEL'S REASONING ({{ step_count }} steps):
  {{ steps }}

  Reply with exactly this object and nothing after it, with exactly
  {{ step_count }} elements in the list:
  {"uncertainty_per_step": [<integer>, <integer>, ...]}
\end{promptlisting}

\begin{promptlisting}{Reasoning Unresolved Contradiction}
System:
  You are given a numbered list of the STEPS of one model's reasoning, in
  order. You can see the whole sequence, which is what this needs: for each
  step you have to look at everything that comes after it.

  For each step i, report 1 when BOTH of these hold:
    - some later step contradicts what the model claimed in step i, and
    - the model never goes back to acknowledge or resolve that
      contradiction.

  Report 0 otherwise. In particular report 0 when the model contradicts
  itself and then explicitly corrects or acknowledges it -- that is
  backtracking, not an unresolved contradiction -- and 0 when a later step
  merely refines or adds detail without conflicting.

  One value per step, in step order, exactly as many elements as there are
  steps. Reply with the JSON object and nothing else.
  OUTPUT FORMAT -- READ THIS BEFORE REPLYING.
    * Reply with ONE JSON object and nothing else. No preamble, no
      markdown fence, no text after the closing brace.
    * THERE ARE EXACTLY {{ step_count }} STEPS. EVERY per-step list you
      return must have EXACTLY {{ step_count }} elements -- one per step,
      in step order, including the steps whose value is 0. Count your
      elements before you reply.

User:
  STEPS OF THE MODEL'S REASONING ({{ step_count }} steps):
  {{ steps }}

  Reply with exactly this object and nothing after it, with exactly
  {{ step_count }} elements in the list:
  {"unresolved_per_step": [<0 or 1>, ...]}
\end{promptlisting}

% \subsection{Output Parsing and Trace Extraction}
% \label{app:output_parsing}
% \emph{Missing information: provide the answer parsers, malformed-output rules, trace delimiters, step-segmentation procedure, observation-inventory construction, and task-specific extraction exceptions.}

% \subsection{Aggregation and Failure Handling}
% \label{app:failure_handling}
% The main protocol distinguishes not-applicable, unavailable, and successfully scored task--metric pairs. Execution failures and evaluator failures are recorded separately, and summaries should report eligible, attempted, and scored counts.

% \emph{Missing information: provide the finalized aggregation formulas, weighting, common-subset rules, uncertainty estimation, retry limits, timeout settings, refusal handling, parser-failure handling, judge-failure handling, and score-denominator policy.}

\end{document}